\documentclass[sigplan,10pt]{acmart}

\usepackage{bookmark}

\usepackage{enumitem}
\setitemize{
        noitemsep,
        leftmargin=*,
		topsep=2pt,
		parsep=0pt,
		partopsep=2pt}
		
\usepackage{fontawesome}
\usepackage{colortbl}
\usepackage{wrapfig}
\usepackage{graphicx}
\usepackage{footnote}
\usepackage{tcolorbox}
\usepackage[T1]{fontenc}
\usepackage{doi}
\setlist{nosep} 
\setlist{noitemsep}
\usepackage{footnote}
\makesavenoteenv{tabular}

\usepackage{amsmath}
\usepackage{amsfonts}

\usepackage{algorithmicx}
\usepackage{multirow}
\usepackage{multicol}
\usepackage{booktabs}

\usepackage{array}
\usepackage{xspace}
\usepackage{bigstrut}
\usepackage[export]{adjustbox}

\usepackage[caption=false,font=footnotesize]{subfig}

\usepackage{stfloats}

\usepackage{boldline}

\usepackage{cleveref}
\crefformat{section}{\S#2#1#3}
\crefmultiformat{section}{\S\S#2#1#3}{and~#2#1#3}{, #2#1#3}{, and~#2#1#3}
\Crefname{equation}{Eq.}{Eqs.}
\Crefname{figure}{Fig.}{Figs.}
\Crefname{tabular}{Tab.}{Tabs.}
\crefname{algocf}{pseudocode}{pseudocodes}
\Crefname{algocf}{Pseudocode}{Pseudocodes}

\definecolor{gray05}{gray}{0.95}
\definecolor{gray10}{gray}{0.90}
\definecolor{gray12}{gray}{0.88}
\definecolor{gray15}{gray}{0.85}
\definecolor{gray20}{gray}{0.80}
\definecolor{gray25}{gray}{0.75}
\definecolor{gray30}{gray}{0.70}
\definecolor{gray40}{gray}{0.60}
\definecolor{gray50}{gray}{0.50}
\definecolor{gray60}{gray}{0.40}
\definecolor{gray70}{gray}{0.30}
\definecolor{gray75}{gray}{0.25}
\definecolor{gray80}{gray}{0.20}
\definecolor{gray85}{gray}{0.15}
\definecolor{gray90}{gray}{0.10}
\definecolor{gray95}{gray}{0.05}

\newcommand\eq[1]{\Cref{eq:#1}\xspace}
\newcommand\fig[1]{\Cref{fig:#1}\xspace}
\newcommand\tab[1]{\Cref{tab:#1}\xspace}
\newcommand\tion[1]{\Cref{sect:#1}\xspace}

\newcommand{\etal}{\textit{et al.}\xspace}
\newcommand{\ie}{i.e.}
\newcommand{\eg}{e.g.}

\newcommand{\tool}{\textsc{Unicorn}\xspace}
\newcommand{\ourapproach}{\textsc{Unicorn}\xspace}
\newcommand{\ourtool}{$\textsc{Unicorn}_{\textsc{Tool}}$\xspace}

\newcommand{\encore}{\textsc{EnCore}\xspace}
\newcommand{\bugdoc}{\textsc{BugDoc}\xspace}
\newcommand{\cbi}{\textsc{CBI}\xspace}

\newcommand\txone{\textsc{TX1}\xspace}
\newcommand\txtwo{\textsc{TX2}\xspace}
\newcommand\xavier{\textsc{Xavier}\xspace}

\newcommand{\be}{\begin{enumerate}}
\newcommand{\smalleq}{\begin{equation}\small}
\newcommand{\smallereq}{\begin{equation}\footnotesize}
\newcommand{\eeq}{\end{equation}}
\newcommand{\beqml}{\begin{multline}}
\newcommand{\eeqml}{\end{multline}}
\newcommand{\besq}{\begin{enumerate}[leftmargin=*,wide=0pt,topsep=0pt]}
\newcommand{\ee}{\end{enumerate}}
\newcommand{\bi}{\begin{itemize}}
\newcommand{\bicirc}{\begin{itemize}[leftmargin=*]\renewcommand\labelitemi{$\circ$}}
\newcommand{\bisq}{\begin{itemize}[leftmargin=*,wide=0pt,topsep=1pt]}
\newcommand{\ei}{\end{itemize}}

\makeatletter
\edef\textFontName{\fontname\csname
  \f@encoding/\f@family/\f@series/\f@shape/\f@size\endcsname}

\newcommand{\removelatexerror}{\let\@latex@error\@gobble}

\makeatother

\usepackage{tikz}
\usetikzlibrary{arrows.meta}

\newcommand\edgeone{
\begin{tikzpicture}
  \draw[black,  arrows={-Triangle[angle=90:3pt,black,fill=black]}] (0,0.0) -- (0.5,0.0);
\end{tikzpicture}\xspace}

\newcommand\edgetwo{
\begin{tikzpicture}
  \draw[black,  arrows={Triangle[angle=90:3pt,black,fill=black]-Triangle[angle=90:3pt,black,fill=black]}] (0,0.0) -- (0.5,0.0);   
\end{tikzpicture}\xspace}

\newcommand\edgethree{
\begin{tikzpicture}
  \draw[black,  arrows={Circle[open]-Triangle[angle=90:3pt,black,fill=black]}] (0,0.0) -- (0.5,0.0);   
\end{tikzpicture}\xspace}

\newcommand\edgefour{
\begin{tikzpicture}
  \draw[black,  arrows={Circle[open]-Circle[open]}] (0,0.0) -- (0.5,0.0);   
\end{tikzpicture}\xspace}

\usepackage{amsthm}
\DeclareMathOperator*{\argmax}{argmax}
\usepackage{thmtools}
\usepackage{tikz}
\usetikzlibrary{arrows.meta}
\declaretheoremstyle[headfont=\scshape]{schead}
\declaretheoremstyle[headfont=\bf]{bfhead}

\usepackage[linesnumbered,ruled,vlined, noend]{algorithm2e}

\SetCommentSty{mycommfont}
\SetAlCapNameFnt{\footnotesize}
\SetAlCapFnt{\footnotesize}
\SetAlgorithmName{Pseudocode}{Pseudocode}{Pseudocode of Algorithms}

\usepackage[framemethod=tikz]{mdframed}
\usetikzlibrary{shadows}
\usepackage{graphics}
\newmdenv[
    tikzsetting= {fill=gray05!0},
    skipabove=0.33em,
    skipbelow=0.33em,
    linewidth=1pt,
    innerleftmargin=4pt,
    innerrightmargin=4pt,
    innertopmargin=2pt,
    innerbottommargin=2pt,
    linecolor=gray85,
    roundcorner=2pt, 
    shadow=true,
    shadowsize=4pt,
    shadowcolor=black
]{myshadowbox}
\newmdenv[
    tikzsetting= {fill=gray05!0},
    skipabove=0.33em,
    skipbelow=0.33em,
    linewidth=1.25pt,
    innerleftmargin=4pt,
    innerrightmargin=4pt,
    innertopmargin=2pt,
    innerbottommargin=2pt,
    linecolor=gray85,
    roundcorner=3pt, 
    shadow=false,
    shadowsize=3pt,
    shadowcolor=black
]{mygoalbox}
\usepackage{tikz}

\usepackage{url}

\usepackage{titlesec}
\author{Md Shahriar Iqbal}
\affiliation{%
  \institution{University of South Carolina}}
\email{miqbal@email.sc.edu}

\author{Rahul Krishna}
\affiliation{%
  \institution{IBM Research}}
\email{rkrsn@ibm.com}

\author{Mohammad Ali Javidian}
\affiliation{%
  \institution{Purdue University}}
\email{mjavidia@purdue.edu}

\author{Baishakhi Ray}
\affiliation{%
  \institution{Columbia University}}
\email{rayb@cs.columbia.edu}

\author{Pooyan Jamshidi}
\affiliation{%
  \institution{University of South Carolina}}
 \email{pjamshid@cse.sc.edu}

\AtBeginDocument{%
  \providecommand\BibTeX{{%
    \normalfont B\kern-0.5em{\scshape i\kern-0.25em b}\kern-0.8em\TeX}}}

\copyrightyear{2022}
\acmYear{2022}
\setcopyright{rightsretained}
\acmConference[EuroSys '22]{Seventeenth European Conference on Computer Systems}{April 5--8, 2022}{RENNES, France}
\acmBooktitle{Seventeenth European Conference on Computer Systems (EuroSys '22), April 5--8, 2022, RENNES, France}\acmDOI{10.1145/3492321.3519575}
\acmISBN{978-1-4503-9162-7/22/04}
\begin{document}
\title{\ourapproach: Reasoning about Configurable System Performance through the Lens of Causality}

\graphicspath{{figures-vg/}}
\begin{abstract}

Modern computer systems are highly configurable, with the total variability space sometimes larger than the number of atoms in the universe. Understanding and reasoning about the performance behavior of highly configurable systems, over a vast and variable space, is challenging. State-of-the-art methods for performance modeling and analyses rely on predictive machine learning models, therefore, they become (i)~\emph{unreliable in unseen environments} (e.g., different hardware, workloads), and (ii)~\emph{may produce incorrect explanations}. To tackle this, we propose a new method, called \ourapproach, which (i)~\emph{captures intricate interactions} between configuration options across the software-hardware stack and (ii)~describes how such interactions can impact \emph{performance variations} via causal inference. We evaluated \ourapproach on six highly configurable systems, including three on-device machine learning systems, a video encoder, a database management system, and a data analytics pipeline. The experimental results indicate that \ourapproach outperforms state-of-the-art performance debugging and optimization methods in finding effective repairs for performance faults and finding configurations with near-optimal performance. Further, unlike the existing methods, the learned causal performance models reliably predict performance for new environments.  

\end{abstract}

\maketitle

\section{Introduction}
 \begin{figure}[tp!]
  \subfloat[]{
	\begin{minipage}[c][1\width]{
	   0.3\linewidth}
	   \centering
	   \includegraphics[width=\textwidth]{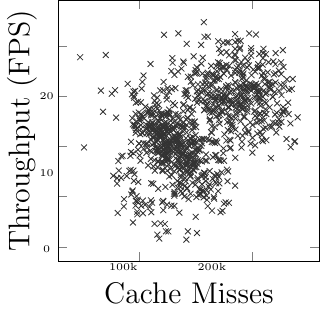}
	\end{minipage}}
 \hfill 	
  \subfloat[]{
	\begin{minipage}[c][1\width]{
	   0.3\linewidth}
	   \centering
	   \includegraphics[width=\textwidth]{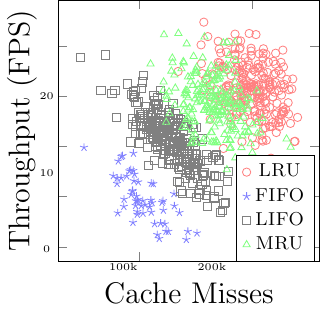}
	\end{minipage}}
 \hfill	
  \subfloat[]{
	\begin{minipage}[c][1\width]{
	   0.3\linewidth}
	   \centering
	   \includegraphics[width=\textwidth]{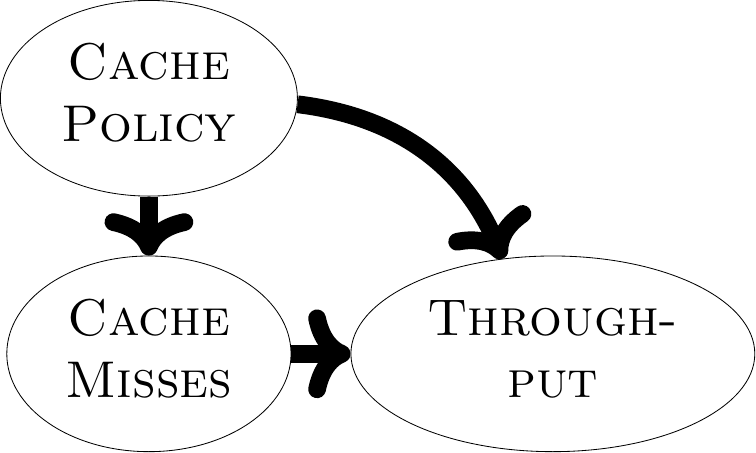}
	\end{minipage}}
\caption{\small{An example showing the effectiveness of causality in reasoning about system performance behavior. 
    (a)  Observational data shows that the increase in \texttt{Cache Misses} 
    leads to high \texttt{Throughput} and such trend is typically captured by statistical reasoning in ML models; (b) incorporating \texttt{Cache Policy} as a confounder correctly shows increase of \texttt{Cache Misses} corresponding to decrease in \texttt{Throughput}; (c) the corresponding causal model correctly captures \texttt{Cache Policy} as a common cause to explain performance behavior.}}
    \label{fig:intro_fig}
\end{figure}

Modern computer systems, such as data analytics pipelines, are typically composed of multiple components, where each component has a plethora of configuration options that can be deployed individually or in conjunction with other components on different hardware platforms. The configuration space of such highly configurable systems is combinatorially large, with 100s if not 1000s of software and hardware configuration options that interact non-trivially with one another~\cite{wang2018understanding,halin2019test,JC:MASCOTS16,velez2022study}. Individual component developers typically have a relatively localized, and thus limited, understanding of the performance behavior of the systems that comprise the components. Therefore, developers and end-users of the final system are often overwhelmed with the complexity of composing and configuring components, making it challenging and error-prone to configure these systems to reach desired performance goals.



Incorrect configuration (\textit{misconfiguration}) elicits unexpected interactions between software and hardware, resulting in \textit{non-functional faults}\footnote{ \emph{Non-functional} and \emph{Performance faults} are used interchangeably to refer to severe performance degradation caused by certain type of misconfigurations, (aka. specious configuration)~\cite{hu2020automated}.}, \ie, degradations in \textit{non-functional} system properties like latency and energy consumption. These non-functional faults, unlike regular software bugs, do not cause the system to crash or exhibit any obvious misbehavior~\cite{reddy2016fault, tsakiltsidis2016automatic, nistor2013discovering}. Instead, misconfigured systems remain operational but degrade in performance~\cite{bryant2003computer, molyneaux2009art,sanchez2020tandem,nistor2015caramel} that can cause major issues in cloud infrastructure~\cite{amazon:config:outage}, internet-scale systems~\cite{fb:config:outage}, and on-device machine learning (ML) systems~\cite{slowimag79:online}. For example, a developer complained that \textit{``I have a complicated system composed of multiple components running on NVIDIA Nano and using several sensors and I observed several performance issues.~\cite{super_frustrated_power_perf:online}.''} 
In another instance, a developer asks \textit{``I’m quite upset with CPU usage on Jetson TX2 while running TCP/IP upload test program''~\cite{HighCPUu7:online}}. 
After struggling in fixing the issues over several days, the developer concludes, \textit{``there is a lot of knowledge required to optimize the network stack and measure CPU load correctly. I tried to play with every configuration option explained in the kernel documents.''}
In addition, they would like to \emph{understand} the impact of configuration options and their interactions, e.g., \textit{``What is the effect of swap memory on increasing throughput?~\cite{slowimag79:online}''}.

\noindent\textbf{Existing works and gap.} Understanding the performance behavior of configurable systems can enable (i) performance debugging~\cite{SGAK:ESECFSE15,GSKA:SPPEXA}, (ii) performance tuning~\cite{H:CACM,SHM:LIO,H:AS,VPGFC:ICPE17,HPHL:ICSE15,WWHJK:GECCO15,NMSA:Arxive17,MARC:SPLC13,ORGC:SPLC14}, and (iii) architecture adaptation~\cite{JGAP:CSMR13,ABDD:VISSOFT13,KK:WSR16,JVKSK:SEAMS17,HSCMAR:SIGPLAN,FHM:FSE15,EEM:TSE,EEM:FSE10}. 
A common strategy is to build performance influence models such as regression models that explain the influence of individual options and their interactions~\cite{SGAK:ESECFSE15,VPGFC:ICPE17,GCASW:ASE13,pereira2019learning}. These approaches are adept at inferring the correlations between configuration options and performance objectives, however, as illustrated in~\fig{intro_fig} performance influence models suffer from several shortcomings (detailed in \S\ref{sec:motivation}): (i) they become ~\emph{unreliable in unseen environments} and (ii) ~\emph{produce incorrect explanations}.

\noindent\textbf{Our approach.}~
Based on the several experimental pieces of evidence presented in the following sections, this paper proposes \ourapproach--a methodology that enables reasoning about configurable system performance with causal inference and counterfactual reasoning. \ourapproach first recovers the underlying causal structure from performance data. The causal performance model 
allows users to (i)~identify the root causes of performance faults, (ii)~estimate the causal effects of various configurable parameters on the performance objectives, and (iii)~prescribe candidate configurations to fix the performance fault or optimize system performance. 
\begin{figure}[tp!]
    \centering
    \includegraphics*[width=\linewidth]{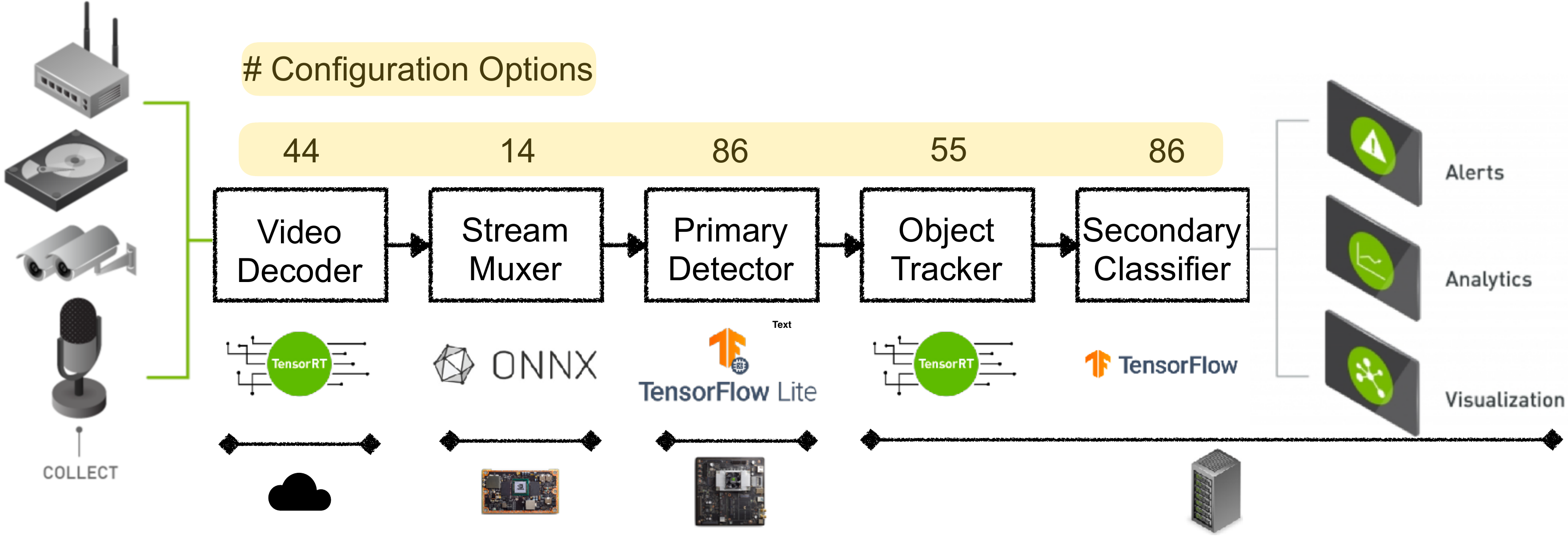}
    \caption{\small{\textsc{Deepstream}: An example of a highly-configurable composed system, a big data analytics pipeline system, with several configurable components: (i) Video Decoder performs video encoding/decoding with different formats; (ii) Stream Muxer accepts input streams and converts them to sequential batch frames; (iii) Primary Detector transforms the input frames based on input NN requirements and makes model inference to detect objects; (iv) Object Tracker supports multi-object tracking; (v) Secondary Classifier improves performance by avoiding re-inferencing.}}
    \label{fig:deepstream_pipeline}
    \vspace{-4mm}
\end{figure}

\noindent\textbf{Contributions.}~Our contributions are as follows:
\begin{itemize}
    \item We propose \ourapproach (\S\ref{sec:methodology}), a novel approach that allows causal reasoning about system performance.
    \item We \emph{have conducted a thorough evaluation} of \ourapproach in a controlled case study (\S\ref{sec:casestudy}) as well as real-world large-scale experiments. In particular, we evaluated \emph{effectiveness} (\S\ref{sec:effectiveness}), \emph{transferability} (\S\ref{sec:transfer}), and \emph{scalability} (\S\ref{sec:scalability}) by comparing \ourapproach with: (i)~state-of-the-art performance debugging approaches, including \textsc{CBI}~\cite{song2014statistical}, \textsc{DD}~\cite{artho2011iterative}, \textsc{EnCore}~\cite{zhang2014encore}, and \textsc{BugDoc}~\cite{lourencco2020bugdoc} and (ii)~performance optimization techniques, including \textsc{SMAC}~\cite{hutter2011sequential} and \textsc{PESMO} \cite{hernandez2016predictive}. The evaluations were conducted on six real-world highly configurable systems, including a video analytic pipeline, \textsc{Deepstream}~\cite{DeepStream}, three deep learning-based systems, \textsc{Xception} \cite{chollet2017xception}, \textsc{Deepspeech}~\cite{hannun2014deep}, and \textsc{BERT}~\cite{devlin2018bert}, a video encoder, X264~\cite{x264}, and a database engine, SQLite~\cite{SQLite}, deployed on \textsc{NVIDIA Jetson} hardware (\txone, \txtwo, and \xavier). \item In addition to sample efficiency and accuracy of \ourapproach in finding root causes of performance issues, we show that the learned causal performance model is transferable across different workload and deployment environments. Finally, we demonstrate the scalability of \ourapproach to large systems consisting of 500 options and several trillion potential configurations.  
    \item The artifacts and supplementary materials can be found at \href{github}{\color{blue!80} https://github.com/softsys4ai/unicorn}.
\end{itemize}

\section{Motivating Scenarios}
\label{sec:motivation}
\begin{figure}[tp!]

 \subfloat[Performance Distribution]{\raisebox{-7.5em}{\includegraphics*[width=0.65\linewidth, valign=b]{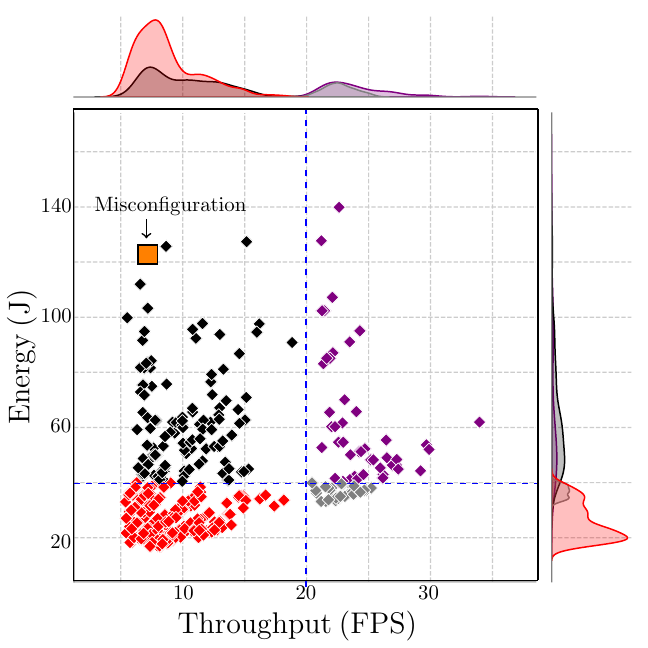}}\label{fig:perf_behavior}} 
 \subfloat[Misconfiguration]{
        \resizebox{0.32\linewidth}{!}{\adjustbox{valign=m}{%
       \begin{tabular}{lV{2.5}rV{2.5}}
    \clineB{1-2}{2.5}
    \multicolumn{1}{V{2.5}lV{2.5}}{\textbf{Config. Option}}& {Value} \\ 
     \clineB{1-2}{2.5}
    \multicolumn{1}{l}{} & \multicolumn{1}{l}{} \\[-0.3em] \hlineB{2.5}
     
    \multicolumn{1}{V{2.5}lV{2.5}}{CPU Cores} & 2 \\

    \multicolumn{1}{V{2.5}lV{2.5}}{CPU Frequency (GHz)} & 0.1 \\

    \multicolumn{1}{V{2.5}lV{2.5}}{EMC Frequency (GHz)} & 1.6\\

    \multicolumn{1}{V{2.5}lV{2.5}}{GPU Frequency (GHz)} & 0.7\\

    \multicolumn{1}{V{2.5}lV{2.5}}{Scheduler Policy} & \textsc{noop}\\

    \multicolumn{1}{V{2.5}lV{2.5}}{Scheduler Runtime ($\mu$s)} &  950k \\

    \multicolumn{1}{V{2.5}lV{2.5}}{Scheduler Child Runs First} & 1 \\
    
     \multicolumn{1}{V{2.5}lV{2.5}}{Batched Push Timeout} & 1 \\
     
      \multicolumn{1}{V{2.5}lV{2.5}}{Batch Size} & 1 \\
      
      \multicolumn{1}{V{2.5}lV{2.5}}{Interval} & 1 \\
      
      \multicolumn{1}{V{2.5}lV{2.5}}{Buffer Size} & 6000 \\

    \multicolumn{1}{V{2.5}lV{2.5}}{ Dirty Background Ratio} & 5 \\

    \multicolumn{1}{V{2.5}lV{2.5}}{Dirty Ratio} &  10  \\

    \multicolumn{1}{V{2.5}lV{2.5}}{Drop Caches}  &0  \\

    \multicolumn{1}{V{2.5}lV{2.5}}{Cache Pressure} & 500 \\
    
    \multicolumn{1}{V{2.5}lV{2.5}}{Swappiness} & 60  \\

    \multicolumn{1}{V{2.5}lV{2.5}}{Enable Padding} & 1 \\

    \multicolumn{1}{V{2.5}lV{2.5}}{Offset} & 0 \\
    
     \multicolumn{1}{V{2.5}lV{2.5}}{Swap Memory (Gb)} & 1 \\
     
      \multicolumn{1}{V{2.5}lV{2.5}}{Sched Time Avg (ms)} & 1000 \\
      
       \multicolumn{1}{V{2.5}lV{2.5}}{Dirty Bytes} & 30 \\
       
        \multicolumn{1}{V{2.5}lV{2.5}}{Overcommit Memory} & 2 \\
        
         \multicolumn{1}{V{2.5}lV{2.5}}{Overcommit Hugepages} & 1 \\

    \hlineB{2.5}

    \multicolumn{1}{l}{} & \multicolumn{1}{l}{}  \\[-0.9em] 
    \clineB{1-2}{2.5}
     
    \multicolumn{1}{V{2.5}lV{2.5}}{Throughput (FPS)} & 6.8 \\
       
    \multicolumn{1}{V{2.5}lV{2.5}}{Energy (J)} & 122 \\
     \clineB{1-2}{2.5}
        \end{tabular}}}
        \label{tab:example_fault}}

\caption{\small {(a) Performance distribution when \textsc{Deepstream} is deployed on NVIDIA Jetson Xavier (b) Misconfiguration that caused the multi-objective non-functional fault, shown as {\color{orange!60} $\square$} in the performance distribution.}}
\vspace{-3mm}
\label{tab:sample_fault}\end{figure}

\noindent
\textbf{Simple motivating scenario.} In this simple scenario, we motivate our work by demonstrating why performance analyses solely based on correlational statistics may lead to an incorrect outcome. Here, the collected performance data indicates that \texttt{Throughput} is positively correlated with increased \texttt{Cache Misses}\footnote{we used a \texttt{distinct font} to indicate variables such as configuration options or performance metrics and events throughout this paper.} (as in  \fig{intro_fig}(a)). A simple ML model built on this data will predict with high confidence that larger \texttt{Cache Misses} leads to higher \texttt{Throughput}---this is misleading as higher \texttt{Cache Misses} should, in theory, lower \texttt{Throughput}. By further investigating the performance data, we noticed that the caching policy was automatically changed during measurement. We then segregated the same data on \texttt{Cache Policy} (as in~ \fig{intro_fig}(b)) and found out that within each group of \texttt{Cache Misses}, as \texttt{Cache Misses} increases, the \texttt{Throughput} decreases. One would expect such behavior, as the more \texttt{Cache Misses} the higher number of access to external memory, and, therefore, the \texttt{Throughput} would be expected to decrease. The system resource manager may change the \texttt{Cache Policy} based on some criteria; this means that for the same number of \texttt{Cache Misses}, the \texttt{Throughput} may be lower or higher; however, in all \texttt{Cache Policies}, the increases of \texttt{Cache Misses} resulting in a decrease in \texttt{Throughput}. Thus, \texttt{Cache Policy} acts as a confounder that explains the relation between \texttt{Cache Misses} and \texttt{Throughout}, which a correlation-based model will not be able to capture. In contrast, a causal performance model, as shown in \fig{intro_fig}(c), finds the relation between \texttt{Cache Misses}, \texttt{Cache Policy}, and \texttt{Throughput} and thus can reason about the observed behavior correctly. 

In reality, performance analysis and debugging of heterogeneous multi-component systems is non-trivial and often compared with finding the needle in the haystack~\cite{whitaker2004configuration}. In particular, the end-to-end performance analysis is not possible by reasoning about individual components in isolation, as components may interact with one another in such a composed system. Below, we use a highly configurable multi-stack system to motivate why causal reasoning is a better choice for understanding the performance behavior of complex systems. 

\noindent
\textbf{Motivating scenario based on a highly configurable data analytics system.} We deployed a data analytics pipeline, \textsc{Deepstream}~\cite{DeepStream}. \textsc{Deepstream} has many components, and each component has many configuration options, resulting in several variants of the same system as shown in~\fig{deepstream_pipeline}. Specifically, the variability arises from: 
(i)~the configuration options of each software component in the pipeline, (ii)~configurable low-level libraries that implement functionalities required by different components (e.g., the choice of tracking algorithm in the tracker or different neural network architectures), (iii)~the configuration options associated with each component's deployment stack (e.g., \texttt{CPU Frequency} of \xavier).  Further, there exist many configurable events that can be measured/observed at the OS level by the event tracing system. More specifically, the configuration space of the system includes (i) 27 Software options (Decoder: 6, Stream Muxer: 7, Detector: 10, Tracker: 4), (ii) 22 Kernel options (e.g., \texttt{Swappiness}, \texttt{Scheduler Policy}, etc.), and (iii) 4 Hardware options (\texttt{CPU Frequency}, \texttt{CPU Cores}, etc.). We use 8 camera streams as the workload, x264 as the decoder, TrafficCamNet model that uses ResNet 18 architecture for the detector, and an NvDCF tracker, which uses a correlation filter-based online discriminative learning algorithm for tracking. Such a large space of variability makes performance analysis challenging. This is further exacerbated by the fact that the configuration options among the components \emph{interact} with each other. Additional details about our \textsc{Deepstream} implementation can be found in the \href{https://github.com/softsys4ai/unicorn}{\color{blue!80}supplementary materials}.

To better understand the potential of the proposed approach, we measured (i) application performance metrics including throughput and energy consumption by instrumenting the \textsc{Deepstream} code, and (ii) 288 system-wide performance events (hardware, software, cache, and tracepoint) using $perf$ and measured performance for 2461 configurations of \textsc{Deepstream} in two different hardware environments, \xavier, and \txtwo.  As it is depicted in  \fig{perf_behavior}, performance behavior of \textsc{Deepstream}, like other highly configurable systems, is non-linear, multi-modal, and non-convex~\cite{jamshidi2016uncertainty}. In this work, we focus on two performance tasks: 
(i) \textit{Performance Debugging}: here, one observes a performance issue (e.g., latency), and the task involves replacing the current configurations in the deployed environment with another that fixes the observed performance issue; 
(ii) \textit{Performance Optimization}: here, no performance issue is observed; however, one wants to get a near-optimal performance by finding a configuration that enables the best trade-off in the multi-objective space (e.g., throughput vs. energy consumption vs. accuracy in \textsc{Deepstream}).

\begin{figure}[tp!]
\small
    \centering
    \includegraphics*[width=\linewidth]{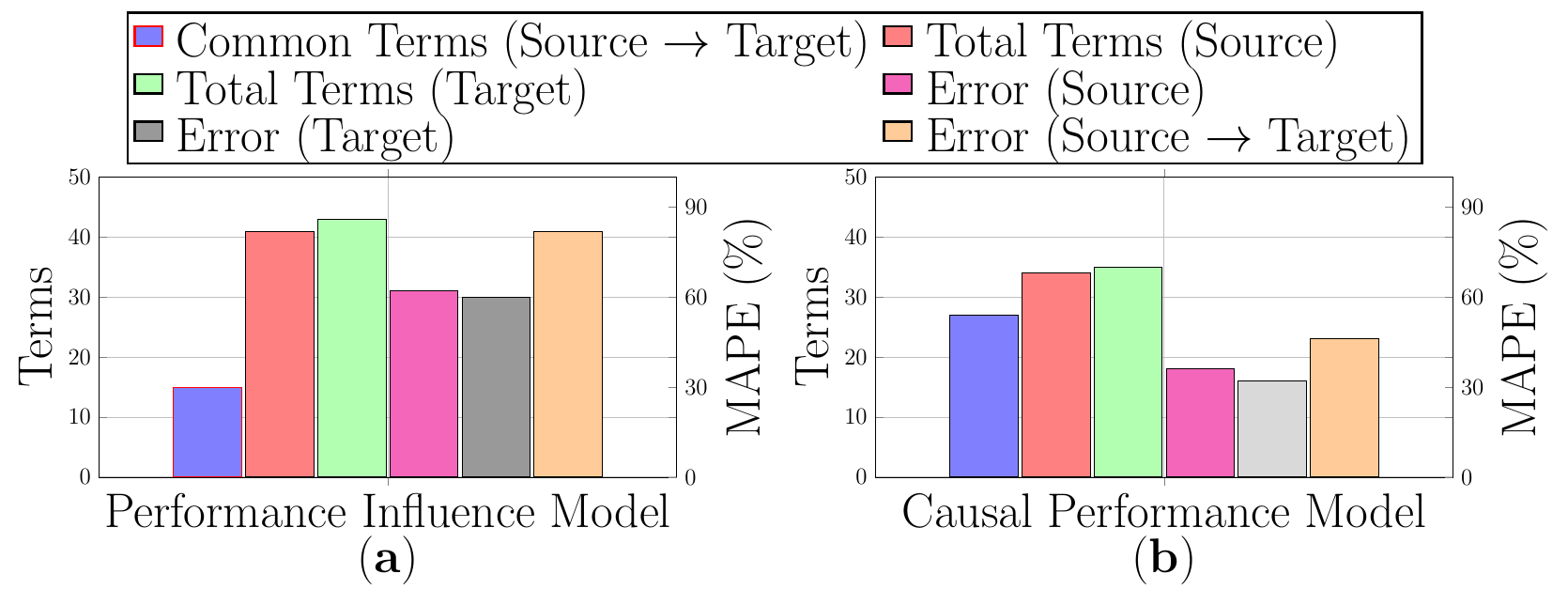}
    \caption{\small {(a) Performance influence models do not generalize well as the number of common terms, total terms and prediction error of these models change from source (\xavier) to target (\txtwo). The rank correlation between source and target is 0.07 (p-value=0.73).} (b) \small {Causal performance models generalize better as the number of common terms, total terms and prediction error of the structural does not change much from source (\xavier) to target (\txtwo). The rank correlation between source and target is 0.49 (p-value=0.76).}}
    \label{fig:barplot_hw_change}
\end{figure}

\begin{figure}[tp!]
\small
    \centering
    \includegraphics*[width=\linewidth]{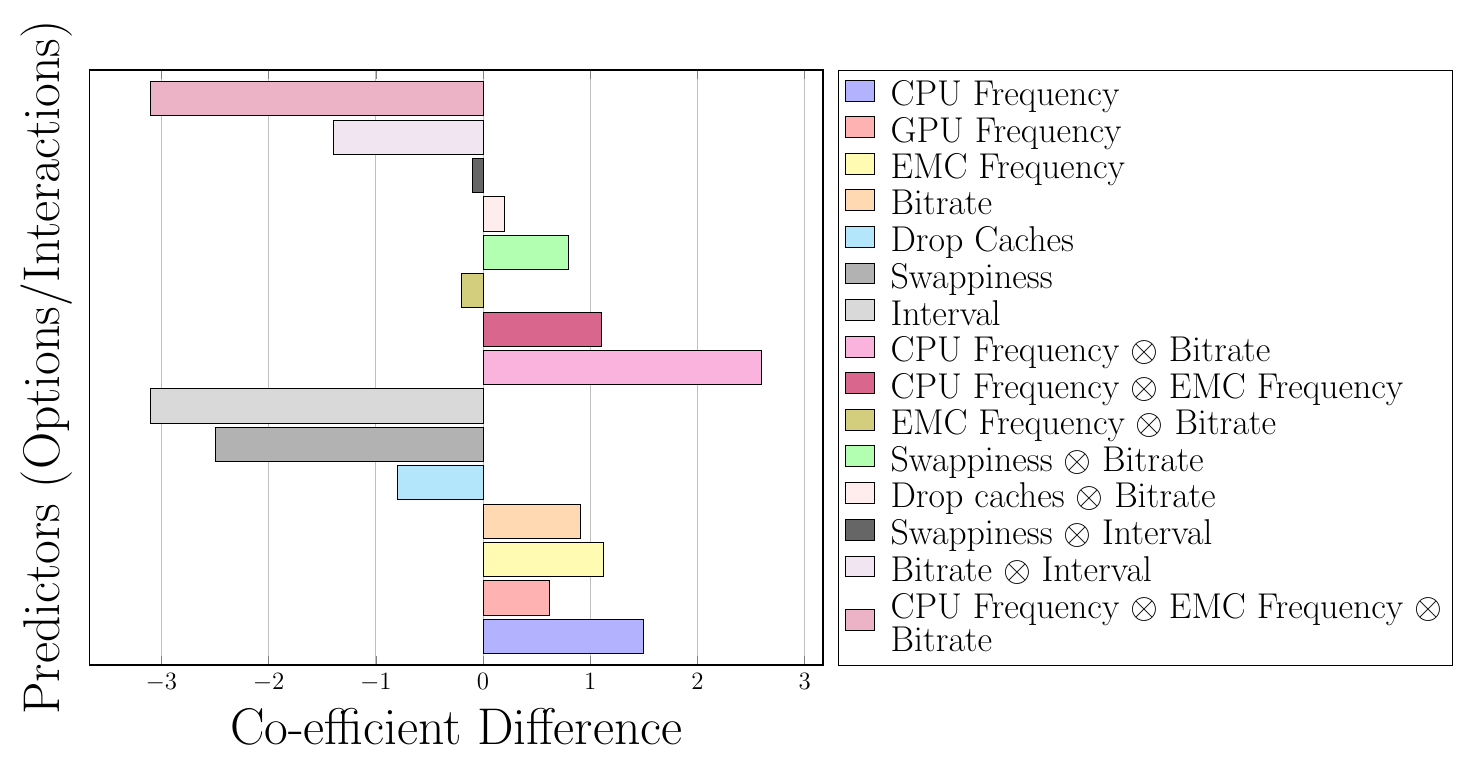}
    \caption{\small {Visualizing co-efficient differences from the source (\xavier) performance influence model to the target (\txtwo) performance influence model for the common terms for both options and interactions (shown by $\otimes$). }}
    \label{fig:coeff_across_environments}
    
\end{figure}

To show major shortcomings of existing state-of-the-art performance models, we built performance influence models that have extensively been used in the systems' literature~\cite{siegmund2015performance,JSVKPA:ASE17,GCASW:ASE13,GSKA:SPPEXA,kolesnikov2019tradeoffs,kaltenecker2020interplay,muhlbauer2019accurate,grebhahn2019predicting,siegmund2020dimensions} and it is the standard approach in industry~\cite{kolesnikov2019tradeoffs,kaltenecker2020interplay}. Specifically, we built non-linear regression models with forward and backward elimination using a step-wise training method on the \textsc{Deepstream} performance data. We then performed several sensitivity analyses and identified the following issues: 
\besq
    \item \textbf{Performance influence models could not reliably predict performance in unseen environments.} 
    Performance behavior of configurable systems vary across environments, e.g., when we deploy software on new hardware with a different microarchitecture or when the workload changes~\cite{JSVKPA:ASE17,JVKSK:SEAMS17,JVKS:FSE18,iqbal_transfer_2019,VPGFC:ICPE17}. When building a performance model, it is important to capture predictors that transfer well, i.e., remain \emph{stable} across environmental changes. The predictors in performance models are options ($o_i$) and interactions ($\phi(o_i..o_j)$) that appear in the explainable models of form $f(c)=\beta_0+\Sigma_{i}{\phi(o_i)}+\Sigma_{i..j}{\phi(o_i..o_j)}$. The transferability of performance predictors is expected from performance models since they are learned based on one environment (e.g., staging as the source environment) and are desirable to reliably predict performance in another environment (e.g., production as the target environment). 
    Therefore, if the predictors in a performance model become unstable, even if they produce accurate predictions in the current environment, there is no guarantee that it performs well in other environments, i.e., they become unreliable for performance predictions and performance optimizations due to large prediction errors. 
    To investigate how transferable performance influence models are across environments, we performed a thorough analysis when learning a performance model for \textsc{DeepStream} deployed on two different hardware platforms that have two different microarchitectures. Note that such environmental changes are common, and it is known that performance behavior changes when, in addition to a change in hardware resources (e.g., higher \texttt{CPU Frequency}), we have major differences in terms of architectural constructs~\cite{ding2021generalizable,curtsinger2013stabilizer}, also supported by a thorough empirical study~\cite{JSVKPA:ASE17}. The results in \fig{barplot_hw_change} (a) indicate that the number of stable predictors is too small for the total number of predictors that appear in the learned regression models. Additionally, the coefficients of the common predictors change across environments as shown in \fig{coeff_across_environments} making them unreliable to be resued in the new scenario.  
    
    

    \item \textbf{Performance influence models could produce incorrect explanations.} In addition to performance predictions, where developers are interested to know the effect of configuration changes on performance objectives, they are also interested to estimate and explain the effect of a change in particular \emph{configuration options} (e.g., changing \texttt{Cache Policy}) toward performance variations. It is therefore desirable that the strength of the predictors in performance models, determined by their coefficients, remain consistent across environments~\cite{ding2021generalizable,JSVKPA:ASE17}.  
    In the context of our simple scenario in \fig{intro_fig}, the learned performance influence model indicates that $0.16 \times \texttt{Cache Misses}$ is the most influential term that determines throughput, however, the (causal) model in \fig{intro_fig}(c) show that the interactions between configuration option \texttt{Cache Policy} and system event \texttt{Cache Misses} is a more reliable predictor of the throughput, indicating that the performance influence model, due to relying on superficial correlational statistics, incorrectly explains factors that influence performance behavior of the system. The low Spearman rank correlation between predictors coefficients indicates that a performance model based on regression could be highly unstable and thus would produce unreliable explanations as well as unreliable estimation of the effect of change in specific options for performance debugging or optimization.
\ee

\section{Causal Reasoning for Systems}
\label{sec:causalreasoning}

We hypothesize that the reason behind unreliable predictions and incorrect explanations of performance influence models (see \S\ref{sec:causalreasoning}) is the inability of correlation-based models to capture causally relevant predictors in the learned performance models. The theoretical and empirical results~\cite{JSVKPA:ASE17,javidian2019transfer} also show that predictive models that contain non-causal predictors, even though they might be accurate in the environment that the training data come from, such models are not typically transferable in unseen environments.

Hence, we introduce a new abstraction for performance modeling, called \emph{Causal Performance Model}, which gives us the leverage for performing causal reasoning for computer systems. In particular, we introduce the causal performance model to serve as a \emph{modeling abstraction} that allow building \emph{reusable performance models} for downstream performance tasks, including performance predictions, performance testing and debugging, performance optimization, and more importantly, it serves as a \emph{transferable model} that allow performance analyses across environments~\cite{JSVKPA:ASE17,javidian2019transfer}.

\begin{figure}[tp!]
    \centering
    \includegraphics*[width=\linewidth]{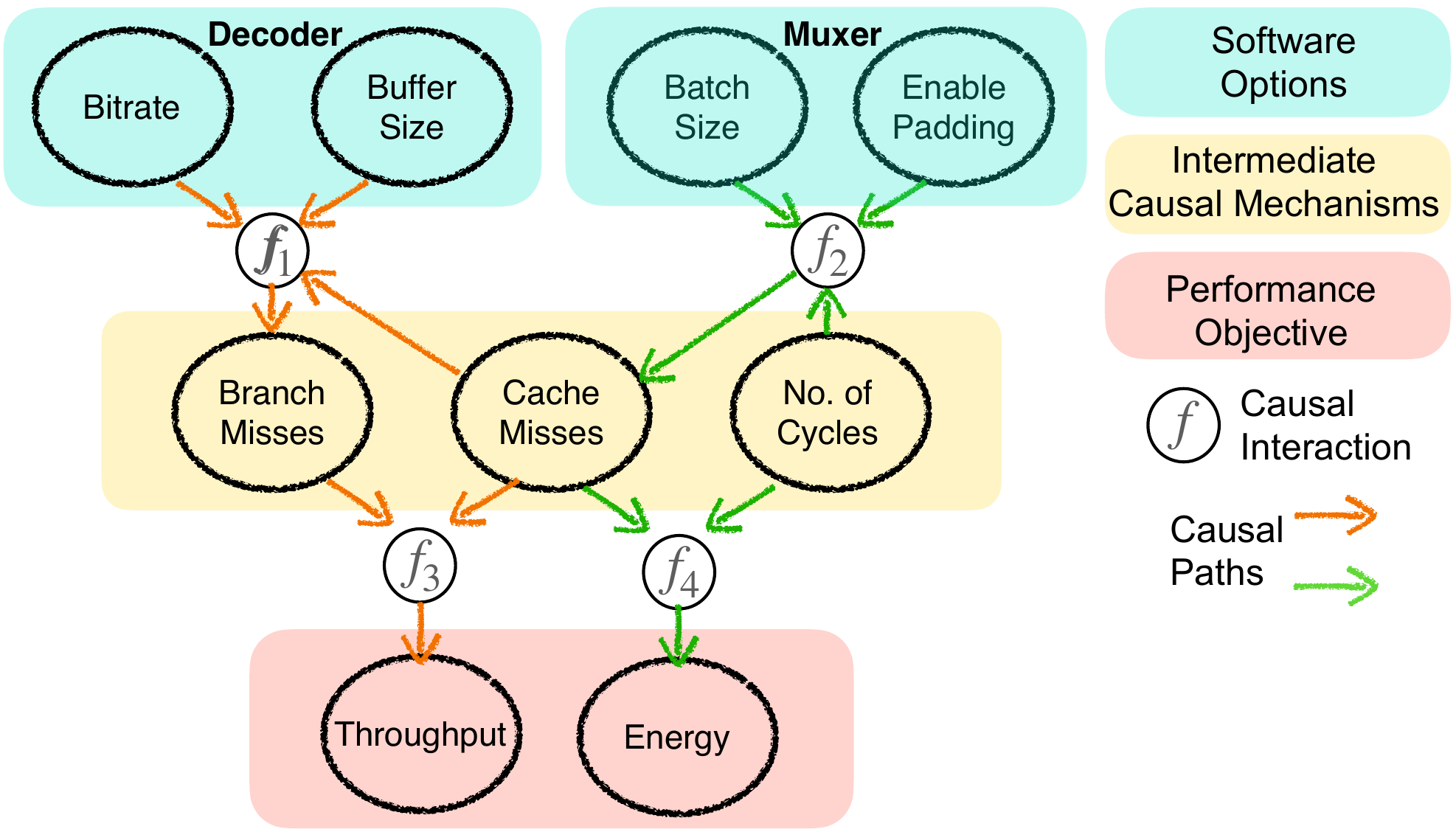}
    \caption{\small {A partial causal performance model for \textsc{Deepstream} discovered in our experiments. 
    }}
    
    \label{fig:causal_model_example}
    \vspace{-3mm}
\end{figure}

\textbf{Causal performance models}. We define a causal performance model as an instantiation of Probabilistic Graphical Models~\cite{pearl1998graphical} with new types and structural constraints to enable performance modeling and analyses. Formally, causal performance models (cf., \fig{causal_model_example}) are Directed Acyclic Graphs (DAGs)~\cite{pearl1998graphical} with (i) performance variables, (ii) functional nodes that define functional dependencies between performance variables (i.e., how variations in one or multiple variables determine variations in other variables), (iii) causal links that interconnect performance nodes with each other via functional nodes, and (iv) constraints to define assumptions we require in performance modeling (e.g., software configuration options cannot be the child node of performance objectives; or \texttt{Cache Misses} as a performance variable takes only positive integer values). In particular, we define three new variable types: (i) Software-level configuration options associated with a software component in the composed system (e.g., \texttt{Bitrate} in the decoder component of \textsc{Deepstream}), and hardware-level options (e.g., \texttt{CPU Frequency}), (ii) intermediate performance variables relating the effect of configuration options to performance objectives including middleware traces (e.g., \texttt{Context Switches}), performance events (e.g., \texttt{Cache Misses}) and (iii) end-to-end performance objectives (e.g., \texttt{Throughput}). In this paper, we characterize the functional nodes with polynomial models, because of their simplicity and their explainable nature, however, they could be characterized with any functional forms, e.g., neural networks~\cite{xia2021causal,scherrer2021learning}. We also define two specific constraints over causal performance models to characterize the assumptions in performance modeling: (i) defining variables that can be intervened (note that some performance variables can only be observed (e.g., \texttt{Cache Misses}) or in some cases where a variable can be intervened, the user may want to restrict the variability space, e.g., the cases where the user may want to use prior experience, restricting the variables that do not have a major impact to performance objectives); (ii) structural constraints, e.g., configuration options do not cause other options. Note that such constraints enable incorporating domain knowledge and enable further sparsity that facilitates learning with low sample sizes.

\textbf{How causal reasoning can fix the reliability and explainability issues in current performance analyses practices?}.
The causal performance models contain more detail than the joint distribution of all variables in the model. For example, the causal performance model in \fig{causal_model_example} encodes not only \texttt{Branch Misses} and \texttt{Throughput} readings are dependent but also that lowering \texttt{Cache Misses} causes the \texttt{Throughput} of \textsc{Deepstream} to increase and not the other way around. The arrows in causal performance models correspond to the assumed direction of causation, and the absence of an arrow represents the absence of direct causal influence between variables, including configuration options, system events, and performance objectives. The only way we can make predictions about how performance distribution changes for a system when deployed in another environment or when its workload changes are if we know how the variables are causally related. This information about causal relationships is not captured in non-causal models, such as regression-based models. 
Using the encoded information in causal performance models, we can benefit from analyses that are only possible when we explicitly employ causal models, in particular, interventional and counterfactual analyses~\cite{pearl2009causality, pearl2018book}. For example, imagine that in a hardware platform, we deploy the \textsc{Deepstream} and observed that the system throughput is below 30 \texttt{FPS} and \texttt{Buffer Size} as one of the configuration options was determined dynamically between 8k-20k. The system maintainers may be interested in estimating the likelihood of fixing the performance issue in a counterfactual world where the \texttt{Buffer Size} is set to a fixed value, 6k. The estimation of this counterfactual query is only possible if we have access to the underlying causal model because setting a specific option to a fixed value is an intervention as opposed to conditional observations that have been done in the traditional performance model for performance predictions.

Causal performance models are not only capable of predicting system performance in certain environments, they encode the causal structure of the underlying system performance behavior, i.e., the data-generating mechanism behind system performance. Therefore, the causal model can reliably transfer across environments~\cite{scholkopf2021toward}. To demonstrate this for causal performance models as a particular characterization of causal models, we performed a similar sensitivity analysis to regression-based models and observed that causal performance models can reliably predict performance in unseen environments (see \fig{barplot_hw_change} (b)). In addition, as opposed to performance influence models that are only capable of performance predictions, causal performance models can be used for several downstream heterogeneous performance tasks. For example, using a causal performance model, we can determine the \emph{causal effects} of configuration options on performance objectives. Using the estimated causal effects, one can determine the effect of change in a particular set of options towards performance objectives and therefore can select the options with the highest effects to fix a performance issue, i.e., bring back the performance objective that has violated a specific quality of service constraint without sacrificing other objectives. Causal performance models are also capable of predicting performance behavior by calculating conditional expectation, $E(Y|X)$, where $Y$ indicates performance objectives, e.g., throughput, and $X=x$ is the system configurations that have not been measured.

\section{\ourapproach}
\label{sec:methodology}

This section presents \ourapproach--our methodology for performance analyses of highly configurable and composable systems with causal reasoning. 

\smallskip
\begin{figure}[tp!]
    \centering
    \includegraphics*[width=\linewidth]{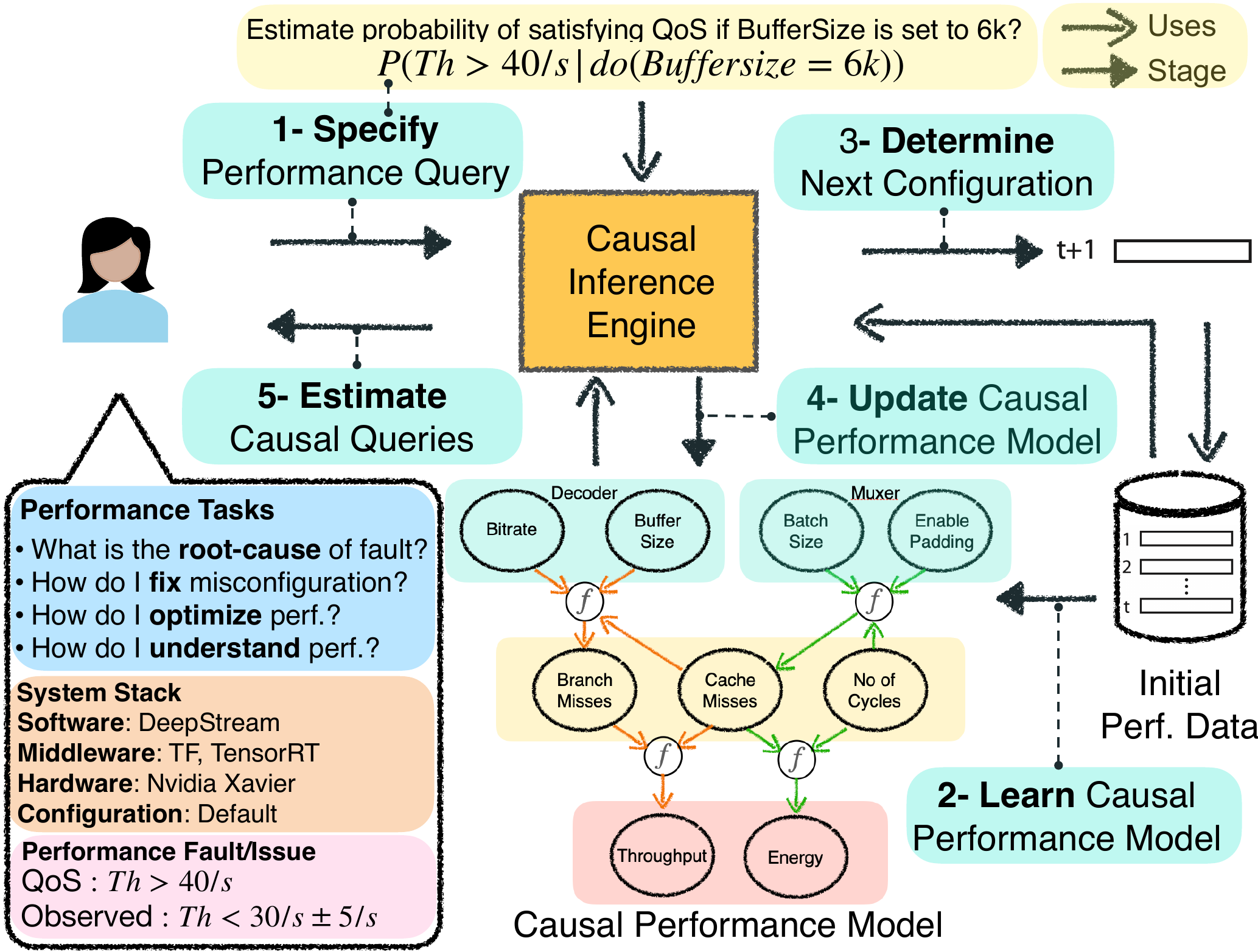}
    \caption{\small {Overview of \ourapproach}.}
    \label{fig:overview}
\end{figure}

\begin{figure}[tp!]
    \centering
    \includegraphics*[width=\linewidth]{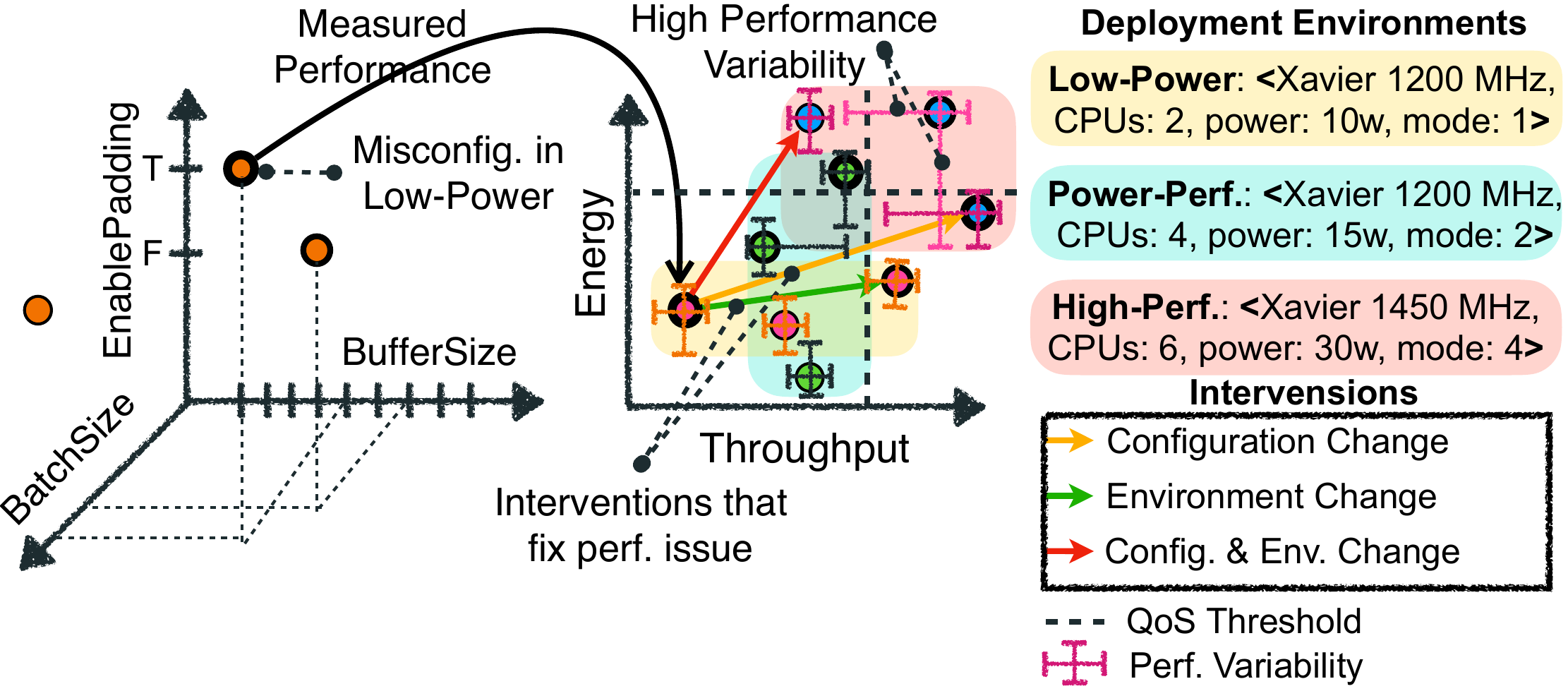}
    \caption{\small {Mapping configuration space to multi-objective performance space.}}
    
    \label{fig:config_to_objective_space}
    \vspace{-4mm}
\end{figure}
\paragraph{Overview.}
\ourapproach works in five stages, implementing an active learning loop (cf. \fig{overview}): (i) Users or developers of a highly-configurable system \emph{specify}, in a human-readable language, the performance task at hand in terms of a query in the Inference Engine. For example, a \textsc{Deepstream} user may have experienced a throughput drop when they have deployed it on NVIDIA Xavier in low-power mode (cf. \fig{config_to_objective_space}). Then, \ourapproach's main process starts by (ii) collecting some predetermined number of samples and \emph{learning a causal performance model}; Here, a sample contains a system configuration and its corresponding measurement—including low-level system events and end-to-end system performance. Given a certain budget, which in practice either translates to time~\cite{iqbal2020flexibo} or several samples~\cite{jamshidi2016uncertainty}, \ourapproach, at each iteration, (iii) \emph{determines the next configuration(s)} and measures system performance when deployed with the determined configuration--i.e. new sample; accordingly, (iv) the \emph{learned causal performance model is incrementally updated}, reflecting a model that captures the underlying causal structure of the system performance. \ourapproach terminates if either budget is exhausted or the same configuration has been selected a certain number of times consecutively, otherwise, it continues from Stage III. Finally, (v) to automatically derive the quantities which are needed to conduct the performance tasks, the specified performance queries are \emph{translated} to formal causal queries, and they will be \emph{estimated} based on the final causal model.


\begin{figure}[tp!]
    \centering
    \includegraphics*[width=\linewidth]{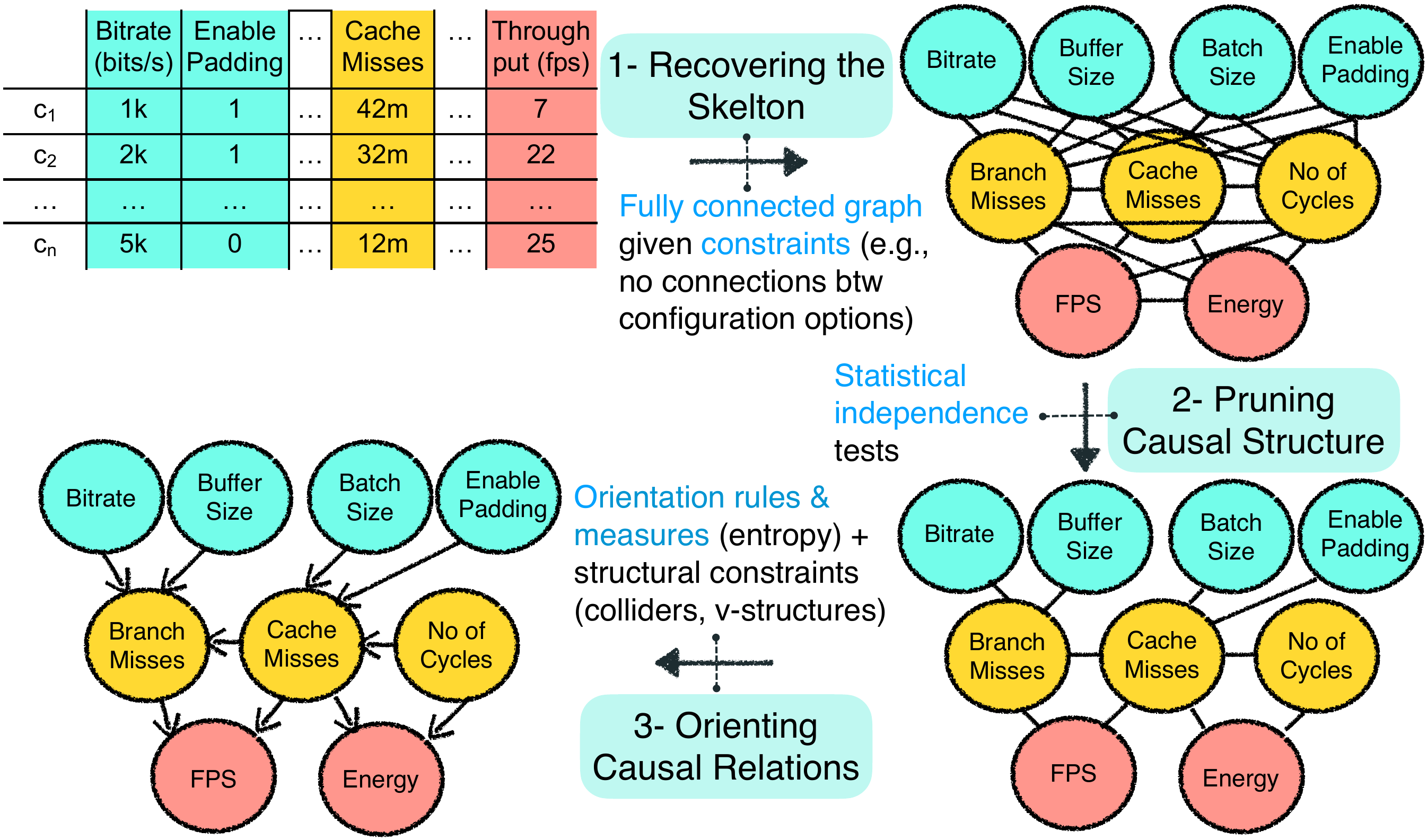}
    \caption{\small {Causal model learning from performance data.}}
    
    \label{fig:causal_model_learning}
\end{figure}

\smallskip
\noindent
\paragraph{Stage I: Formulate Performance Queries.}
\ourapproach enables ~\textit{developers} and ~\textit{users} of highly-configurable systems to conduct performance tasks, including performance debugging, optimization, and tuning, n particular, when they need to answer several performance queries:
(i) What configuration options \emph{caused} the performance fault? (ii) What are \emph{important options and their interactions} that influence performance? (iii) How to \emph{optimize} one quality or navigate \textit{tradeoffs} among multiple qualities in a reliable and explainable fashion? (iv) How can we \emph{understand} what options and possible interactions are most responsible for the performance degradation in production? 

\begin{figure}[tp!]
    \centering
    \includegraphics*[width=\linewidth]{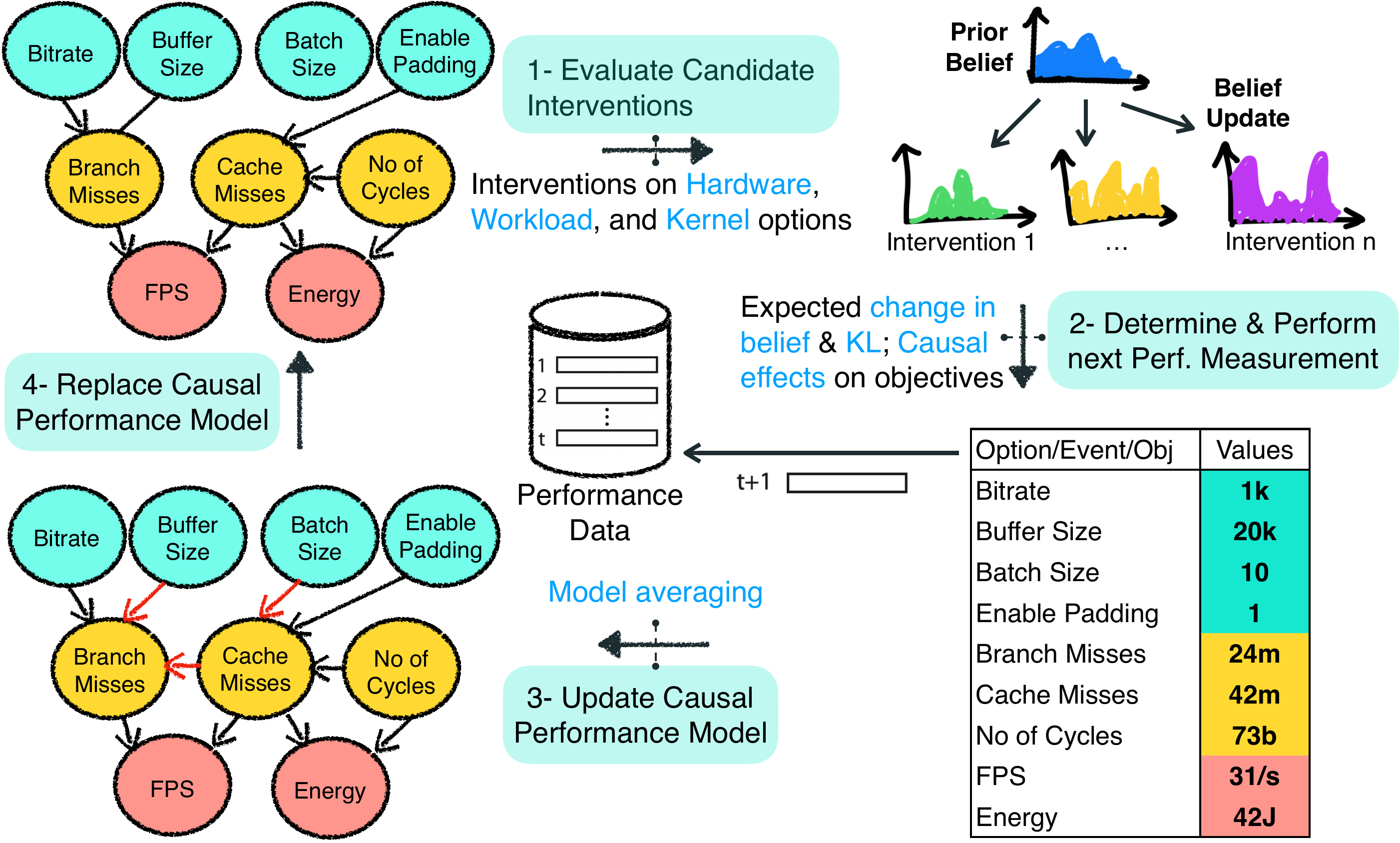}
    \caption{\small {Causal model update.}}
    
    \label{fig:causal_model_update}
\end{figure}

At this stage, the performance queries are translated to formal causal queries using the interface of the causal inference engine (cf. \fig{overview}). Note that in the current implementation of \ourapproach, this translation is performed manually, however, this process could be made automated by creating a grammar for specifying performance queries and the translations can be made between the performance queries into the well-defined causal queries, note that such translation has been done in domains such as genomics~\cite{farahmand2019causal}.  

\paragraph{Stage II: Learn Causal Performance Model.}
\label{sect:structure_discovery}
In this stage, \ourapproach learns a causal performance model (see Section \ref{sec:motivation}) that explains the causal relations between configuration options, the intermediate causal mechanism, and performance objectives. Here, we use an existing structure learning algorithm  called \textit{Fast Causal Inference} (hereafter, FCI)~\cite{spirtes2000causation}. We selected FCI because: (i) it accommodates for the existence of unobserved confounders~\cite{spirtes2000causation,ogarrio2016hybrid, glymour2019review}, \ie, it operates even when there are latent common causes that have not been, or cannot be, measured. This is important because we do not assume absolute knowledge about configuration space, hence there could be certain configurations we could not modify or system events we have not observed. (ii) FCI, also, accommodates variables that belong to various data types such as nominal, ordinal, and categorical data common across the system stack (cf. \fig{config_to_objective_space}).
To build the causal performance model, we, first, gather a set of initial samples (cf.~\fig{causal_model_learning}). To ensure reliability~\cite{curtsinger2013stabilizer,ding2021generalizable}, we measure each configuration multiple times, and we use the median (as an unbiased measure) for the causal model learning. As depicted in \fig{causal_model_learning}, \ourapproach implements three steps for causal structure learning: (i) recovering the skeleton of the causal performance model by enforcing structural constraints; (ii) pruning the recovered structure using standard statistical tests of independence. In particular, we use mutual info for discrete variables and Fisher z-test for continuous variables; (iii) orienting undirected edges using entropy~\cite{spirtes2000causation,ogarrio2016hybrid, glymour2019review,colombo2012learning,colombo2014order}.

\noindent \textbf{Orienting undirected causal links.} We orient undirected edges using prescribed edge orientation rules~\cite{spirtes2000causation,ogarrio2016hybrid, glymour2019review,colombo2012learning,colombo2014order} to produce a \textit{partial ancestral graph} (or PAG). A PAG contains the following types of (partially) directed edges: 
\bi[leftmargin=*, topsep=0pt]
\item $X$\edgeone$Y$ indicating that vertex $X$ causes $Y$. 
\item $X$\edgetwo$Y$ which indicates that there are unmeasured confounders between vertices $X$ and $Y$.
\ei
\noindent In addition, a PAG produces two types of edges:
\bi[leftmargin=*, topsep=0pt]
\item $X$\edgethree$Y$ indicating that either $X$ causes $Y$, or that there are \textit{unmeasured confounders} that cause both $X$ and $Y$.
\item $X$\edgefour~$Y$ which indicates that either: (a) vertices $X$ causes $Y$, or (b) vertex $Y$ causes $X$, or (c) there are \textit{unmeasured confounders} that cause both $X$ and $Y$.
\ei
\begin{figure}[tp!]
    \centering
    \includegraphics*[width=\linewidth]{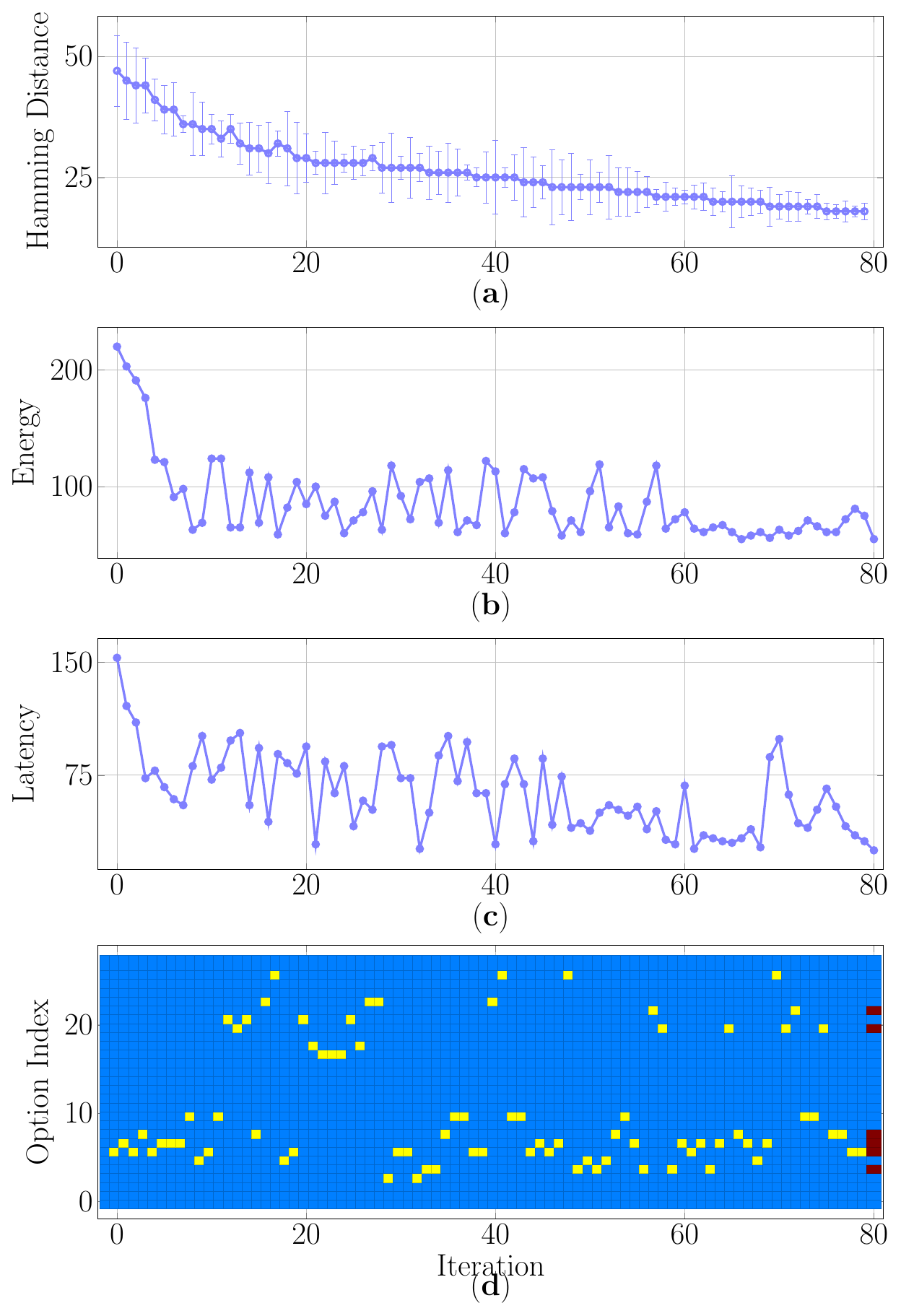}
    \caption{\small {(a) The hamming distance between the learned causal model and ground truth model decreases as the algorithms measure more configuration samples. Incremental update of (b) Latency and (c) Energy, using \ourapproach for debugging a multi-objective fault. Configuration options selected by \ourapproach at each iteration are during debugging are shown in (d) using the yellow-colored nodes. Red-colored nodes indicate configuration options that are selected as a fix to the multi-objective performance fault. Mapping between option indexes and configuration options are shown in the \href{https://github.com/softsys4ai/unicorn}{\color{blue!80}supplementary materials}.}}
   
    \label{fig:incremental_update}
\end{figure}
\noindent In the last two cases, the circle ($\circ$) indicates that there is an ambiguity in the edge type. In other words, given the current observational data, the circle can indicate an arrowhead (\edgeone) or no arrowhead (---), \ie, for $X$\edgefour$Y$, all three of $X$\edgeone$Y$, $Y$\edgeone$X$, and $X$\edgetwo$Y$ might be compatible with current data, \ie, the current data could be faithful to each of these statistically equivalent causal graphs inducing the same conditional independence relationships.

\noindent{\textbf{Resolving partially directed edges.}}~
For subsequent analyses over the causal graph, the PAG obtained must be fully resolved (directed with no $\circ$ ended edges) in order to generate an ADMG. We use the information-theoretic approach using entropy proposed in \cite{Kocaoglu2017,Kocaoglu2020} to discover the true causal direction between two variables. Our work extends the theoretic underpinnings of entropic causal discovery to generate a fully directed causal graph by resolving the partially directed edges produced by FCI. For each partially directed edge, we follow two steps: (i) establish if we can generate a latent variable (with low entropy) to serve as a common cause between two vertices; (ii) if such a latent variable does not exist, then pick the direction which has the lowest entropy. 

For the first step, we assess if there could be an unmeasured confounder (say $Z$) that lies between two partially oriented nodes (say $X$ and $Y$). For this, we use the \textit{LatentSearch} algorithm proposed by Kocaoglu \etal~\cite{Kocaoglu2020}. \textit{LatentSearch} outputs a joint distribution $q(X, Y, Z)$ of the variables $X$, $Y$, and $Z$ which can be used to compute the entropy $H(Z)$ of the unmeasured confounder $Z$. Following the guidelines of Kocaoglu \etal, we set an entropy threshold $\theta_r=0.8 \times min\left\{H(X), H(Y)\right\}$. If the entropy $H(Z)$ of the unmeasured confounder falls \textit{below} this threshold, then we declare that there is a simple unmeasured confounder $Z$ (with a low enough entropy) to serve as a common cause between $X$ and $Y$ and accordingly, we replace the partial edge with a bidirected (\ie, \edgetwo) edge. 

When there is no latent variable with a sufficiently low entropy, two possibilities exist: {(i)} variable $X$ causes $Y$; then, there is an arbitrary function $f(\cdot)$ such that $Y=f(X,E)$, where $E$ is an exogenous variable (independent of $X$) that accounts for system noise; or {(ii)} variable $Y$ causes $X$; then, there is an arbitrary function $g(\cdot)$ such that $X=g(Y,\tilde{E})$, where $\tilde{E}$ is an exogenous variable (independent of $Y$) that accounts for noise in the system. The distribution of $E$ and $\tilde{E}$ can be inferred from the data~\cite[see~\S3.1]{Kocaoglu2017}. With these distributions, we measure the entropies $H(E)$ and $H(\tilde{E})$. If $H(E) < H(\tilde{E})$, then, it is simpler to explain the $X$\edgeone$Y$ (\ie, the entropy is lower when $Y=f(X,E)$) and we choose $X$\edgeone$Y$. Otherwise, we choose $Y$\edgeone$X$. 

\begin{figure}[tp!]
    \setlength{\belowcaptionskip}{-1em}
    \centering
    \resizebox{\linewidth}{!}{
        \begin{tabular}{V{2.5}p{\linewidth}V{2.5}}
            \hlineB{2}
            \small
            \textbf{Problem~\cite{code_transplant:online}:}~For a real-time scene detection task, \txtwo (faster platform) only processed 4 frames/sec whereas \txone (slower platform) processed 17 frames/sec, \ie, the latency is $4\times$ worse on \txtwo.
            \\
            \small\textbf{Observed Latency (frames/sec):} 4 FPS\\
            \small\textbf{Expected Latency (frames/sec):} 22-24 FPS \textit{(30-40\% better)}\\\hlineB{2}
        \end{tabular}
    }\vspace{0.1em}
    \resizebox{\linewidth}{!}{%
    \begin{tabular}{lV{2.5}ccccV{2.5}r}
    \clineB{2-6}{2.5}
     \multicolumn{1}{lV{2.5}}{\textbf{Configuration Options}}& \rotatebox{90}{\tool~} & \rotatebox{90}{SMAC} & \rotatebox{90}{\bugdoc~} & \rotatebox{90}{Forum} & \multicolumn{1}{cV{2.5}}{\rotatebox{90}{ACE$^\dagger$}} \bigstrut[t]\\ \clineB{2-6}{2.5}
     
    \multicolumn{1}{l}{} & \multicolumn{1}{l}{} & \multicolumn{1}{l}{} & \multicolumn{1}{l}{} & \multicolumn{1}{l}{} & \multicolumn{1}{l}{} \\[-0.9em] \hlineB{2.5}
     
    \multicolumn{1}{V{2.5}lV{2.5}}{\texttt{CPU Cores}} & \cellcolor{blue!10}\color{gray50}{\faCheck} & \cellcolor{blue!10}\color{gray50}{\faCheck} & \cellcolor{blue!10}\color{gray50}{\faCheck} & \cellcolor{blue!10}\color{gray50}{\faCheck} & \multicolumn{1}{cV{2.5}}{3\%} \bigstrut[t]\\

    \multicolumn{1}{V{2.5}lV{2.5}}{\texttt{CPU Frequency}} & \cellcolor{blue!10}\color{gray50}{\faCheck} & \cellcolor{blue!10}\color{gray50}{\faCheck} & \cellcolor{blue!10}\color{gray50}{\faCheck} & \cellcolor{blue!10}\color{gray50}{\faCheck} & \multicolumn{1}{cV{2.5}}{6\%} \\

    \multicolumn{1}{V{2.5}lV{2.5}}{\texttt{EMC Frequency}} & \cellcolor{blue!10}\color{gray50}{\faCheck} & \cellcolor{blue!10}\color{gray50}{\faCheck} & \cellcolor{blue!10}\color{gray50}{\faCheck} & \cellcolor{blue!10}\color{gray50}{\faCheck} & \multicolumn{1}{cV{2.5}}{13\%} \\

    \multicolumn{1}{V{2.5}lV{2.5}}{\texttt{GPU Frequency}} & \cellcolor{blue!10}\color{gray50}{\faCheck} & \cellcolor{blue!10}\color{gray50}{\faCheck} & \cellcolor{blue!10}\color{gray50}{\faCheck} & \cellcolor{blue!10}\color{gray50}{\faCheck} & \multicolumn{1}{cV{2.5}}{22\%} \\

    \multicolumn{1}{V{2.5}lV{2.5}}{\texttt{Scheduler Policy}} & $\cdot$ & \cellcolor{orange!12}\color{gray50}{\faCheck} & \cellcolor{orange!12}\color{gray50}{\faCheck} & $\cdot$ & \multicolumn{1}{cV{2.5}}{.} \\

    \multicolumn{1}{V{2.5}lV{2.5}}{\texttt{kernel.sched\_rt\_runtime\_us}} & $\cdot$ & $\cdot$ & \cellcolor{orange!12}\color{gray50}{\faCheck} & $\cdot$ & \multicolumn{1}{cV{2.5}}{.} \\

    \multicolumn{1}{V{2.5}lV{2.5}}{\texttt{kernel.sched\_child\_runs\_first}} & $\cdot$ & $\cdot$ & \cellcolor{orange!12}\color{gray50}{\faCheck} & $\cdot$ & \multicolumn{1}{cV{2.5}}{.} \\

    \multicolumn{1}{V{2.5}lV{2.5}}{\texttt{vm.dirty\_background\_ratio}} & $\cdot$ & $\cdot$ & $\cdot$ & $\cdot$ & \multicolumn{1}{cV{2.5}}{.} \\

    \multicolumn{1}{V{2.5}lV{2.5}}{\texttt{vm.dirty\_ratio}} & $\cdot$ & $\cdot$ & \cellcolor{orange!12}\color{gray50}{\faCheck} & $\cdot$ & \multicolumn{1}{cV{2.5}}{.} \\

    \multicolumn{1}{V{2.5}lV{2.5}}{\texttt{Drop Caches}} & $\cdot$ & \cellcolor{orange!12}\color{gray50}{\faCheck} & \cellcolor{orange!12}\color{gray50}{\faCheck} & $\cdot$ & \multicolumn{1}{cV{2.5}}{.} \\

    \multicolumn{1}{V{2.5}lV{2.5}}{\texttt{CUDA\_STATIC}} & \cellcolor{blue!10}\color{gray50}{\faCheck} & \cellcolor{blue!10}\color{gray50}{\faCheck} & \cellcolor{blue!10}\color{gray50}{\faCheck} & \cellcolor{blue!10}\color{gray50}{\faCheck} & \multicolumn{1}{cV{2.5}}{55\%} \\

    \multicolumn{1}{V{2.5}lV{2.5}}{\texttt{vm.vfs\_cache\_pressure}} & $\cdot$ & $\cdot$ & $\cdot$ & $\cdot$ & \multicolumn{1}{cV{2.5}}{.} \\

    \multicolumn{1}{V{2.5}lV{2.5}}{\texttt{vm.swappiness}} & $\cdot$ & \cellcolor{orange!12}\color{gray50}{\faCheck} & \cellcolor{orange!12}\color{gray50}{\faCheck} & $\cdot$ & \multicolumn{1}{cV{2.5}}{1\%}\\ \hlineB{2.5}

    \multicolumn{1}{l}{} & \multicolumn{1}{l}{} & \multicolumn{1}{l}{} & \multicolumn{1}{l}{} & \multicolumn{1}{l}{} & \multicolumn{1}{l}{} \\[-0.95em] \clineB{1-5}{2.5}
     
    \multicolumn{1}{V{2.5}lV{2.5}}{Latency (\txtwo frames/sec)} & \textbf{28} & 24 & 21 & \multicolumn{1}{lV{2.5}}{23} &  \bigstrut[t]\\
    \multicolumn{1}{V{2.5}lV{2.5}}{Latency Gain (over \txone)} & \textbf{65\% } & 41\% & 24\% & \multicolumn{1}{lV{2.5}}{35\%} &  \\
    \multicolumn{1}{V{2.5}lV{2.5}}{Latency Gain (over default)} & \textbf{7$\times$} & 6$\times$ & 5.25$\times$ & \multicolumn{1}{lV{2.5}}{5.75$\times$} &  \\
    \multicolumn{1}{V{2.5}lV{2.5}}{Resolution time} & \textbf{22 mins} & 4 hrs & 4 hrs & \multicolumn{1}{lV{2.5}}{2 days} &  \\ \clineB{1-5}{2.5}
\end{tabular}}%
\vspace{1em}

\caption{\small Using \ourapproach on a real-world performance issue.}
\label{fig:real_example}
\end{figure}

\paragraph{Stage III: Iterative Sampling (Active Learning).}
\label{sect:path_discovery}



At this stage, \ourapproach determines the next configuration to be measured. 
\ourapproach first estimates the causal effects of configuration options towards performance objectives using the learned causal performance model. Then, \ourapproach iteratively determines the next system configuration using the estimated causal effects as a heuristic. Specifically, \ourapproach determines the value assignments for options with a probability that is determined proportionally based on their associated causal effects. The key intuition is that such changes in the options are more likely to have a larger effect on performance objectives, and therefore, we can learn more about the performance behavior of the system. Given the exponentially large configuration space and the fact that the span of performance variations is determined by a small percentage of configurations, if we had ignored such estimates for determining the change in configuration options, the next configurations would result in considerable variations in performance objectives comparing with the existing data. Therefore, measuring the next configuration would not provide additional information for the causal model. 

We extract paths from the causal graph (referred to as \textit{causal paths}) and rank them from highest to lowest based on their average causal effect on latency, and energy. Using path extraction and ranking, we reduce the complex causal graph into a few useful causal paths for further analyses. The configurations in this path are more likely to be associated with the root cause of the fault.

\noindent\textbf{Extracting causal paths with backtracking.}~A causal path is a directed path originating from either the configuration options or the system event and terminating at a non-functional property (\ie, throughput and/or energy). To discover causal paths, we backtrack from the nodes corresponding to each non-functional property until we reach a node with no parents. If any intermediate node has more than one parent, then we create a path for each parent and continue backtracking on each parent. 

\noindent\textbf{Ranking causal paths.~}~A complex causal graph can result in many causal paths. It is not practical to reason over all possible paths, as it may lead to a combinatorial explosion. Therefore, we rank the paths in descending of their causal effect on each non-functional property. For further analysis, we use paths with the highest causal effect.
To rank the paths, we measure the causal effect of changing the value of one node (say \texttt{Batch Size} or $X$) on its successor (say \texttt{Cache Misses} or $Z$) in the path (say \texttt{Batch Size}~\edgeone \texttt{Cache Misses} \edgeone \texttt{FPS} and \texttt{Energy}). We express this with the \textit{do-calculus}~\cite{pearl2009causality} notation: $\mathbb{E}[Z~|~\mathit{do}(X=x)]$. This notation represents the expected value of $Z$ (\texttt{Cache Misses}) if we set the value of the node $X$ (\texttt{Batch Size}) to $x$. To compute the \textit{average causal effect} (ACE) of $X\rightarrow Z$ (\ie, \texttt{Batch Size} \edgeone \texttt{Cache Misses}), we find the average effect over all permissible values of $X$ (\texttt{Batch Size}), \ie, $\mathrm{ACE}\left(Z, X\right) = \frac{1}{N}\cdot \sum_{\forall a, b\in X}\mathbb{E}\left[Z~|~\mathit{do}\left(X=b\right)\right]~-~ \mathbb{E}\left[Z~|~\mathit{do}\left(X=a\right)\right]$.  Here $N$ represents the total number of values $X$ (\texttt{Batch Size}) can take. If changes in \texttt{Batch Size} result in a large change in \texttt{Cache Misses}, then $\mathrm{ACE}\left(Z, X\right)$ will be larger, indicating that \texttt{Batch Size} has a large causal effect on \texttt{Cache Misses}.




\paragraph{Stage IV: Update Causal Performance Model.}
\label{sect:incremental_learning}


At each iteration, \ourapproach measures the configuration that is determined in the previous stage and updates the causal performance model incrementally (shown in~\fig{causal_model_update}). 
Since the causal model uses limited observational data, there may be a discrepancy between the underlying performance model and the learned causal performance model, note that this issue exists in all domains using data-driven models, including causal reasoning~\cite{pearl2009causality}. The more accurate the causal graph, the more accurate the proposed intervention will be~\cite{spirtes2000causation,ogarrio2016hybrid, glymour2019review,colombo2012learning,colombo2014order}. \fig{incremental_update} (a) shows an example of an iterative decrease of hamming distance~\cite{norouzi2012hamming} between the learned causal model and (approximate) ground truth causal model. \fig{incremental_update} (b), \ref{fig:incremental_update} (c), and \ref{fig:incremental_update} (d) shows the iterative behavior of \ourapproach while debugging a multi-objective performance fault. In case our repairs do not fix the faults, we update the observational data with this new configuration and repeat the process. Over time, the estimations of causal effects will become more accurate. We terminate the incremental learning once we achieve the desired performance.


    

\paragraph{Stage V: Estimate Performance Queries.}
\label{sect:path_discovery}
At this stage, given the learned causal performance model, \ourapproach's inference engine estimates the user-specified queries using the mathematics of causal reasoning--do-calculus.
Specifically, the causal inference engine provides a quantitative estimate for the identifiable queries on the current causal model and may return some queries as unidentifiable. It also determines what assumptions or new measurements are required to answer the ``unanswerable`` questions, so, the user can decide to incorporate these new assumptions by defining more constraints or increasing the sampling budgets.

\section{Case Study}
\label{sec:casestudy}

\begin{figure}[tp!]
    \centering
    \includegraphics*[width=\linewidth]{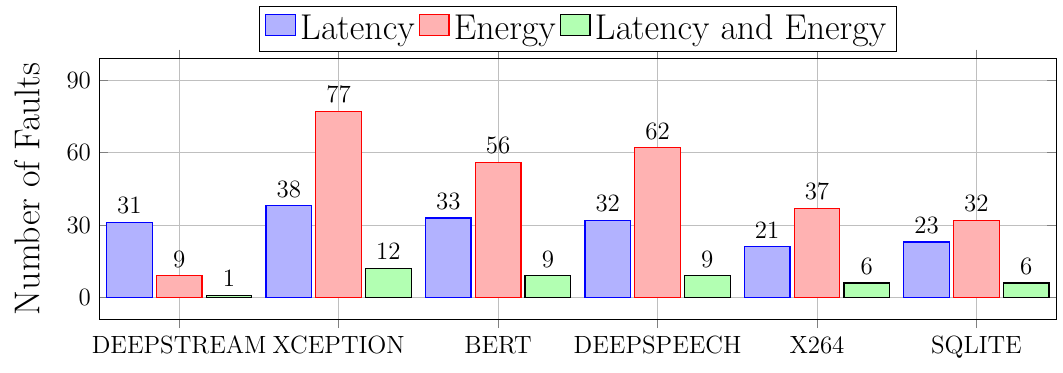}
    \caption{\small {Distribution of 451 single-objective and 43 multi-objective non-functional faults across different software systems used in our study.}}
    
    \label{fig:jetson_faults}
\end{figure}
Prior to a systematic evaluation in \S\ref{sec:evaluation}, here, we show how \ourapproach can enable performance debugging in a real-world scenario discussed in \cite{code_transplant:online}, where a developer migrated a real-time scene detection system from NVIDIA \txone to a more powerful hardware, \txtwo. The developer, surprisingly, experienced $4\times$ worse latency in the new environment (from 17 frames/sec in \txone to 4 frames/sec in \txtwo). 
After two days of discussions, the performance issue was diagnosed with a misconfiguration--an incorrect setting of a compiler option and four hardware options.
Here, we assess whether and how \ourapproach could facilitate the performance debugging by comparing with (i) the fix suggested by NVIDIA in the forum, and two academic performance debugging approaches--\bugdoc~\cite{lourencco2020bugdoc} and SMAC~\cite{hutter2011sequential}.

\noindent\textbf{Findings.~}~\fig{real_example} illustrates our findings. We find that:
\begin{itemize}
    \item \ourapproach could diagnose the root cause of the misconfiguration and recommends a fix within 22 minutes. Using the recommended configuration from \ourapproach, we achieved a throughput of 28 frames/sec ($65\%$ higher than \txone and $7\times$ higher than the fault). This, surprisingly, exceeds the developers' initial expectation of $30-40\%$ improvement. 
    \item \bugdoc (a diagnosis approach) has the least improvement compared to other approaches ($24\%$ improvement over \txone) while taking 4 hours to suggest the fix. \bugdoc also changed several unrelated options (depicted by \colorbox{orange!12}{\color{gray50}{\faCheck}}) not endorsed by the domain experts. 
    \item Using SMAC (an optimization approach), we aimed to find a configuration that achieves optimal throughput. However, after converging, SMAC recommended a configuration which achieved 24 frames/sec ($41\%$ better than \txone and $6\times$ better than the fault), however, could not outperform the configuration suggested by \ourapproach and even took 4 hours ($11\times$ longer than \ourapproach to converge). In addition, SMAC changed several unrelated options (\colorbox{orange!12}{\color{gray50}{\faCheck}} in~\fig{real_example}). 
\end{itemize}


\noindent\textbf{Why \ourapproach works better (and faster)?~}~%
\ourapproach discovers the misconfigurations by constructing a causal model that rules out irrelevant configuration options and focuses on the configurations that have the highest (direct or indirect) causal effect on latency, \eg, we found the root-cause \texttt{CUDA} \texttt{STATIC} in the causal graph which indirectly affects latency via \texttt{Context Switches} (an intermediate system event). Using counterfactual queries, \ourapproach can reason about changes to configurations with the highest average causal effect (ACE) (last column in~\fig{real_example}). The counterfactual reasoning occurs no additional measurements, significantly speeding up inference as shown in \fig{real_example}, \ourapproach accurately finds all the configuration options recommended by the forum (depicted by \colorbox{blue!10}{\color{gray50}{\faCheck}} in~\fig{real_example}).

%


\section{Evaluation}
\label{sec:evaluation}
\begin{table}[tb!]
\caption{Overview of the subject systems used in our study. Details about the configuration options and system events for each system are found in the \href{https://github.com/softsys4ai/unicorn}{\color{blue!80}supplementary materials}.}
\label{tab:subject_systems}
\resizebox{0.98\linewidth}{!}{%
\begin{tabular}{@{}lp{4cm}llllll@{}}

\toprule
System      & Workload     & $|\mathcal{C}|$ & $|\mathcal{O}|$ & $|\mathcal{S}|$ & $|\mathcal{H}|$ & $|\mathcal{W}|$ & $|\mathcal{P}|$  \\ \midrule
\textsc{Deepstream}~\cite{DeepStream} & Video analytics pipeline for detection and tracking from 8 camera streams. & 2461    & 53 & 288  & 2  & 1 & 2 \\

\textsc{Xception}~\cite{chollet2017xception}   & Image recognition system to classify 5000/5000 test images from CIFAR10. & 6443  & 28 & 19  & 3  & 3 & 3 \\

\textsc{Deepspeech}~\cite{hannun2014deep} &Speech-to-text from 0.5/1932 hours of Common Voice Corpus 5.1 (English) data.      & 6112  & 28 & 19  & 3  & 1 & 3 \\

\textsc{Bert}~\cite{devlin2018bert}       & NLP system for sentiment analysis of 1000/25000 test reviews from IMDb.           & 6188  & 28 & 19  & 3  & 1 & 3 \\

\textsc{x264}~\cite{x264}       & Encodes a 20 second 11.2 MB video of resolution 1920 x 1080 from UGC.  & 17248  & 32 & 19  & 3  & 1 & 3 \\

\textsc{SQLite}~\cite{SQLite}     & Database engine for sequential \& batch \& random reads, writes, deletions.     & 15680   & 242 & 288  & 3  & 3 & 3 \\

\bottomrule
\end{tabular}}
{* \tiny $\mathcal{C}$: Configurations, $\mathcal{O}$: Options, $\mathcal{S}$: System Events, $\mathcal{H}$: Hardware, $\mathcal{W}$: Workload, $\mathcal{P}$: Objectives}
\vspace{-4mm}
\end{table}

For a thorough evaluation of \ourapproach, we have developed \ourtool that implements the methodology that we explained in \S\ref{sec:methodology}.
We used \ourtool (see \S\ref{sec:artifact}) to facilitate comparing \ourapproach with state-of-the-art performance debugging and optimization approaches for: 
\begin{itemize}
\item \textbf{Effectiveness} in terms of sample efficiency and performance gain (\S\ref{sec:effectiveness}).
\item \textbf{Transferability} of learned models across environmental changes such as hardware and workload changes (\S\ref{sec:transfer}).
\item \textbf{Scalability} to large-scale configurable systems (\S\ref{sec:scalability}).
\end{itemize}



\noindent
\textbf{Systems.} We selected six configurable systems including a video analytic pipeline, three deep learning-based systems (for image, speech, and NLP), a video encoder, and a database, see Table~\ref{tab:subject_systems}. We use heterogeneous deployment platforms, including \textsc{NVIDIA} \txone, \txtwo, and \xavier, each having different resources (compute, memory) and microarchitectures.

\noindent\textbf{Configurations.} We choose a wide range of configuration options and system events (see Table~\ref{tab:subject_systems}), following \textsc{NVIDIA}'s configuration guides/tutorials and other related work~\cite{Halawa2017}. As opposed to prior works (e.g.,~\cite{velez2021white,VJSSAK:ASE20}) that only support binary options due to scalability issues, we included options with binary, discrete, and continuous.



\noindent \textbf{Ground truth.} We measured several thousands samples (proportional to the configuration space of the system, see \href{https://github.com/softsys4ai/unicorn}{\color{blue!80}supplementary materials} for specific dataset size) for each 18 deployment settings (6 systems and 3 hardware; see Table \ref{tab:subject_systems} for more details). To ensure reliable and replicable results, following the common practice~\cite{ding2021generalizable,JSVKPA:ASE17,curtsinger2013stabilizer,kaltenecker2020interplay}, we repeated each measurement $5$ times and used the median in the evaluation metrics. 
We curated a ground truth of performance issues, called {\sc Jetson Faults}, for each of the studied software and hardware systems using the ground truth data. By definition, non-functional faults are located in the tail of performance distributions~\cite{gunawi2018fail,kleppmann2017designing}. We, therefore, selected and labeled configurations that are worse than the $99^\text{th}$ percentile as `\textit{faulty}.' ~\fig{jetson_faults} shows the total 494 faults discovered across different software. Out of these 494 non-functional faults, 43 are faults with multiple types (both energy and latency). Of all the 451 single-objective and 43 multi-objective faults discovered in this study, only 2 faults had a single root cause, 411 faults had five or more root causes, and 81 remaining faults had two to four root causes. 

\noindent \textbf{Experimental parameters}. To facilitate replication of the results, we made some choices for specific parameters. In particular, we disabled dynamic voltage and frequency scaling (DVFS) before starting any experiment and start with 25 samples for each method~(10\% of the total sampling budget). We repeat the entire process 3 times for consistent analyses.

\begin{table*}[tb!]
    \centering
    \caption{\small  Efficiency of \tool compared to other approaches. Cells highlighted in \colorbox{blue!10}{\bfseries blue} indicate improvement over faults.
    }
\vspace{-0.85em}
\subfloat[Single objective performance fault for \textit{latency and energy} in \txtwo and \xavier, respectively.]{\scriptsize
    \label{tab:single_1}
    \resizebox{\textwidth}{!}{
    \begin{tabular}{@{}l|l|l|lllll|lllll|lllll|lllll|ll|}
        \clineB{4-25}{2}
        \multicolumn{1}{c}{}&\multicolumn{1}{c}{}  &  & \multicolumn{5}{c|}{Accuracy} & \multicolumn{5}{c|}{Precision} & \multicolumn{5}{c|}{Recall} & \multicolumn{5}{c|}{Gain} & \multicolumn{2}{c|}{Time$^\dagger$} \bigstrut\\ \clineB{4-25}{2}
        \multicolumn{1}{c}{}& \multicolumn{1}{c}{} &  & \multicolumn{1}{c}{\rotatebox{90}{\bfseries \tool}} &
        \multicolumn{1}{c}{\rotatebox{90}{\cbi}} & \multicolumn{1}{c}{\rotatebox{90}{DD}} & \multicolumn{1}{c}{\rotatebox{90}{\encore}} & \multicolumn{1}{c|}{\rotatebox{90}{\bugdoc~}} & \multicolumn{1}{c}{\rotatebox{90}{\bfseries \tool}} &  
        \multicolumn{1}{c}{\rotatebox{90}{\cbi}} & \multicolumn{1}{c}{\rotatebox{90}{DD}} & \multicolumn{1}{c}{\rotatebox{90}{\encore}} & \multicolumn{1}{c|}{\rotatebox{90}{\bugdoc~}} & \multicolumn{1}{c}{\rotatebox{90}{\bfseries \tool}} &
        \multicolumn{1}{c}{\rotatebox{90}{\cbi}} & \multicolumn{1}{c}{\rotatebox{90}{DD}} & \multicolumn{1}{c}{\rotatebox{90}{\encore}} & \multicolumn{1}{c|}{\rotatebox{90}{\bugdoc~}} & \multicolumn{1}{c}{\rotatebox{90}{\bfseries \tool}} & \multicolumn{1}{c}{\rotatebox{90}{\cbi}} & \multicolumn{1}{c}{\rotatebox{90}{DD}} & \multicolumn{1}{c}{\rotatebox{90}{\encore}} & \multicolumn{1}{c|}{\rotatebox{90}{\bugdoc
        ~}} & \multicolumn{1}{c}{\rotatebox{90}{\bfseries \tool}} &  \multicolumn{1}{c|}{\rotatebox{90}{Others}} \bigstrut[t]
        \\ \clineB{4-25}{2}
    \multicolumn{1}{c}{}&\multicolumn{1}{c}{}  & \multicolumn{1}{c}{} & \multicolumn{1}{c}{} & \multicolumn{1}{c}{} & \multicolumn{1}{c}{} & \multicolumn{1}{c}{} & \multicolumn{1}{c}{} \bigstrut\\[-1.4em]\hlineB{2}
    &  & \textsc{DeepStream} & \cellcolor{blue!10}\bfseries87 & 61 & 62 & 65 & 81 & \cellcolor{blue!10}\bfseries 83 & 66 & 59 & 60 & 71 & \cellcolor{blue!10}\bfseries80  & 61 & 65 & 60 & 70 & \cellcolor{blue!10}\bfseries88  & 66 & 67 & 68 & 79 & \cellcolor{blue!10}\bfseries0.8 &4 \\
     &  & \textsc{Xception} & \cellcolor{blue!10}\bfseries86 & 53 & 42 & 62 & 65 & \cellcolor{blue!10}\bfseries 86 & 67 & 61 & 63 & 67 & \cellcolor{blue!10}\bfseries83  & 64 & 68 & 69 & 62 & \cellcolor{blue!10}\bfseries82  & 48 & 42 & 57 & 59 & \cellcolor{blue!10}\bfseries0.6 &4 \\
     &  & \textsc{BERT} & \cellcolor{blue!10}\bfseries81 &56 & 59 & 60 & 57 & \cellcolor{blue!10}\bfseries76  & 57 & 55 & 61 & 73 & \cellcolor{blue!10}\bfseries 71 & 74 & 68 & 67 & 65 & \cellcolor{blue!10}\bfseries74  & 54 & 59 & 62 & 58 & \cellcolor{blue!10}\bfseries0.4 & 4 \\
     &  & \textsc{Deepspeech} & \cellcolor{blue!10}\bfseries81  & 61 & 59 & 60 & 72 & \cellcolor{blue!10}\bfseries76  & 58 & 69 & 61 & 71 & \cellcolor{blue!10}\bfseries81  & 73 & 61 & 63 & 69 & \cellcolor{blue!10}\bfseries76 & 59 & 53 & 55 & 66 & \cellcolor{blue!10}\bfseries0.7  & 4 \\
     \multirow{-4}{*}{\rotatebox{90}{\txtwo}} & \multirow{-4}{*}{\rotatebox{90}{Latency}} & \textsc{x264} & \cellcolor{blue!10}\bfseries83  & 59 & 63 & 62 & 62 & \cellcolor{blue!10}\bfseries82  &69 & 58 & 65 & 66 & \cellcolor{blue!10}\bfseries78 & 64 & 67 & 63 & 72 & \cellcolor{blue!10}\bfseries85  & 69 & 72 & 68 & 71 & \cellcolor{blue!10}\bfseries1.4  & 4 \\ \hlineB{2}
    \multicolumn{1}{c}{}&\multicolumn{1}{c}{}  & \multicolumn{1}{c}{} & \multicolumn{1}{c}{} & \multicolumn{1}{c}{} & \multicolumn{1}{c}{} & \multicolumn{1}{c}{} & \multicolumn{1}{c}{} \bigstrut\\[-1.55em]\hlineB{2}
    &  & \textsc{DeepStream} & \cellcolor{blue!10}\bfseries91 & 81 & 79 & 77 & 87 & \cellcolor{blue!10}\bfseries81 &61 & 62 & 64 & 73 & \cellcolor{blue!10}\bfseries85 & 63 & 61 & 62 & 75 & \cellcolor{blue!10}\bfseries86 & 68 & 62 & 61 & 78 & \cellcolor{blue!10}\bfseries0.7 & 4 \\
     &  & \textsc{Xception} & \cellcolor{blue!10}\bfseries84 & 66 & 63 & 63 & 81 & \cellcolor{blue!10}\bfseries78 &56 & 58 & 66 & 65 & \cellcolor{blue!10}\bfseries80 & 69 & 55 & 63 & 68 & \cellcolor{blue!10}\bfseries83 & 59 & 50 & 51 & 62 & \cellcolor{blue!10}\bfseries0.4 & 4 \\
     &  & \textsc{BERT} & 66& 59 & 53 & 63 & \cellcolor{blue!10}\bfseries72 & \cellcolor{blue!10}\bfseries70  & 62 & 64 & 64 & 65 & \cellcolor{blue!10}\bfseries79  & 61 & 54 & 63 & 66 & \cellcolor{blue!10}\bfseries62  & 49 & 36 & 49 & 53 & \cellcolor{blue!10}\bfseries0.5  & 4 \\
     &  & \textsc{Deepspeech} & \cellcolor{blue!10}\bfseries73  &68 & 63 & 72 & 71 & \cellcolor{blue!10}\bfseries75 & 55 & 59 & 54 & 68 & \cellcolor{blue!10}\bfseries78  &53 & 52 & 59 & 71 & \cellcolor{blue!10}\bfseries78 & 64 & 48 & 65 & 63 & \cellcolor{blue!10}\bfseries1.2  & 4 \\
     \multirow{-4}{*}{\rotatebox{90}{\xavier}} & \multirow{-4}{*}{\rotatebox{90}{Energy}}& \textsc{x264} & \cellcolor{blue!10}\bfseries77  &71 & 70 & 74 & 74 & \cellcolor{blue!10}\bfseries83  & 63 & 53 & 61 & 66 & \cellcolor{blue!10}\bfseries78  & 67 & 53 & 54 & 72 & \cellcolor{blue!10}\bfseries 87  & 73 & 71 & 76 & 76 & \cellcolor{blue!10}\bfseries0.3  &4 \\ \hlineB{2}
    \end{tabular}
    }}
    \\
    \subfloat[Multi-objective non-functional faults in \textit{Energy, Latency} in \xavier.]{
        \scriptsize
        \label{tab:multi_1}
        \resizebox{\textwidth}{!}{
            \begin{tabular}{@{}r@{}ll|llll|llll|llll|llll|llll|ll|}
            \clineB{4-25}{2}
            &  &  & \multicolumn{4}{c|}{Accuracy} & \multicolumn{4}{c|}{Precision} & \multicolumn{4}{c|}{Recall} & \multicolumn{4}{c|}{Gain (Latency)} & \multicolumn{4}{c|}{Gain (Energy)}  & \multicolumn{2}{c|}{Time$^\dagger$} \bigstrut\\ \clineB{4-25}{2}
            
            &  &  & \rotatebox{90}{\bfseries \tool~} & \rotatebox{90}{\cbi} & \rotatebox{90}{\encore} & \rotatebox{90}{\bugdoc} & \rotatebox{90}{\bfseries \tool~}  & \rotatebox{90}{\cbi} & \rotatebox{90}{\encore} & \rotatebox{90}{\bugdoc} & \rotatebox{90}{\bfseries \tool~} & \rotatebox{90}{\cbi} & \rotatebox{90}{\encore} & \rotatebox{90}{\bugdoc} & \rotatebox{90}{\bfseries \tool~} & \rotatebox{90}{\cbi} & \rotatebox{90}{\encore} & \rotatebox{90}{\bugdoc} & \rotatebox{90}{\bfseries \tool~} & \rotatebox{90}{\cbi} & \rotatebox{90}{\encore} & \rotatebox{90}{\bugdoc} & \rotatebox{90}{\bfseries \tool~}  & \rotatebox{90}{Others} \\ \clineB{4-25}{2}
            
            \multicolumn{1}{l}{}&\multicolumn{1}{l}{}  & \multicolumn{1}{l}{} & \multicolumn{1}{l}{} & \multicolumn{1}{l}{} & \multicolumn{1}{l}{} & \multicolumn{1}{l}{} & \multicolumn{1}{l}{} & \multicolumn{1}{l}{} & \multicolumn{1}{l}{} \\[-0.85em]\hlineB{2}

            & \multicolumn{1}{l|}{} & \textsc{Xception} & \cellcolor{blue!10}\textbf{89} & 76 & 81 & 79 & \cellcolor{blue!10}\textbf{77} & 53 & 54 & 62 & \cellcolor{blue!10}\textbf{81} & 59 & 59 & 62 & \cellcolor{blue!10}\textbf{84} & 53 & 61 & 65 & \cellcolor{blue!10}\textbf{75} & 38 & 46 & 44 & \cellcolor{blue!10}\textbf{0.9} & 4 \\
            
            & \multicolumn{1}{l|}{} & \textsc{BERT} & {71} &72 & \cellcolor{blue!10}\textbf{73} & 71 & \cellcolor{blue!10}\textbf{77} & 42 & 56 & 63 & \cellcolor{blue!10}\textbf{79} & 59 & 62 & 65 & \cellcolor{blue!10}\textbf{84} & 53 & 59 & 61 & \cellcolor{blue!10}\textbf{67}  & 41 & 27 & 48 & \cellcolor{blue!10}\textbf{0.5}  & 4 \\
            
            & \multicolumn{1}{l|}{} & \textsc{Deepspeech} & \cellcolor{blue!10}\textbf{86}  & 69 & 71 & 72 & \cellcolor{blue!10}\textbf{80} & 44 & 53 & 62 & \cellcolor{blue!10}\textbf{81}  & 51 & 59 & 64 & \cellcolor{blue!10}\textbf{88}  & 55 & 55 & 62 & \cellcolor{blue!10}\textbf{77}  & 43 & 43 & 41  & \cellcolor{blue!10}\textbf{1.1}  & 4 \\
            \multirow{-4}{*}{\rotatebox{90}{Energy +}} & \multicolumn{1}{l|}{\multirow{-4}{*}{\rotatebox{90}{Latency}}} & \multicolumn{1}{l|}{\textsc{x264}} & \cellcolor{blue!10}\textbf{85}  & 73 & 83 & 81 & \cellcolor{blue!10}\textbf{83}  & 50 & 54 & 67 & \cellcolor{blue!10}\textbf{80}  & 63 & 62 & 61 & \cellcolor{blue!10}\textbf{75}  & 62 & 64 & 66 & \cellcolor{blue!10}\textbf{76}  & 64 & 66 & 64  & \cellcolor{blue!10}\textbf{1}  & 4 \\\hlineB{2}
           \multicolumn{10}{l}{$^\dagger$ Wallclock time in hours}\bigstrut
           \end{tabular}
    }}\vspace{-2.5em}
\end{table*}

\noindent
\textbf{Baselines.} 
We evaluate \ourapproach for two performance tasks: (i) performance debugging and repair and (ii) performance optimization. We compare \ourapproach against state-of-the-art, including \cbi~\cite{song2014statistical}---a statistical debugging method that uses a feature selection algorithm; \textsc{DD}~\cite{artho2011iterative}---a delta debugging technique, that minimizes the difference between a pair of configurations; 
\encore~\cite{zhang2014encore}---a debugging method that learns to debug from correlational information about misconfigurations;
\bugdoc~\cite{lourencco2020bugdoc}---a debugging method that infers the root causes and derives succinct explanations of failures using decision trees;
\textsc{SMAC}~\cite{hutter2011sequential}---a sequential model-based auto-tuning approach; and \textsc{PESMO}~\cite{hernandez2016predictive}---a multi-objective Bayesian optimization approach.

\noindent
\textbf{Evaluation metrics.}
(i) \emph{Accuracy} is calculated by weighted Jaccard similarity between the predicted and true root causes, where the weight vector was derived based on the average causal effect of options to performance based on the ground-truth causal performance model. For example, if  $A$ is the recommended configuration by an approach and $B$ is the configuration that fixes the performance issue in the ground truth, we measure $accuracy=\frac{\sum_{\text{ACE}} (\text{A} \cap \text{B})}{\sum_{\text{ACE}} (\text{A} \cup \text{B})}$. The key intuition is that an ideal causal model underlying the system should identify the most important options that affect performance objectives. In other words, an ideal causal model should provide recommendations for changing the values of options that have the highest average causal effects on system performance. (ii) \emph{Precision} is calculated by the percentage of true root causes among the predicted ones.  (iii)~\emph{Recall} is calculated by the percentage of true root causes that are correctly predicted.
(iv) \emph{Gain} is calculated by percentage improvement of suggested fix over the observed fault--$\Delta_{gain}=\frac{\text{NFP}_\textsc{fault}-\text{NFP}_\textsc{nofault}}{\text{NFP}_\textsc{fault}}\times 100$, where $\text{NFP}_{\textsc{fault}}$ the observed faulty performance and $\text{NFP}_{\textsc{no}~\textsc{fault}}$ is the performance of suggested fix. (v) \emph{Error} is calculated by the hypervolume error (in multi-objective)~\cite{zitzler2007hypervolume}.  (vi) \emph{Time} is measured by wallclock time (in hours) to suggest a fix.

\section{Effectiveness and Sample Efficiency}
\label{sec:effectiveness}

\noindent \textbf{Setting}. We only show the partial results, however, our results generalize to all evaluated settings. For \emph{debugging}, we use latency faults in \txtwo and energy faults in \xavier. For \emph{single-objective optimization}, we compare \ourapproach with SMAC for \textsc{Xception} for latency and energy and for \emph{multi-objective optimization} we compare with PESMO in \txtwo. 

\begin{figure}[tb!]
    \setlength{\belowcaptionskip}{-1.5em}
    \subfloat[Latency]{ \vspace{-1.5em}
        \includegraphics[width=\linewidth]{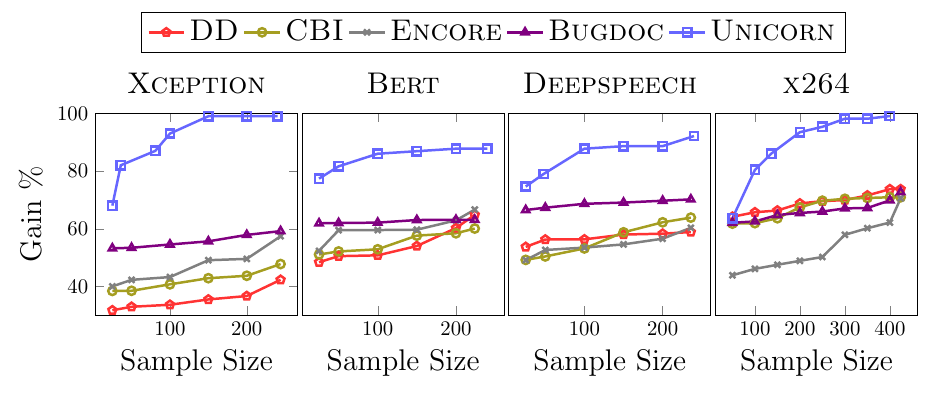}
        \label{fig:rq2_1}
    }\\
    \vspace{-1.5em}
    \subfloat[Energy]{
        \includegraphics[width=\linewidth]{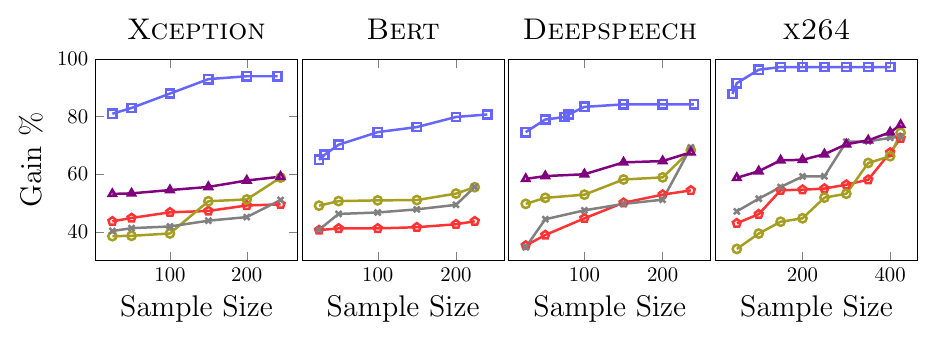}
        \label{fig:rq2_2}
    }
 
    \caption{\small{\ourapproach has significantly higher sampling efficiency than other baselines in debugging non-functional faults: (a) latency faults in \txtwo and (b) energy faults in \xavier.}}
    \label{tab:rq2_1}
\end{figure}
\noindent{\textbf{Results (debugging).}}~\Cref{tab:single_1,tab:multi_1} shows \ourapproach ~{\em significantly outperforms correlation-based methods in all cases}. For example, in  \textsc{Deepstream} on TX2, \ourapproach achieves 6\% more accuracy, 12\% more precision, and 10\% more recall compared to the next best method, \bugdoc. We observed latency gains as high as $88\%$ ($9\%$ more than \bugdoc) on \txtwo and energy gain of $86\%$ ($9\%$ more than \bugdoc) on \xavier for \textsc{Xception}. We observe similar trends for multi-objective faults as well. The results confirm that \ourapproach~{\em can recommend repairs for faults that significantly improve latency and energy}. By applying the changes to the configurations recommended by \ourapproach improves performance drastically.

\fig{rq2_1} and \fig{rq2_2} demonstrate the sample efficiency results for different systems. We observe that, for both latency and energy faults, \ourapproach achieved significantly higher gains with substantially fewer samples. For \textsc{Xception}, \ourapproach required a 8$\times$ fewer samples to obtain 32\% higher gain than \textsc{DD}. The higher gain in \ourapproach in comparison to correlation-based methods indicates that \ourapproach's causal reasoning is more effective in guiding the search in the objective space. \ourapproach does not waste budget evaluating configurations with lower causal effects and finds a fix faster.

\begin{figure}[tp!]
  
    \includegraphics[width=\linewidth]{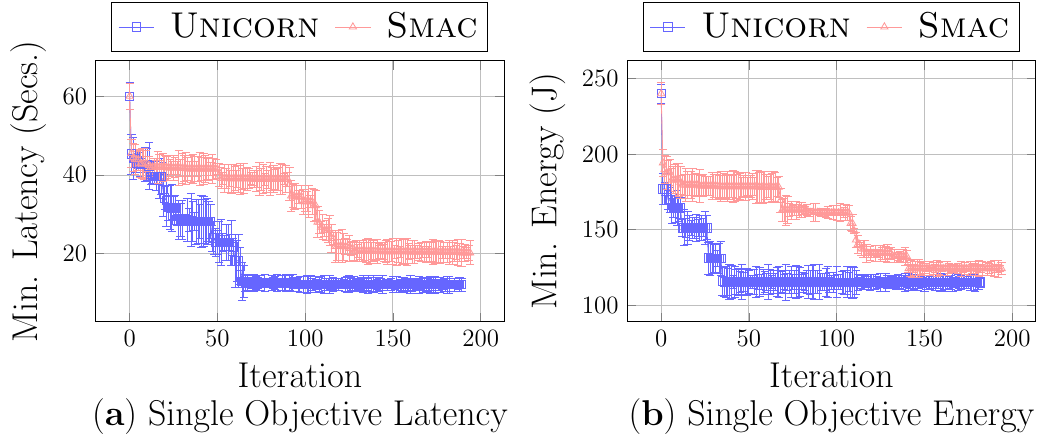} 
    \includegraphics[width=\linewidth]{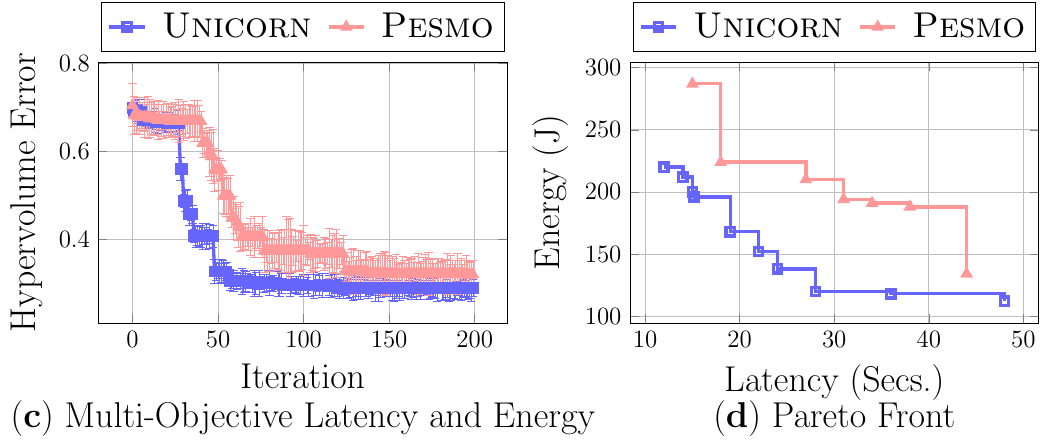}
    \vspace{-4mm}
    \caption{\small{\ourapproach vs. single and multi-objective optimization with SMAC and PESMO in \txtwo.}}
     \label{fig:rq1_opt_se}
    \vspace{-3mm}
\end{figure}

\ourapproach~{\em resolves misconfiguration faults significantly faster than correlation-based approaches}. In~\Cref{tab:single_1,tab:multi_1}, the last two columns indicate the time taken (in hours) by each approach to diagnosing the root cause. For all correlation-based methods, we set a maximum budget of 4 hours. We find that, while other approaches use the entire budget to diagnose and resolve the faults, \ourapproach can do so significantly faster. In particular, we observed that \ourapproach is $13\times$ faster in diagnosing and resolving faults in energy usage for \textsc{x264} deployed on \xavier and $10\times$ faster for latency faults for \textsc{Bert} deployed on \txtwo.

\noindent\textbf{Results (optimization).}~\fig{rq1_opt_se}(a) and \fig{rq1_opt_se}(b) demonstrate the single-objective optimization results---\ourapproach finds configurations with optimal latency and energy for both cases. \fig{rq1_opt_se}(a)  illustrates that the optimal configuration discovered by \ourapproach has 43\% lower latency (12 seconds) than that of SMAC (21 seconds). Here, \ourapproach reaches near-optimal configuration by only exhausting one-third of the entire budget. In \fig{rq1_opt_se}(b), the optimal configuration discovered by \ourapproach and SMAC had almost the same energy, but \ourapproach reached this optimal configuration 4x faster than SMAC. In both single-objective optimizations, the iterative variation of \ourapproach is less than SMAC--i.e., \ourapproach finds more stable configurations. \fig{rq1_opt_se}(c) compares \ourapproach with PESMO to optimize both latency and energy in \txtwo (for image recognition). Here, \ourapproach has 12\% lower hypervolume error than  PESMO and reaches the same level of hypervolume error of PESMO 4x times faster. \fig{rq1_opt_se}(d) illustrates the Pareto optimal configurations obtained by \ourapproach and PESMO. The Pareto front discovered by \ourapproach has higher coverage, as it discovers a larger number of Pareto optimal configurations with lower energy and latency value than PESMO.

\begin{figure}[tp!]
    \setlength{\belowcaptionskip}{-1.5em}
    \centering
    \includegraphics*[width=\linewidth]{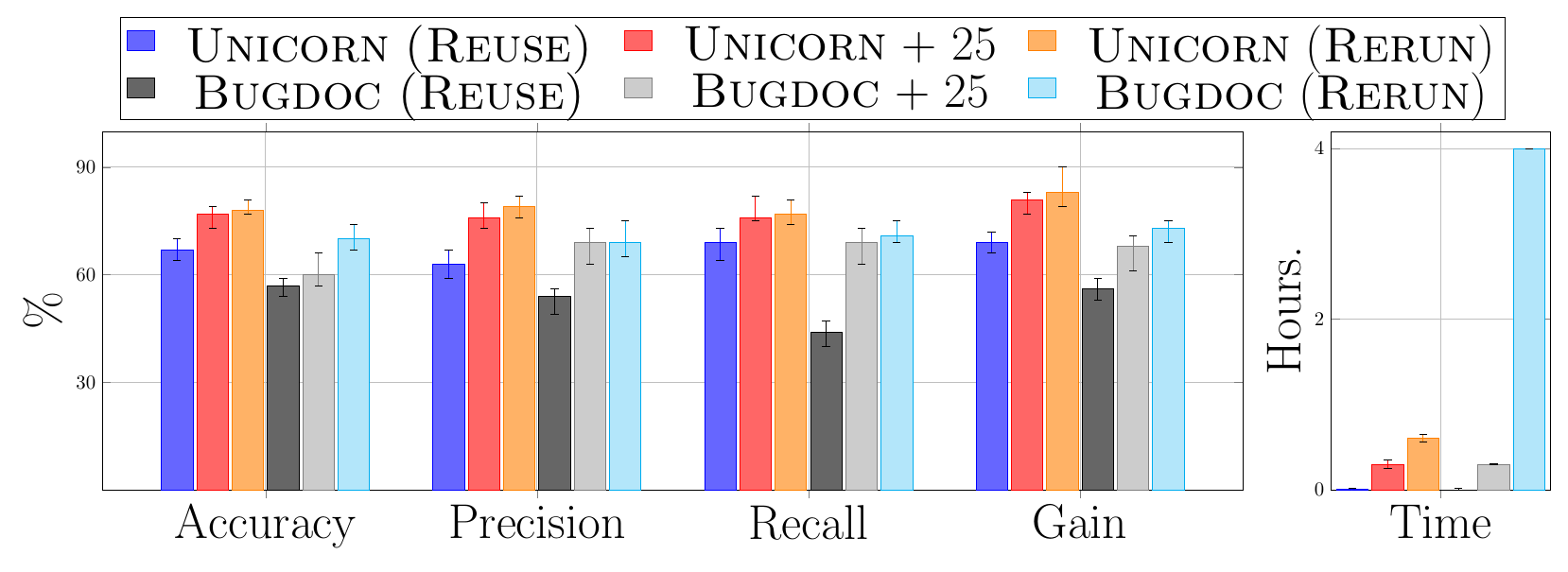}
    \caption{\small{{\ourapproach has higher accuracy, precision, recall, and gain in debugging non-functional energy faults when hardware changes (\xavier to \txtwo)}}.}
    \label{fig:rq1_debugging_multi}
\end{figure}

\begin{figure}[tp!]
    \centering
    \includegraphics*[width=\linewidth]{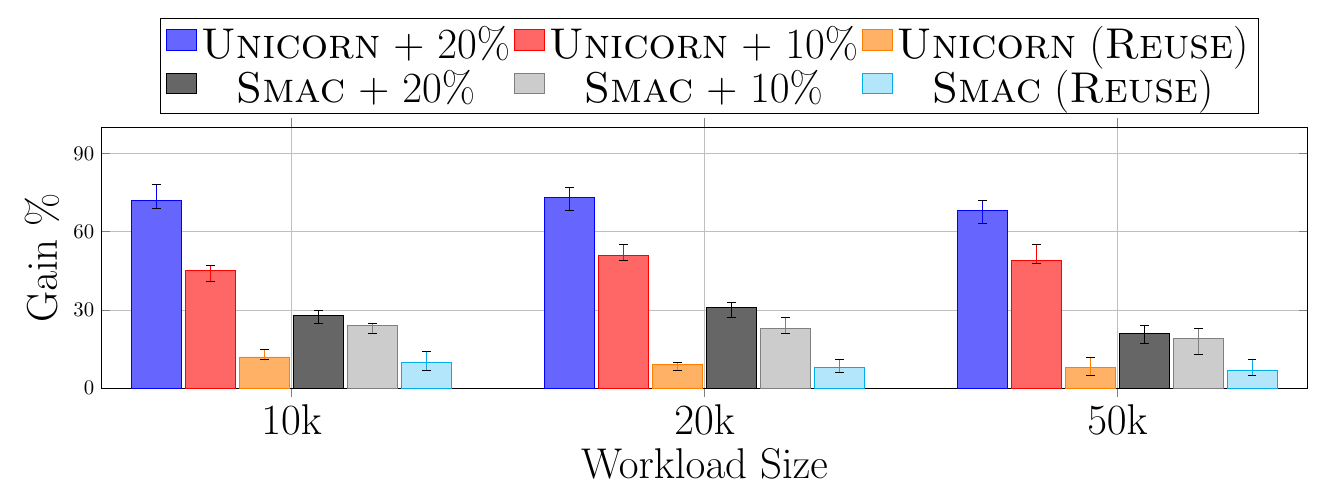}
    \caption{\small {\ourapproach finds configurations with higher gain when workloads are changed for performance (latency) optimization task in \txtwo.}}
    
    \label{fig:rq1_opt_me}
\end{figure}

\section{Transferability} 
\label{sec:transfer}



\noindent \textbf{Setting.} We reuse the causal performance model constructed from a source environment, e.g., \txone, to resolve a non-functional fault in a target environment, e.g., \xavier. We evaluated \ourapproach for debugging energy faults for \textsc{Xception} and used \xavier as the source and \txtwo as the target, since they have different microarchitectures, expecting to see large differences in their performance behaviors. We only compared with \bugdoc as it discovered fixes with higher energy gain in \xavier than other correlation-based baseline methods (see ~\Cref{tab:single_1}). We compared \ourapproach and \bugdoc in the following scenarios: (i) \bugdoc (\textsc{Reuse}) and \ourapproach (\textsc{Reuse}): reusing the recommended configurations from Source to Target, (ii) \bugdoc \textsc{+ 25} and \ourapproach \textsc{+ 25}: reusing the performance models (i.e., causal model and decision tree) learned in Source and fine-tuning the models with 25 new samples in Target, and (iii) \bugdoc (\textsc{Rerun}) and \ourapproach (\textsc{Rerun}): we rerun \ourapproach and \bugdoc from scratch to resolve energy faults in Target. 
For optimization tasks, we use three larger additional \textsc{Xception} workloads: 10000 (10k), 20000 (20k), and 50000 (50k) test images (previous experiments used 5000 (5k) test images). We evaluated three variants of SMAC and \ourapproach: 
(i)~SMAC (\textsc{Reuse}) and \ourapproach (\textsc{Reuse}), where we \textit{reuse} the near-optimum found with 5k test images on the larger workloads; 
(ii)~SMAC + 10\% and \ourapproach + 10\%, where we rerun with 10\% budget in target and update the optimization and causal performance model with 10\% additional budget; and
(iii)~SMAC + 20\% and \ourapproach + 20\%, where we rerun with 20\% budget in target and update the models with 20\% additional budget.

\noindent \textbf{Results.} 
\fig{rq1_debugging_multi} indicates the results in resolving energy faults in \txtwo. We observe that \ourapproach + 25 obtains 8\% more accuracy, 7\% more precision, 5\% more recall and 8\% more gain than \bugdoc \textsc{(Rerun)}. Here, \bugdoc takes significantly longer time than \ourapproach, \ie, \bugdoc (\textsc{Rerun}) exhausts the entire 4-hour budget whereas \ourapproach takes at most 20 minutes to fix the energy faults. Moreover, we have to rerun \bugdoc every time the hardware changes, and this limits its practical usability. In contrast, \ourapproach incrementally updates the internal causal model with new samples from the newer hardware to learn new relationships. We also observe that with little updates, \ourapproach + 25 ($\sim$20 minutes) achieves a similar performance of \ourapproach \textsc{(Rerun)} ($\sim$36 minutes). Since the causal mechanisms are sparse, the causal performance model from \xavier in \ourapproach quickly reaches a fixed structure in \txtwo using incremental learning by judiciously evaluating the most promising fixes until the fault is resolved.

Our experimental results demonstrate that \ourapproach performs better than the two variants of three SMAC (c.f. \fig{rq1_opt_me}). SMAC (\textsc{Reuse}) performs the worst when the workload changes. With 10K images, reusing the near-optimal configuration from 5K images results in a latency gain of 10\%, compared to 12\% with \ourapproach in comparison with the default configuration. We observe that \ourapproach + 20\% achieves 44\%, 42\%, and 47\% higher gain than SMAC + 20\% for workload sizes of 10k, 20k, and 50k images, respectively.

\begin{table}[tp!]
\caption{Scalability for \textsc{SQLite} and \textsc{DeepStream} on \xavier.}
\label{tab:scalability}
\resizebox{0.98\linewidth}{!}{
\begin{tabular}{V{2}rrrrrrrV{2}rrrV{2}}
\clineB{8-10}{2}
\multicolumn{1}{l}{}              & \multicolumn{1}{l}{}   & \multicolumn{1}{l}{}           &
\multicolumn{1}{l}{}           &
\multicolumn{1}{l}{}           &
\multicolumn{1}{l}{}         & \multicolumn{1}{cV{2}}{}              & \multicolumn{3}{cV{2}}{Time/Fault (in sec.)}                                                    \bigstrut\\ \clineB{1-7}{2}
\multicolumn{1}{V{2}r}{\rotatebox{90}{System}} &{\rotatebox{90}{Configs}}&
{\rotatebox{90}{Events}} &\multicolumn{1}{r}{\rotatebox{90}{Paths}} & \multicolumn{1}{r}{\rotatebox{90}{Queries}} &
\multicolumn{1}{r}{\rotatebox{90}{Degree}} &
\multicolumn{1}{rV{2}}{\rotatebox{90}{Gain (\%)}} & \multicolumn{1}{r}{\rotatebox{90}{Discovery}} & \multicolumn{1}{r}{\rotatebox{90}{Query Eval}} & \multicolumn{1}{rV{2}}{\rotatebox{90}{\bfseries Total}} \bigstrut\\ \hlineB{2}

\multicolumn{1}{c}{}& \multicolumn{1}{c}{}&\multicolumn{1}{c}{}& \multicolumn{1}{c}{}& \multicolumn{1}{c}{}& \multicolumn{1}{c}{}& \multicolumn{1}{c}{}& \multicolumn{1}{c}{}\bigstrut\\[-1.5em] \hlineB{2}

\multicolumn{1}{V{2}r}{\textsc{SQLite}}   & \multicolumn{1}{r}{34}  & \multicolumn{1}{r}{19}                    & \multicolumn{1}{r}{32}        & \multicolumn{1}{r}{191} &   \multicolumn{1}{r}{3.6} & 93                              & 9                            & 14                            & \cellcolor{blue!10}\textbf{291}                     \bigstrut\\ 
\multicolumn{1}{V{2}r}{}    & \multicolumn{1}{r}{242}  & \multicolumn{1}{r}{19}                   & \multicolumn{1}{r}{111}       & \multicolumn{1}{r}{2234}   & \multicolumn{1}{r}{1.9}& 94                               & 57                           & 129                           & \cellcolor{blue!10}\textbf{1345}                    \bigstrut\\
\multicolumn{1}{V{2}r}{}&  \multicolumn{1}{r}{242} & \multicolumn{1}{r}{288}                      & \multicolumn{1}{r}{441}       & \multicolumn{1}{r}{22372} & \multicolumn{1}{r}{1.6}    & 92                             & 111                         & 854                             & \cellcolor{blue!10}\textbf{5312}           \bigstrut\\ \hlineB{2}
\multicolumn{1}{V{2}r}{\small \textsc{Deepstream}}   & \multicolumn{1}{r}{53}  & \multicolumn{1}{r}{19}                    & \multicolumn{1}{r}{43}        & \multicolumn{1}{r}{497} &   \multicolumn{1}{r}{3.1} & 86                              & 16                            & 32                            & \cellcolor{blue!10}\textbf{1509}                     \bigstrut\\ 

\multicolumn{1}{V{2}r}{}&  \multicolumn{1}{r}{53} & \multicolumn{1}{r}{288}                      & \multicolumn{1}{r}{219}       & \multicolumn{1}{r}{5008} & \multicolumn{1}{r}{2.3}    & 85                             & 97                         & 168                             & \cellcolor{blue!10}\textbf{3113}           \bigstrut\\ \hlineB{2}

\end{tabular}}\vspace{-1em}

\end{table}

\section{Scalability}
\label{sec:scalability}



\noindent{\textbf{Setting.}}~ We evaluated \ourapproach for scalability with \textsc{SQLite} (large configuration space) and \textsc{Deepstream} (large composed system). 
In \textsc{SQLite}, we evaluated three scenarios: (a)~selecting the most relevant software, hardware options, and events (34 configuration options and 19 system events), (b) selecting all modifiable software and hardware options and system events (242 configuration options and 19 events), and (c) selecting not only all modifiable software and hardware options and system events but also intermediate {\tt tracepoint events} (242 configuration options and 288 events). In \textsc{Deepstream}, there are two scenarios: (a) 53 configuration options and 19 system events, and (b) 53 configuration options and 288 events when we select all modifiable software and hardware options, and system/tracepoint events. 

\noindent{\textbf{Results.}} In large systems, there are significantly more causal paths and therefore, causal learning and estimations of queries take more time. 
The results in ~\tab{scalability} indicate that \ourapproach can scale to a much larger configuration space without an exponential increase in runtime for any of the intermediate stages. This can be attributed to the sparsity of the causal graph. For example, the average degree of a node for \textsc{SQLite} in \tab{scalability} is at most 3.6, and it reduces to 1.6 when the number of configurations increases. Similarly, the average degree reduces from 3.1 to 2.3 in \textsc{Deepstream} when systems events are increased. 

\section{Related Work}
\label{sec:related}
\noindent

\noindent\textbf{Performance faults in configurable systems.} Previous empirical studies have shown that a majority of performance issues are due to misconfigurations~\cite{han2016empirical}, with severe consequences in production environments~\cite{tang2015holistic,maurer2015fail}, and configuration options that cause such performance faults force the users to tune the systems themselves~\cite{zhang2021evolutionary}. Previous works have used static and dynamic program analysis to identify the influence of configuration options on performance~\cite{VJSSAK:ASE20,velez2021white,li2020statically} and to detect and diagnose misconfigurations~\cite{XJHZLJP:OSDI16,attariyan2010automating,zhang2013automated,attariyan2012x}. 
Unlike \ourapproach, none of the white-box analysis approaches target configuration space across the system stack, where it limits their applicability in identifying the true causes of a performance fault.

\noindent\textbf{Statistical and model-based debugging.}
Debugging approaches such as \textsc{Statistical Debugging}~\cite{song2014statistical}, 
\textsc{HOLMES}~\cite{chilimbi2009holmes}, \textsc{XTREE}~\cite{krishna2017less}, \bugdoc~\cite{lourencco2020bugdoc}, \encore~\cite{lourencco2020bugdoc}, \textsc{Rex}~\cite{mehta2020rex}, and \textsc{PerfLearner}~\cite{han2018perflearner} have been proposed to detect root causes of system faults. These methods make use of statistical diagnosis and pattern mining to rank the probable causes based on their likelihood of being the root causes of faults. However, these approaches may produce correlated predicates that lead to incorrect explanations.



\noindent\textbf{Causal testing and profiling.} 
Causal inference has been used for fault localization \cite{baah2010causal,feyzi2019inforence}, resource allocations in cloud systems~ \cite{geiger2016causal}, and causal effect estimation for advertisement recommendation systems \cite{bottou2013counterfactual}. More recently, \textsc{AID}~\cite{fariha2020causality} detects root causes of an intermittent software failure using fault injection as an intervention. \textsc{Causal Testing} ~\cite{johnson2020causal} modifies the system inputs to observe behavioral changes and utilizes counterfactual reasoning to find the root causes of bugs. Causal profiling approaches like \textsc{CoZ}~\cite{curtsinger2015coz} point to developers where optimizations will improve performance and quantify their potential impact. Causal inference methods like \textsc{X-Ray}~\cite{attariyan2012x} and \textsc{ConfAid}~\cite{attariyan2010automating} had previously been applied to analyze program failures. All approaches above are either orthogonal or complementary to \ourapproach, mostly they focus on functional bugs (e.g., \textsc{Causal Testing}) or if they are performance-related, they are not configuration-aware (e.g., \textsc{CoZ}).

\section{Limitations and Future Directions}
\label{sec:discussion}

\noindent \textbf{Learning a predictive model vs learning the underlying structure.}
Building a causal performance model could be more expensive than performance influence models. The reason for having a potentially higher learning cost is that in addition to learning a predictive model, we also need to learn the structure of the input configuration space.
However, exploiting causal knowledge is more helpful in search-like tasks (e.g., performance optimization~\cite{JC:MASCOTS16,JVKS:FSE18}) that looks for 
higher quality samples, making it possible to debug or optimize with a few samples.

\noindent \textbf{Dealing with an incomplete causal model.} Existing off-the-shelf causal graph discovery algorithms like FCI remain ambiguous while data is insufficient and returns partially directed edges. For highly configurable systems, gathering high-quality data is challenging. To address this issue, we develop a novel pipeline for causal model discovery by combining FCI with entropic causality, an information-theoretic approach~\cite{Kocaoglu2017} to causality that takes the direction across which the entropy is lower as the causal direction. Such an approach helps to reduce ambiguity and thus allows the causal graph to converge faster. Note that estimating a theoretical guarantee for convergence is out of scope, as having a global view of the entire configuration space is infeasible. Moreover,  the presence of too many confounders can affect the correctness of the causal models, and this error may propagate along with the structure if the dimensionality is high. Therefore, we use a greedy refinement strategy to update the causal graph incrementally with more samples; at each step, the resultant graph can be approximate and incomplete, but asymptotically, it will be refined to its correct form given enough time and samples.

\noindent \textbf{Algorithmic innovations for faster convergence.} The efficacy of \ourapproach depends on several factors such as the representativeness of the observational data or the presence of unmeasured confounders that can negatively affect the quality of the causal model. There are instances where the causal model may be incorrect or lack some crucial connections that may result in detecting spurious root causes or recommending incorrect repairs. One promising direction to address this problem would be to develop new algorithms for Stage II \& III of \ourapproach (see Section~\ref{sec:methodology}). Specifically, we see the potential for developing innovative approaches for learning better structure, incorporating domain knowledge by restricting the structure of the underlying causal model. In addition, there are potentials for developing better sampling algorithms by either shrinking the search space (e.g., using transfer learning~\cite{JVKS:FSE18}) or searching the space more efficiently to determine effective interventions that enable faster convergence to the true underlying structure. 

\noindent \textbf{Incorporating domain knowledge.} Additionally, there is scope for developing new approaches for either automatically extracting constraints (e.g., from source code or other downstream artifacts) to incorporate in learning causal performance model or approaches to make humans part of the loop for correcting the causal performance model during learning. Specifically, new approaches could provide infrastructure as well as algorithms to determine when to ask for human feedback and what to ask for, e.g., feedback regarding a specific part of the causal model or feedback regarding the determined intervention at each step.

\noindent \textbf{Developing new domain-specific languages.} \ourapproach uses a query engine to translate common user queries into counterfactual statements. A domain-specific language to facilitate automated specification of queries from written unstructured data could potentially lead to the adoption of causal reasoning in the system development lifecycle. 
\section{Conclusion}
\label{sect:conclusion}

Modern computer systems are highly-configurable with thousands of interacting configurations with a complex performance behavior. Misconfigurations in these systems can elicit complex interactions between software and hardware configuration options, resulting in non-functional faults. We propose \ourapproach, a novel approach for diagnostics that learns and exploits the system's causal structure consisting of configuration options, system events, and performance metrics. Our evaluation shows that \ourapproach effectively and quickly diagnoses the root cause of non-functional faults and recommends high-quality repairs to mitigate these faults. We also show that the learned causal performance model is transferable across different workload and deployment environments. Finally, we demonstrate the scalability of \ourapproach scales to large systems consisting of 500 options and several trillion potential configurations. 




\section*{Acknowledgements} This work has been supported in part by NASA (Awards 80NSSC20K1720 and 521418-SC) and NSF (Awards 2007202, 2107463, and 2038080), Google, and Chameleon Cloud. We are grateful to all who provided feedback on this work, including Christian K$\ddot{\text{a}} $stner, Sven Apel, Yuriy Brun, Emery Berger, Tianyin Xu, Vivek Nair, Jianhai Su, Miguel Velez, Tobius D$\ddot{\text{u}}$rschmid, and the anonymous EuroSys'22 (as well as EuroSys'21 and FSE'21) reviewers. 


\bibliographystyle{acm}
\balance
\bibliography{references,bib_pjamshidi}
\clearpage 
\balance
\appendix
.
%


\section{Artifact Appendix}
\label{sec:artifact}
\begin{tcolorbox}[colback=blue!5!white,colframe=blue!75!black]
\textbf{DOI:} \url{doi:10.5281/zenodo.6360540} \\
\textbf{Code:} \url{https://github.com/softsys4ai/unicorn}
\end{tcolorbox}

This appendix provides additional information regarding the tool that we have developed for evaluating \ourapproach. In this section, we call this tool \ourtool. In addition, we describe the steps using our \ourtool to reproduce the results reported in \S\ref{sec:effectiveness}, \S\ref{sec:transfer}, and \S\ref{sec:scalability}. We provide the source code and data in a publicly accessible GitHub repository that can be tested on any hardware once the software dependencies are met. 

\subsection{Description}
\ourapproach is used for performing tasks such as performance optimization and performance debugging in both offline and online modes. 
\begin{itemize}
    \item In the offline mode, \ourtool can be run on any device that uses previously measured configurations.
    \item In the online mode, the performance metrics are measured directly on the hardware on which the underlying configurable system is deployed, while the experiments are running. In the experiments, we have used \txtwo and \xavier. To collect measurements from these devices, \emph{sudo} privilege is needed, as it requires setting a device to a new configuration before measurement.
\end{itemize}
 


\subsection{Setup}








\subsubsection{Software Dependencies}
\begin{figure}[hb!]
    \centering
    \includegraphics[width=\linewidth]{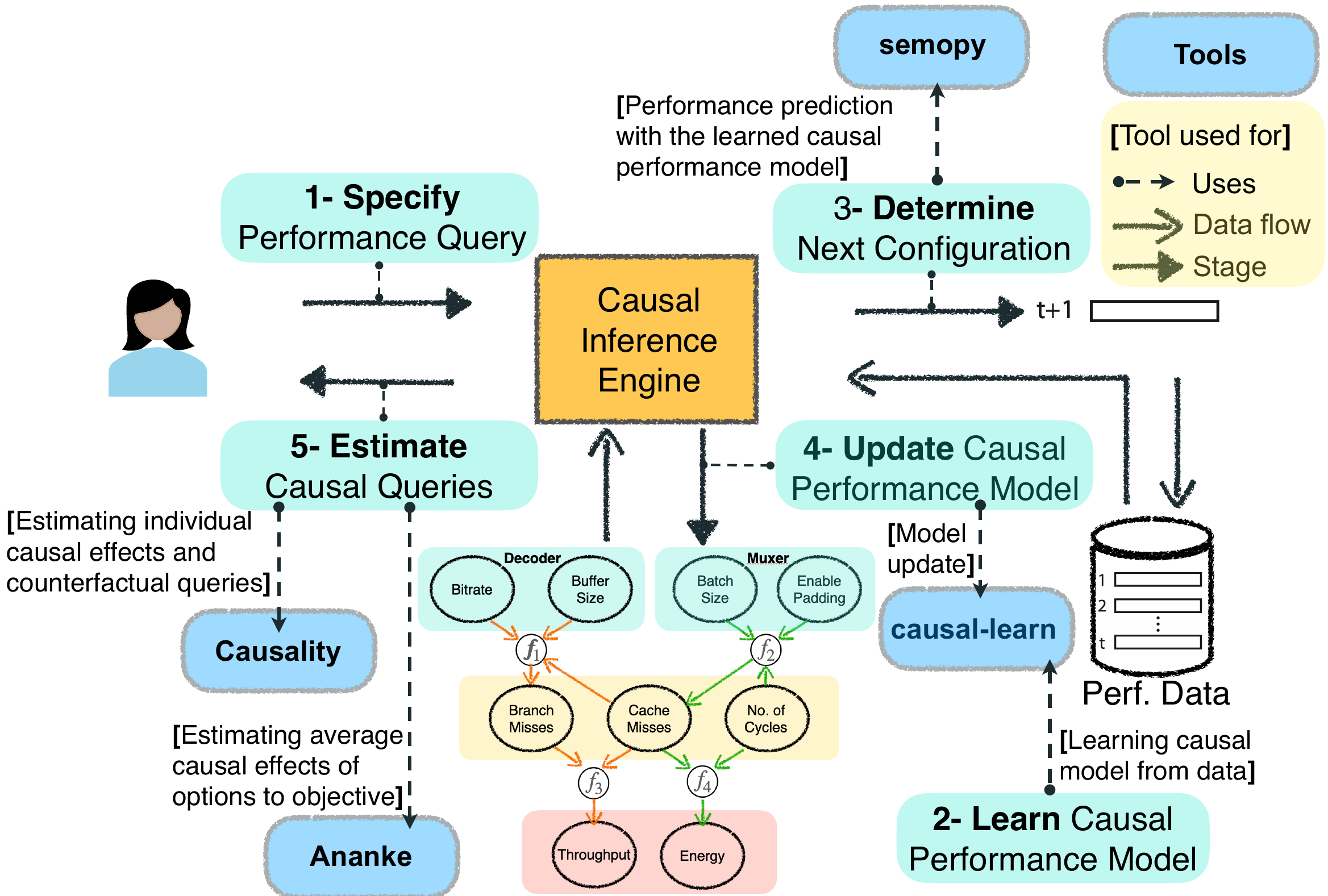}
    \caption{Toolchain in \ourtool.}
    \label{fig:unicorn-toolchain}
\end{figure}

\ourtool is implement-ed by integrating and building on top of several existing tools (see \fig{unicorn-toolchain}): 
\begin{itemize}
    \item \href{https://semopy.com/}{\color{blue!80}semopy} for predictions with causal models.
    \item \href{https://ananke.readthedocs.io/en/latest/}{\color{blue!80}ananke} and \href{https://github.com/akelleh/causality}{\color{blue!80}causality} for estimating the causal effects.
    \item \href{https://github.com/cmu-phil/causal-learn}{\color{blue!80}causal-learn} for structure learning. 
\end{itemize}

\subsubsection{Hardware Dependencies}
\ourtool is implemented both in offline and online modes. There are no particular hardware dependencies to run \ourtool in offline mode. To run \ourtool in online mode, we used hardware that has sensors for performance measurements. In particular, we used \txone, \txtwo, and \xavier with \emph{Jetpack 4.3} and \emph{Ubuntu 20.04 LTS}.

\subsubsection{Installation}
We use \textit{docker-compose} to install the necessary software required to run \ourtool. The necessary steps to install the dependencies and third-party libraries used to test our approach can be done with the following commands.

\begin{verbatim}
git clone git@github.com:softsys4ai/unicorn.git
cd unicorn
docker-compose up --build --detach
\end{verbatim}

Once this step is completed, \ourtool is ready to be tested.

\subsection{Data} 
All the datasets required to run experiments are already included in the \textit{./unicorn/data} directory.  

\subsection{Major Claims}
We make the following major claims in our paper:
\begin{itemize}
\item \ourapproach can be used to detect root causes of non-functional performance (latency and energy) faults with higher accuracy and gain.

\item \ourapproach can support performing downstream performance tasks such as performance optimization.

\item The causal performance models are transferable across environments (different workload or hardware) and can be efficiently re-used from the source environment where it is trained to a target environment.
\end{itemize}

\subsection{Experiments}
We run the following experiments to support our claims.

\subsubsection{E1: Performance Debugging Experiment} To support the claim of efficiency of \ourapproach in debugging non-functional faults, we reproduce energy faults results for \textsc{Xception} in \textsc{Nvidia Jeston Xavier} from Table~\ref{tab:single_1}. Our initial study discovered 29 energy faults for \textsc{Xception} in \textsc{Nvidia Jetson Xavier,} that is 12\% of the faults reported in Table~\ref{tab:single_1}. This would require ~1.5 hours to run the experiments in offline mode and ~11 hours to run the experiments in online mode.

\noindent\textbf{Execution}. To run \ourtool on a single bug, execute the following command:

\begin{verbatim}
docker-compose exec unicorn python \\ 
./tests/run_unicorn_debug.py -o \\
total_energy_consumption -s Image -k Xavier \\
-m offline\online -i 0
\end{verbatim}

To run \ourtool and other debugging baselines reported in this paper on all the bugs, please use the following commands one by one:
\begin{verbatim}
docker-compose exec unicorn python \\ 
./tests/run_unicorn_debug.py -o \\
total_energy_consumption -s Image -k Xavier \\
-m offline\online 
    
docker-compose exec unicorn python \\ 
./tests/run_baseline_debug.py -o \\
total_energy_consumption -s Image -k Xavier \\
-m offline\online -b cbi 
    
docker-compose exec unicorn python \\ 
./tests/run_baseline_debug.py -o \\
total_energy_consumption -s Image -k Xavier \\
-m offline\online -b encore
    
docker-compose exec unicorn python \\ 
./tests/run_baseline_debug.py -o \\
total_energy_consumption -s Image -k Xavier \\
-m offline\online -b bugdoc
    
\end{verbatim}

\noindent \textbf{Results}. We save the evaluation metrics such as accuracy, precision, recall, gain, and time required for debugging. A separate plot is generated using the recommended fixes to compare \ourapproach with other baseline approaches with their evaluation metrics. Note, in the offline mode the reported time is different (usually less) from the main text as instead of running the measurements online we reuse recorded measurements. However, we can get a sense of the efficiency by comparing the number of samples required to resolve a fault.

\subsubsection{E2: Performance Optimization Experiment} \ourapproach supports can support performing downstream performance tasks such as performance optimization. To support this claim, we reproduce single-objective latency optimization results reported in \fig{rq1_opt_se} (a). This experiment would require around ~1.5 hours to complete in the offline mode and ~4 hours to complete in the online mode. We also compare the results with a baseline optimization approach, \textsc{SMAC}, reported in the paper.

\noindent \textbf{Execution}. To run the experiment, we need to execute the following commands:

\begin{verbatim}
docker-compose exec unicorn python \\ 
./tests/run_unicorn_optimization.py -o \\
inference_time -s Image -k TX2 \\
-m offline\online 
    
docker-compose exec unicorn python \\ 
./tests/run_baseline_optimization.py -o \\
inference_time -s Image -k TX2 \\
-m offline\online -b smac
\end{verbatim}

\noindent \textbf{Results}. We display the results similar to \fig{rq1_opt_se} (a) using a line plot. Note that this experiment is run once without repeating, so there are no error bars.

\subsubsection{E3: Transferability Experiment.} To support this claim, we initially build a causal performance model to resolve the latency faults in \xavier and reuse the causal performance model to resolve the latency faults in \txtwo. We only use one bug to demonstrate this result. This would require ~10 minutes to run the experiment in the offline mode and ~25 minutes in the online mode. 

\noindent \textbf{Execution}. The following command runs the experiments:

\begin{verbatim}
docker-compose exec unicorn python \\ 
./tests/run_unicorn_transferability.py -o \\
inference_time -s Image -k Xavier \\
-m offline\online 

\end{verbatim}

\noindent \textbf{Results}. The evaluation metrics, including accuracy, precision, recall, gain, and time required for debugging for different scenarios reported in the paper are saved to a separate CSV file after the experiments are over and plotted. Note that the reported time is different from the time reported in the main text in the offline mode. 

\subsection{Using \ourtool with external data}
We added \href{https://github.com/softsys4ai/unicorn/blob/master/artifact/OTHERS.md}{\color{blue!80}instructions} to describe the required steps to use \ourtool with any other external dataset.

\subsection{Extending \ourtool}
We welcome any contribution for extending either \ourapproach (see \S\ref{sec:discussion} for several possible future directions) and \ourtool for performance improvements or feature extensions.

\clearpage
\section{Appendix}
\label{sec:appendix}
\subsection{Causal Performance Modeling and Analyses: Motivating Scenarios (Additional details)}

\begin{figure}[h]
\small
    \centering
    \includegraphics*[width=0.5\linewidth]{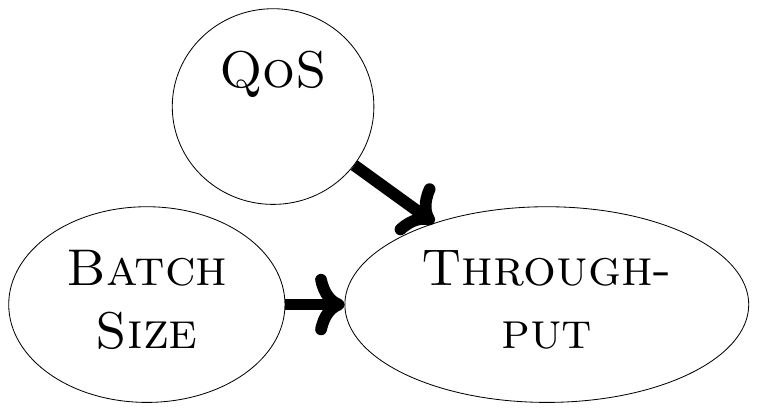}
    \caption{\small {Performance influence model incorrectly identifies \texttt{Batch Size} and \texttt{QoS} are positively correlated with the term $\texttt{0.08 Batch Size} \times \texttt{QoS}$ whereas they are unconditionally independent. Causal model correctly identifies the dependence (no causal connection) relationship between \texttt{Batch Size} and \texttt{QoS} (no arrow between \texttt{Batch Size} and \texttt{QoS})}.}
    \label{fig:spurious_two}
    
\end{figure}

\begin{figure}[h]
\small
    \centering
    \includegraphics*[width=0.7\linewidth]{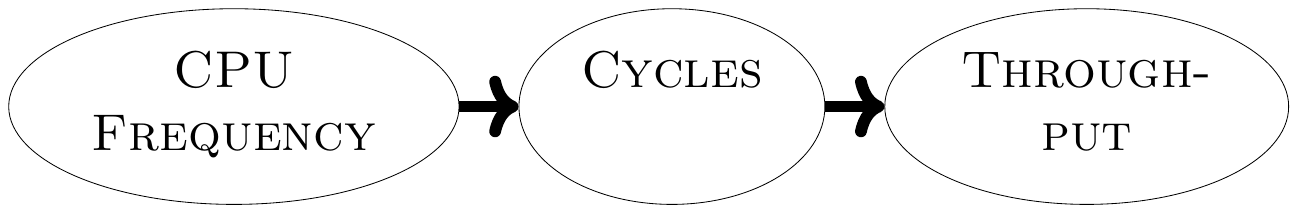}
    \caption{\small {Causal model correctly identifies how \texttt{CPU Frequency} causally influences \texttt{Throughput} via \texttt{Cycles} whereas the performance influence model $\texttt{Throughput} = 0.05 \times \texttt{CPU Frequency} \times \texttt{Cycles}$ identified incorrect interactions. }}
    \label{fig:spurious_three}
    
\end{figure}

\fig{spurious_two} and ~\fig{spurious_three} present additional scenarios where performance influence models could produce incorrect explanations. The regression terms presented here incorrectly identify spurious correlations, whereas the causal model correctly identifies the cause-effect relationships.

\begin{figure}[h]
\small
    \centering
    \includegraphics*[width=\linewidth]{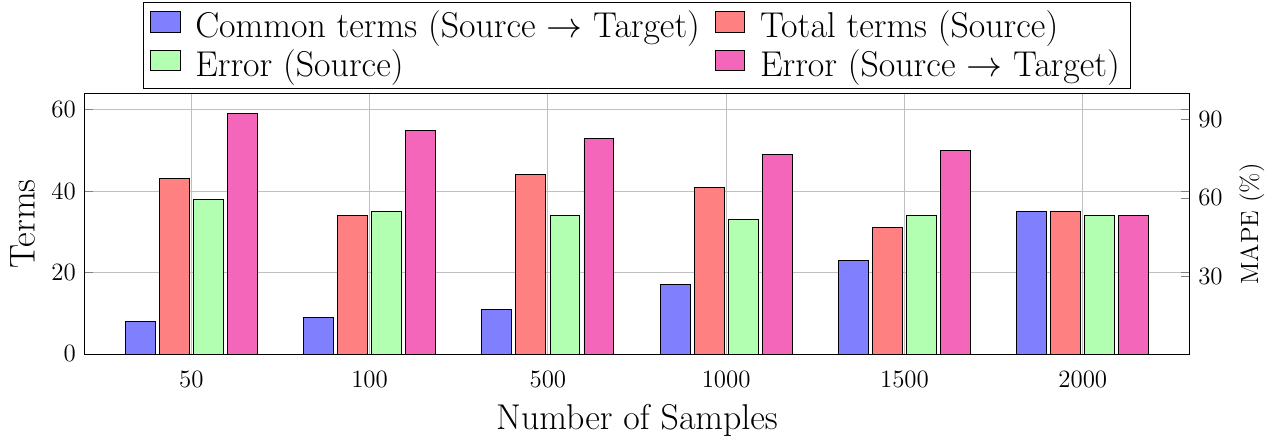}
    \caption{\small {Performance influence models relying on correlational statistics are not stable as new samples are added and do not generalize well. Common terms refers to the individual predictors (i.e., options and interactions) in the performance models that are similar across envirnments.}}
    \label{fig:reg_eq_noise}
    
\end{figure}

\begin{figure}[h]
\small
    \centering
    \includegraphics*[width=\linewidth]{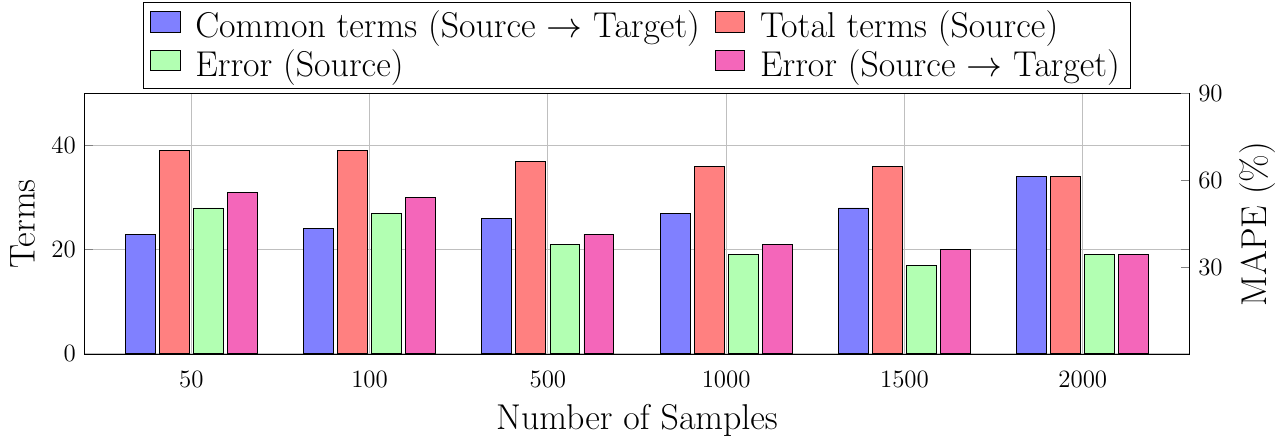}
    \caption{\small {Causal performance models are relatively more stable as new samples are added and do generalize well.}}
    \label{fig:reg_eq_noise_cpm}
    
\end{figure}

The performance behavior of regression models for configurable systems varies when sample size varies. ~\fig{reg_eq_noise} shows the change of a number of stable terms and error with different numbers of samples used for building a performance influence model. Here, we vary the number of samples from 50 to 1500 to build a source regression model. We use a sample size of 2000 to build the target regression model. We observe that regression models cannot be reliably used in performance tasks, as they are sensitive to the number of training samples. The results indicate that these model classes as opposed to causal models cannot identify causal variables underlying system performance, so depending on the training sample, they try to find the best predictor to increase the prediction power with the i.i.d. assumption that does not hold in system performance. On the contrary, the number of stable predictor's variation is less in causal performance models and leads to better generalization, as shown in ~\fig{reg_eq_noise_cpm}. In addition to the number of stable predictors, the difference in error between source and target is negligible when compared to the performance regression models. 


\noindent \textbf{Extraction of predictor terms from the causal performance model.} The constructed causal performance models have performance objective nodes at the bottom (leaf nodes) and configuration options nodes at the top level. The intermediate levels are filled with the system events. To extract a causal term from the causal model, we backtrack starting from the performance objective until we reach a configuration option. If there is more than one path through a system event from performance objective to configuration options, we consider all possible interactions between those configuration options to calculate the number of causal terms.

\subsection{\ourapproach (Additional details)}

Note, if $X$ is a continuous variable, we would replace the summation of $ACE$ with an integral. 
%
For the entire path, we extend it as:
\smalleq
\label{eq:path_ace}
{\footnotesize
\setlength{\abovedisplayskip}{1pt}
\setlength{\belowdisplayskip}{1pt}
\mathrm{Path}_{ACE} = \frac{1}{K} \cdot \sum \mathrm{ACE}(Z, X) \hspace{2em} \footnotesize \forall X, Z \in path 
}
\eeq
\eq{path_ace} represents the average causal effect of the causal path. The configuration options that lie in paths with larger $P_{ACE}$ tend to have a greater causal effect on the corresponding non-functional properties in those paths. We select the top $K$ paths with the largest $\mathrm{P}_{ACE}$ values, for each non-functional property. In this paper, we use K=3 to 25, however, this may be modified in our replication package. 


\begin{table}[]
\centering
\caption{Mapping between configuration options and options indexes. Only a subset of configuration options are shown here.}
\resizebox{\linewidth}{!}{
\begin{tabular}{llll}
\hline
Option&Configuration& Option& Configuration\\
Index & Options & Index & Options\\
\hline
0& \texttt{Swap Memory} &14& \texttt{kernel.numa\_balancing}  \\
1& \texttt{Scheduler Policy}&15& \texttt{kernel.sched\_latency\_ns} \\
2&\texttt{Drop Caches}&16& \texttt{kernel.sched\_nr\_migrate} \\
3& \texttt{Batch Size}& 17& \texttt{kernel.sched\_rt\_period\_us} \\
4& \texttt{Bitrate}&18& \texttt{kernel.sched\_rt\_runtime\_us}\\
5& \texttt{Buffer Size} &19& \texttt{kernel.sched\_time\_avg\_ms}\\
6& \texttt{CPU Freqeuncy}&20& \texttt{kernel.sched\_child\_runs\_first}\\
7& \texttt{GPU Frequency}& 21& \texttt{vm.vfs\_cache\_pressure} \\
8& \texttt{EMC Frequency}& 22& \texttt{vm.swappiness}\\
9& \texttt{CPU Cores}& 23& \texttt{Enable Padding}\\
10& \texttt{vm.overcommit\_memory}& 24& \texttt{vm.dirty\_background\_ratio}\\
11& \texttt{vm.overcommit\_hugepages}& 25& \texttt{vm.dirty\_background\_bytes}\\
12& \texttt{kernel.cpu\_time\_max\_percent}& 26& \texttt{vm.dirty\_ratio}\\
13& \texttt{kernel.max\_pids}& 27& \texttt{vm.nr\_hugepages}\\
\hline
\end{tabular}}
\label{tab:mapping}
\end{table}

\begin{table}[]
\centering
\caption{Configuration options in \textsc{Xception}, \textsc{Bert}, and \textsc{Deepspeech}.}
\begin{tabular}{lll}
\hline
Configuration Options & Option Values/Range  \\
\hline 
\texttt{Memory Growth}                      &    -1, 0.5, 0.9            \\ 
\texttt{Logical Devices}                      &  0, 1               \\ 
\hline
\end{tabular}
\label{tab:ml_conf}
\end{table}

\begin{table}[]
\centering
\caption{\textsc{x264} software configuration options.}
\begin{tabular}{lll}
\hline
Configuration Options & Option Values/Range  \\
\hline 
\texttt{CRF} &  13, 18, 24, 30               \\  
\texttt{Bit Rate} & 1000, 2000, 2800, 5000         \\  
\texttt{Buffer Size} &  6000, 8000, 20000          \\  
\texttt{Presets} &   ultrafast, veryfast, faster\\
& medium, slower  \\  
\texttt{Maximum Rate} &       600k, 1000k         \\  
\texttt{Refresh} &       OFF, ON         \\  
\hline
\end{tabular}
\label{tab:x264_conf}
\end{table}

\begin{table}[]
\centering
\caption{\textsc{SQLite} software configuration options.}
\resizebox{\linewidth}{!}{\begin{tabular}{lll}
\hline
Configuration Options & Option Values/Range \\
\hline 
\texttt{PRAGMA TEMP\_STORE}           &   DEFAULT, FILE, MEMORY              \\  
\texttt{PRAGMA JOURNAL\_MODE}           &      DELETE, TRUNCATE,PERSIST,MEMORY, OFF          \\  
\texttt{PRAGMA SYNCHRONOUS}           &    FULL, NORMAL, OFF          \\  
\texttt{PRAGMA LOCKING\_MODE}           &     NORMAL, EXCLUSIVE            \\  
\texttt{PRAGMA CACHE\_SIZE}           &    0, 1000, 2000, 4000, 10000             \\ 
\texttt{PRAGMA PAGE\_SIZE}           &   2048, 4096, 8192         \\ 
\texttt{PRAGMA MAX\_PAGE\_COUNT}           &   32, 64            \\ 
\texttt{PRAGMA MMAP\_SIZE}           &  30000000000, 60000000000,\\
\hline
\end{tabular}}
\label{tab:sqlite_conf}
\end{table}

\begin{table}[]
\centering
\caption{Linux OS/Kernel configuration options.}
\resizebox{\linewidth}{!}{
\begin{tabular}{lll}
\hline
Configuration Options & Option Values/Range  \\
\hline 
\texttt{vm.vfs\_cache\_pressure}& 1, 100, 500                 \\  
\texttt{vm.swappiness} &  10, 60, 90\\
\texttt{vm.dirty\_bytes} &30, 60\\
\texttt{vm.dirty\_background\_ratio}&10, 80\\
\texttt{vm.dirty\_background\_bytes}&30, 60\\
\texttt{vm.dirty\_ratio}&5, 50\\
\texttt{vm.nr\_hugepages}&0, 1, 2\\
\texttt{vm.overcommit\_ratio}&50, 80\\
\texttt{vm.overcommit\_memory}&0, 2\\
\texttt{vm.overcommit\_hugepages}&0, 1, 2\\
\texttt{kernel.cpu\_time\_max\_percent}&10 - 100\\
\texttt{kernel.max\_pids}&32768, 65536\\
\texttt{kernel.numa\_balancing} &0, 1\\
\texttt{kernel.sched\_latency\_ns} &24000000, 48000000\\
\texttt{kernel.sched\_nr\_migrate} &32, 64, 128\\
\texttt{kernel.sched\_rt\_period\_us} &1000000, 2000000&\\
\texttt{kernel.sched\_rt\_runtime\_us} &500000, 950000\\
\texttt{kernel.sched\_time\_avg\_ms} &1000, 2000\\
\texttt{kernel.sched\_child\_runs\_first} &0, 1&\\
\texttt{Swap Memory}&1, 2, 3, 4 (GB)\\
\texttt{Scheduler Policy}&CFP, NOOP\\
\texttt{Drop Caches}&0, 1, 2, 3\\

\hline
\end{tabular}}
\label{tab:os_conf}
\end{table}

\begin{table}[]
\centering
\caption{Hardware configuration options.}
\begin{tabular}{lll}
\hline
Configuration Options & Option Values/Range \\
\hline 
\texttt{CPU Cores}           &   1 - 4              \\
\texttt{CPU Frequency}           &  0.3 - 2.0 (GHz)              \\
\texttt{GPU Frequency}           &  0.1 - 1.3 (GHz)               \\
\texttt{EMC Frequency}           &  0.1 - 1.8 (GHz)               \\
\hline
\end{tabular}
\label{tab:hw_conf}
\end{table}

\begin{table}[]
\centering
\caption{Performance system events and tracepoints. }
\begin{tabular}{lll}
\hline
System Events \\
\hline 
\texttt{Context Switches}\\ 
\texttt{Major Faults}         \\ 
\texttt{Minor Faults }                              \\ 
\texttt{Migrations}                           \\ 
\texttt{Scheduler Wait Time}                             \\ 
\texttt{Scheduler Sleep Time}                             \\ 
\texttt{Cycles}                               \\
\texttt{Instructions}                             \\ 
\texttt{Number of Syscall Enter}                       \\ 
\texttt{Number of Syscall Exit}                       \\ 
\texttt{L1 dcache Load Misses}            \\                   
\texttt{L1 dcache Loads}                               \\
\texttt{L1 dcache Stores}                               \\
\texttt{Branch Loads}                               \\
\texttt{Branch Loads Misses}                            \\
\texttt{Branch Misses}                            \\
\texttt{Cache References}\\
\texttt{Cache Misses}\\
\texttt{Emulation Faults}                               \\
\hline
Tracepoint Subsystems\\
\hline
\texttt{Block}\\
\texttt{Scheduler}\\
\texttt{IRQ}\\
\texttt{ext4}\\
\hline
\end{tabular}
\label{tab:event_conf}
\end{table}

Counterfactual queries can be different for different tasks. For debugging, we use the top $K$ paths to (a)~identify the root cause of non-functional faults; and (b)~prescribe ways to fix the non-functional faults. Similarly, we use the top $K$ paths to identify the options that can improve the non-functional property values near-optimal.
For both tasks, a developer may ask specific queries to \ourapproach and expect an actionable response. For debugging, we use the example causal graph of where a developer observes low FPS and high energy, \ie, a multi-objective fault, and has the following questions:  

\noindent\textbf{\faQuestionCircle~\textbf{``What are the root causes of my multi-objective (\texttt{FPS} and \texttt{Energy}) fault?''}} To identify the root cause of a non-functional fault, we must identify which configuration options have the most causal effect on the performance objective. 
For this, we use the steps outlined in~\tion{path_discovery} to extract the paths from the causal graph and rank the paths based on their average causal effect (\ie, $\mathrm{Path}_{ACE}$ from \eq{path_ace}) on latency and energy. We return the configurations that lie on the top $K$ paths. 
For example, in ~\fig{causal_model_example} we may return (say) the following paths: 
\bisq
\small
    \item  \texttt{Batch Size} \edgeone \texttt{Cache Misses} \edgeone \texttt{FPS} and \texttt{Energy}
    \item  \texttt{Enable Padding}  \edgeone \texttt{Cache Misses}  \edgeone \texttt{FPS} and \texttt{Energy}
\ei
 
and the configuration options {\texttt{BatchSize}, and \texttt{Enable Padding}} being the probable root causes.

\noindent\textbf{\faQuestionCircle~\textbf{``How to improve my \texttt{FPS} and \texttt{Energy}?''}} To answer this query, we first find the root causes as described above. Next, we discover what values each of the configuration options must take in order that the new \texttt{FPS} and \texttt{Energy} is better (high \texttt{FPS} and low \texttt{Energy}) than the fault (low \texttt{FPS} and high \texttt{Energy}). For example, we consider the causal path \texttt{Batch Size}~\edgeone \texttt{Cache Misses} \edgeone \texttt{FPS} and \texttt{Energy}, we identify the permitted values for the configuration options {\texttt{Batch Size}} that can result in a high FPS and energy ($Y^{\mathit{\textsc{low}}}$) that is better than the fault ($Y^{\mathit{\textsc{high}}}$).
For this, we formulate the following counterfactual expression: 
\smalleq
\label{eq:cfact_bare}
\footnotesize
\mathrm{Pr}(Y_{repair}^{\textsc{low}}|\neg repair,Y_{\neg repair}^{\textsc{high}})
\eeq
\eq{cfact_bare} measures the probability of ``fixing'' the latency fault with a ``repair'' {\footnotesize $(Y_{repair}^{\textsc{low}})$} given that with no repair {we observed the fault} {\footnotesize $(Y_{\neg repair}^{\text{\textsc{high}}})$}.   
In our example, the repairs would resemble \texttt{Batch Size}=$10$. We generate a \textit{repair set} ($\mathcal{R}_{1}$), where the configurations \texttt{Batch Size} is set to all permissible values, \ie,
{
\small
\begin{multline}\label{eq:repairs}
    \setlength{\abovedisplayskip}{5pt}
    \setlength{\belowdisplayskip}{5pt}
    \mathcal{R}_{1}\equiv~\bigcup~\left\{\texttt{Batch Size} = {x},... \right\}\forall {x} \in \texttt{Batch Size}
\end{multline}}
observe that, in the repair set ($\mathcal{R}_{1}$) a configuration option that is not on the path \texttt{Batch Size}~\edgeone \texttt{Cache Misses} \edgeone \texttt{FPS} and \texttt{Energy} is set to the same value of the fault. For example, \texttt{Bit Rate} is set to $2$ or \texttt{Enable Padding} is set to $1$. This way we can reason about the effect of interactions between \texttt{Batch Size} with other options, i.e., \texttt{Bit Rate}, \texttt{Buffer Size}. Say \texttt{Buffer Size} or \texttt{Enable padding} were changed/recommended to set at any other value than the fault in some previous iteration, i.e., $20$ or $0$, respectively. In that case, we set \texttt{BufferSize} and \texttt{Enable padding}=$0$. Similarly, we generate a repair set $\mathcal{R}_{2}$ by setting \texttt{Enable Padding}to all permissible values. 
{
\small
\begin{multline}\label{eq:repairs}
    \setlength{\abovedisplayskip}{5pt}
    \setlength{\belowdisplayskip}{5pt}
    \mathcal{R}_{2}\equiv~\bigcup~\left\{\texttt{Enable padding} = {x},... \right\} \forall {x} \in \texttt{Enable padding}
\end{multline}}

Now, we combine the repair set for each path to construct a final repair set $\mathcal{R}=\mathcal{R}_{1} \cup~\ldots \cup\mathcal{R}_{k}$. Next, we compute the \textit{Individual Causal Effect} (ICE) on the \texttt{FPS} and \texttt{Energy} ($Y$) for each repair in the repair set $\mathcal{R}$. In our case, for each repair $\mathit{r}~\in~\mathcal{R}$, ICE is given by:
\begin{equation}
    \label{eq:ite}
    \footnotesize
    \mathrm{ICE}(\mathit{r})=\mathrm{Pr}(Y_r^{\textsc{low}}~|~\neg r,~Y_{\neg r}^{\textsc{high}}) - \mathrm{Pr}(Y_r^{\textsc{high}}~|~\neg r,~Y_{\neg r}^{\textsc{high}})\hspace{1em}
\end{equation}
ICE measures the difference between the probability that \texttt{FPS} and \texttt{Energy} \textit{is low} after a repair $r$ and the probability that the \texttt{FPS} and \texttt{Energy} is \textit{still high} after a repair $r$. If this difference is positive, then the repair has a higher chance of fixing the fault. In contrast, if the difference is negative, then that repair will likely worsen both \texttt{FPS} and \texttt{Energy}. To find the most useful repair ($\mathcal{R}_{\mathit{best}}$), we find a repair with the largest (positive) ICE, \ie, $\mathcal{R}_{\mathit{best}} = \argmax_{\forall r~\in~\mathcal{R}}[\mathrm{ICE}(\mathit{r})]$. This provides the developer with a possible repair for the configuration options that can fix the multi-objective \texttt{FPS} and \texttt{Energy} fault.

\noindent \textbf{Remarks.}~The ICE computation of \eq{ite} occurs \textit{only} on the observational data. Therefore, we may generate any number of repairs and reason about them without having to deploy those interventions and measure their performance in the real world. This offers significant runtime benefits.

\subsection{Evaluation (Additional details)}

\subsubsection{Experimental setup}

\label{subsec:experimental_setup}
\begin{table}[]
\centering
\caption{Deepstream software configuration options.}
\resizebox{\linewidth}{!}{
\begin{tabular}{llll}
\hline
Component&Configuration Options & Option Values/Range  \\
\hline 
&\texttt{CRF} &  13, 18, 24, 30               \\  
&\texttt{Bitrate} & 1000, 2000, 2800, 5000         \\  
&\texttt{Buffer Size} &  6000, 8000, 20000          \\  
Decoder&\texttt{Presets} &   ultrafast, veryfast, faster\\
&& medium, slower  \\  
&\texttt{Maximum Rate} &       600k, 1000k         \\  
&\texttt{Refresh} &       OFF, ON         \\  
\hline
&\texttt{Batch Size} &       0 - 30         \\  
&\texttt{Batched Push Timeout}&  0 - 20           \\  
&\texttt{Num Surfaces per Frame} & 1, 2, 3, 4\\
Stream Mux&\texttt{Enable Padding} & 0, 1          \\  
&\texttt{Buffer Pool Size} &  1 - 26               \\  
&\texttt{Sync Inputs} &  0, 1          \\ 
&\texttt{Nvbuf Memory Type} &     0, 1, 2, 3         \\ 
\hline
&\texttt{Net Scale Factor} &  0.01 - 10              \\  
&\texttt{Batch Size}& 1 - 60                \\  
&\texttt{Interval} &  1 - 20               \\  
&\texttt{Offset} &       0, 1          \\  
Nvinfer&\texttt{Process Mode} &   0, 1              \\ 
&\texttt{Use DLA Core} &      0, 1           \\ 
&\texttt{Enable DLA} &           0, 1      \\ 
&\texttt{Enable DBSCAN} &       0, 1       \\ 
&\texttt{Secondary Reinfer Interval} &       0 - 20          \\ 
&\texttt{Maintain Aspect Ratio} &       0, 1          \\ 
\hline
&\texttt{IOU Threshold} &       0 - 60          \\
&\texttt{Enable Batch Process} &       0, 1          \\
Nvtracker&\texttt{Enable Past Frame} &       0, 1          \\
&\texttt{Compute HW} &       0, 1, 2, 3, 4          \\ 
\hline
\end{tabular}}
\label{tab:deepstream_conf}
\end{table}

We used the following four components for Deepstream implementation:

\begin{itemize}
\item \textbf{Decoder}: For the decoder, we use x264. It uses the x264 and takes the encoded H.64, VP8, VP9 streams, and produces an NV12 stream.

\item \textbf{Stream Mux}:
The streammux module takes the NV12 stream and outputs the NV12 batched buffer with information about input frames, including the original timestamp and frame number.

\item \textbf{Nvinfer}:
For object detection and classification, we use the TrafficCamNet model that uses ResNet 18 architecture. This model is pre-trained in 4 classes on a dataset of 150k frames and has an accuracy of 83.5\% for detecting and tracking cars from a traffic camera's viewpoint. The 4 classes are Vehicle, BiCycle, Person, and Roadsign. We use the Keras (Tensorflow backend) pre-trained model from TensorRT.

\item \textbf{Nvtracker}:
The plugin accepts NV12- or RGBA-formated frame data from the upstream component and scales (converts) the input buffer to a buffer in the format required by the low-level library, with tracker width and height. NvDCF tracker uses a correlation filter-based online discriminative learning algorithm as a visual object tracker while using a data association algorithm for multi-object tracking. 
\end{itemize}

\noindent \textbf{Configuration options, events, and hyperparameters used for evaluation.}
Table~\ref{tab:deepstream_conf}, Table~\ref{tab:ml_conf}, Table~\ref{tab:x264_conf}, and Table~\ref{tab:sqlite_conf}, show different software configuration options and their values for different systems considered in this paper. Table~\ref{tab:os_conf} shows the OS/kernel level configuration options and their values for different systems considered in this paper. Additionally, Table~\ref{tab:event_conf}
shows the performance events considered in this paper. The hyperparameters considered for \textsc{Xception}, \textsc{Bert}, and \textsc{Deepspeech} are shown in Table~\ref{tab:dnn_conf}.

\begin{table}[]
\centering
\caption{Hyperparameters for DNNs used in \ourapproach.}
\resizebox{\linewidth}{!}{\begin{tabular}{lll}
\hline
Architecture & Hyperparameters & Option Values\\
\hline 
&\texttt{Number of Filters Entry} flow &32 \\
&\texttt{Filter Size Entry Flow} &(3 $\times$ 3)\\
&\texttt{Number of Filters Middle Flow} &64 \\
&\texttt{Filter Size Middle Flow} &(3 $\times$ 3)\\
\textsc{Xception}&\texttt{Number of Filters Exit Flow} &728\\
&\texttt{Filter Size Exit Flow} &(3 $\times$ 3)\\
&\texttt{Batch Size} &32\\
&\texttt{Number of Epochs}&100\\
\hline
&\texttt{Dropout} & 0.3\\
&\texttt{Maximum Batch Size}&16\\
\textsc{Bert}&\texttt{Maximum Sequence Length}&13\\
&\texttt{Learning Rate} &$1e^{-4}$\\
&\texttt{Weight Decay} &0.3\\
\hline
&\texttt{Dropout} & 0.3\\
&\texttt{Maximum Batch Size}& 16\\
\textsc{Deepspeech}&\texttt{Maximum Sequence Length}&32\\
&\texttt{Learning Rate} &$1e^{-4}$\\
&\texttt{Number of Epochs}&10\\
\hline
\end{tabular}}
\label{tab:dnn_conf}
\end{table}


\begin{table}[]
\centering
\caption{Hyperparameters for FCI used in \ourapproach.}
\begin{tabular}{lll}
\hline
Hyperparameters & Value \\
\hline 
\texttt{depth} &-1 \\
\texttt{testId} &fisher-z-test\\
\texttt{maxPathLength} &-1 \\
\texttt{completeRuleSetUsed}&False\\
\hline 

\end{tabular}
\end{table}

\subsubsection{Case Study.}
\begin{figure}[t!]
\small
    \centering
    \includegraphics*[width=\linewidth]{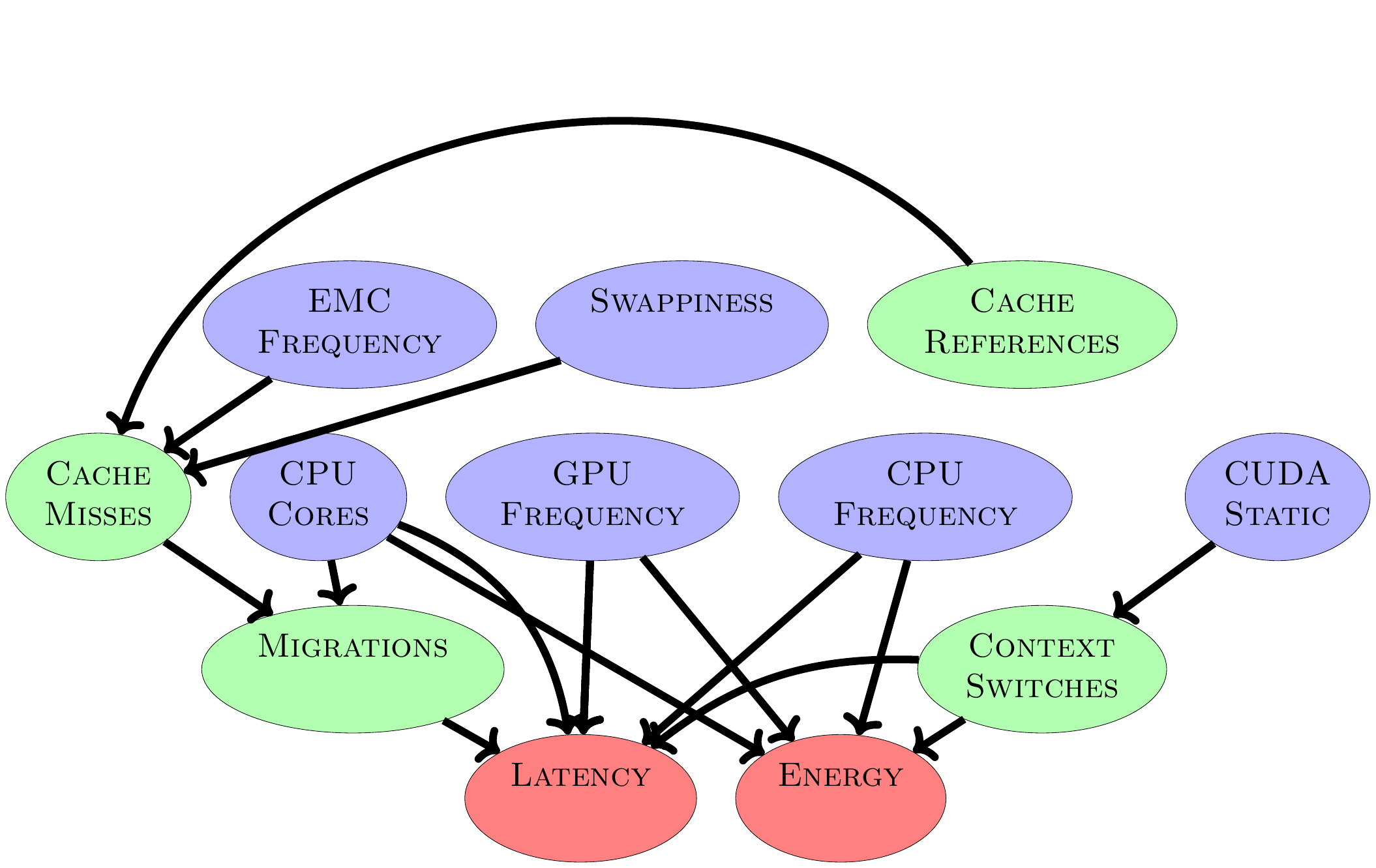}
    \caption{\small {Causal graph used to resolve the latency fault in the real world case study in section~\ref{sec:casestudy}}}
    \label{fig:real_wrold_cpm}
    
\end{figure}

~\fig{real_wrold_cpm} shows the causal graph to resolve the real-world latency fault.

\begin{table*}[h]
    \centering
    \caption{\small  Efficiency of \ourapproach compared to other approaches. Cells highlighted in \colorbox{blue!10}{\bfseries blue} indicate improvement over faults and \colorbox[HTML]{FFCCC9}{red} indicate deterioration. \ourapproach achieves better performance overall and is much faster.
    }
\vspace{-0.85em}
\subfloat[Single objective performance fault for \textit{heat} in \txone.]{\scriptsize
    \label{tab:rq1_1}
    \resizebox{\textwidth}{!}{
    \begin{tabular}{@{}l|l|l|lllll|lllll|lllll|lllll|ll|}
        \clineB{4-25}{2}
        \multicolumn{1}{c}{}&\multicolumn{1}{c}{}  &  & \multicolumn{5}{c|}{Accuracy} & \multicolumn{5}{c|}{Precision} & \multicolumn{5}{c|}{Recall} & \multicolumn{5}{c|}{Gain} & \multicolumn{2}{c|}{Time$^\dagger$} \bigstrut\\ \clineB{4-25}{2}
        \multicolumn{1}{c}{}& \multicolumn{1}{c}{} &  & \multicolumn{1}{c}{\rotatebox{90}{\bfseries \ourapproach}} &
        \multicolumn{1}{c}{\rotatebox{90}{\cbi}} & \multicolumn{1}{c}{\rotatebox{90}{DD}} & \multicolumn{1}{c}{\rotatebox{90}{\encore}} & \multicolumn{1}{c|}{\rotatebox{90}{\bugdoc~}} & \multicolumn{1}{c}{\rotatebox{90}{\bfseries \ourapproach}} &  
        \multicolumn{1}{c}{\rotatebox{90}{\cbi}} & \multicolumn{1}{c}{\rotatebox{90}{DD}} & \multicolumn{1}{c}{\rotatebox{90}{\encore}} & \multicolumn{1}{c|}{\rotatebox{90}{\bugdoc~}} & \multicolumn{1}{c}{\rotatebox{90}{\bfseries \ourapproach}} &
        \multicolumn{1}{c}{\rotatebox{90}{\cbi}} & \multicolumn{1}{c}{\rotatebox{90}{DD}} & \multicolumn{1}{c}{\rotatebox{90}{\encore}} & \multicolumn{1}{c|}{\rotatebox{90}{\bugdoc~}} & \multicolumn{1}{c}{\rotatebox{90}{\bfseries \ourapproach}} & \multicolumn{1}{c}{\rotatebox{90}{\cbi}} & \multicolumn{1}{c}{\rotatebox{90}{DD}} & \multicolumn{1}{c}{\rotatebox{90}{\encore}} & \multicolumn{1}{c|}{\rotatebox{90}{\bugdoc
        ~}} & \multicolumn{1}{c}{\rotatebox{90}{\bfseries \ourapproach}} &  \multicolumn{1}{c|}{\rotatebox{90}{Others}} \bigstrut[t]
        \\ \clineB{4-25}{2}
    \multicolumn{1}{c}{}&\multicolumn{1}{c}{}  & \multicolumn{1}{c}{} & \multicolumn{1}{c}{} & \multicolumn{1}{c}{} & \multicolumn{1}{c}{} & \multicolumn{1}{c}{} & \multicolumn{1}{c}{} \bigstrut\\[-1.4em]\hlineB{2}
    
     &  & \textsc{Xception} & \cellcolor{blue!10}\bfseries69 & 63 & 57 & 64 & 65 & \cellcolor{blue!10}\bfseries 75 & 56 & 56 & 60 & 66 & \cellcolor{blue!10}\bfseries68  & 62 & 58 & 64 & 69 & \cellcolor{blue!10}\bfseries4  & 3 & 2 & 2 & 3 & \cellcolor{blue!10}\bfseries0.6 &4 \\
     
     &  & \textsc{BERT} & \cellcolor{blue!10}\bfseries71 &62 & 61 & 61 & 62 & \cellcolor{blue!10}\bfseries72  & 56 & 59 & 56 & 61 & \cellcolor{blue!10}\bfseries 72 & 65 & 62 & 67 & 62 & \cellcolor{blue!10}\bfseries5  & 3 & 2 & 2 & 3 & \cellcolor{blue!10}\bfseries0.4 & 4 \\
     
     &  & \textsc{Deepspeech} & \cellcolor{blue!10}\bfseries71  & 61 & 64 & 62 & 67 & \cellcolor{blue!10}\bfseries71  & 58 & 59 & 54 & 68 & \cellcolor{blue!10}\bfseries69  & 67 & 66 & 68 & 67 & \cellcolor{blue!10}\bfseries3 & 3 & 2 & 2 & 2 & \cellcolor{blue!10}\bfseries0.7  & 4 \\
     
     & \multirow{-4}{*}{\rotatebox{90}{Heat}} & \textsc{x264} & \cellcolor{blue!10}\bfseries74  & 65 & 57 & 64 & 65 & \cellcolor{blue!10}\bfseries74  &62 & 54 & 55 & 65 & \cellcolor{blue!10}\bfseries74 & 66 & 63 & 68 & 69 & \cellcolor{blue!10}\bfseries7  & 3 & 2 & 2 & 5 & \cellcolor{blue!10}\bfseries1.4  & 4 \\ \hlineB{2}
    \end{tabular}
    }}
    \\
    \subfloat[Multi-objective non-functional faults for \textit{Heat, Latency} in \txtwo.]{
        \scriptsize
        \label{tab:rq2}
        \resizebox{\textwidth}{!}{
            \begin{tabular}{@{}r@{}ll|llll|llll|llll|llll|llll|ll|}
            \clineB{4-25}{2}
            &  &  & \multicolumn{4}{c|}{Accuracy} & \multicolumn{4}{c|}{Precision} & \multicolumn{4}{c|}{Recall} & \multicolumn{4}{c|}{Gain (Latency)} & \multicolumn{4}{c|}{Gain (Heat)}  & \multicolumn{2}{c|}{Time$^\dagger$} \bigstrut\\ \clineB{4-25}{2}
            
            &  &  & \rotatebox{90}{\bfseries \ourapproach~} & \rotatebox{90}{\cbi} & \rotatebox{90}{\encore} & \rotatebox{90}{\bugdoc} & \rotatebox{90}{\bfseries \ourapproach~}  & \rotatebox{90}{\cbi} & \rotatebox{90}{\encore} & \rotatebox{90}{\bugdoc} & \rotatebox{90}{\bfseries \ourapproach~} & \rotatebox{90}{\cbi} & \rotatebox{90}{\encore} & \rotatebox{90}{\bugdoc} & \rotatebox{90}{\bfseries \ourapproach~} & \rotatebox{90}{\cbi} & \rotatebox{90}{\encore} & \rotatebox{90}{\bugdoc} & \rotatebox{90}{\bfseries \ourapproach~} & \rotatebox{90}{\cbi} & \rotatebox{90}{\encore} & \rotatebox{90}{\bugdoc} & \rotatebox{90}{\bfseries \ourapproach~}  & \rotatebox{90}{Others} \\ \clineB{4-25}{2}
            
            \multicolumn{1}{l}{}&\multicolumn{1}{l}{}  & \multicolumn{1}{l}{} & \multicolumn{1}{l}{} & \multicolumn{1}{l}{} & \multicolumn{1}{l}{} & \multicolumn{1}{l}{} & \multicolumn{1}{l}{} & \multicolumn{1}{l}{} & \multicolumn{1}{l}{} \\[-0.85em]\hlineB{2}

            & \multicolumn{1}{l|}{} & \textsc{Xception} & \cellcolor{blue!10}\textbf{62} & 52 & 55 & 57 & \cellcolor{blue!10}\textbf{69} & 57 & 50 & 61 & \cellcolor{blue!10}\textbf{61} & 48 &51 & 60 & \cellcolor{blue!10}\textbf{58} & 42 & 47 & 51 & \cellcolor{blue!10}\textbf{2} & 1 & 1 & 1 & \cellcolor{blue!10}\textbf{0.9} & 4 \\
            
            & \multicolumn{1}{l|}{} & \textsc{BERT} & \cellcolor{blue!10}\textbf{64} & 52 & 47 & 56 & \cellcolor{blue!10}\textbf{62} & 52 & 45 & 60 & \cellcolor{blue!10}\textbf{68} & 54  & 62 & 65 & \cellcolor{blue!10}\textbf{65} & 37 & 48 & 60 & \cellcolor{blue!10}\textbf{4}  & 3 & 2 & 3 & \cellcolor{blue!10}\textbf{0.4}  & 4 \\
            
            & \multicolumn{1}{l|}{} & \textsc{Deepspeech} & \cellcolor{blue!10}\textbf{62}  & 52 & 43 & 55 & \cellcolor{blue!10}\textbf{60} & 48 & 48 & 55 & \cellcolor{blue!10}\textbf{67}  & 58 & 41 & 59 & \cellcolor{blue!10}\textbf{69}  & 37 & 45 & 65 & \cellcolor{blue!10}\textbf{4}  & 1 & 1 & 4  & \cellcolor{blue!10}\textbf{0.3}  & 4 \\
            
            \multirow{-4}{*}{\rotatebox{90}{Latency +}} & \multicolumn{1}{l|}{\multirow{-4}{*}{\rotatebox{90}{Heat}}} & \multicolumn{1}{l|}{\textsc{x264}} & \cellcolor{blue!10}\textbf{61}  & 53 & 53 & 60 & \cellcolor{blue!10}\textbf{63}  & 50 & 54 & 61 & \cellcolor{blue!10}\textbf{60}  & 53 & 55 & 55 & \cellcolor{blue!10}\textbf{67}  & 54 & 54 & 65 & \cellcolor{blue!10}\textbf{5}  & 3 & 3 & 4  & \cellcolor{blue!10}\textbf{0.5}  & 4 \\
            \multicolumn{1}{l}{}&\multicolumn{1}{l}{}  & \multicolumn{1}{l}{} & \multicolumn{1}{l}{} & \multicolumn{1}{l}{} & \multicolumn{1}{l}{} & \multicolumn{1}{l}{} & \multicolumn{1}{l}{} & \multicolumn{1}{l}{} & \multicolumn{1}{l}{} \\[-0.85em]\hlineB{2}

           \end{tabular}
           
    }}
     \\
    \subfloat[Multi-objective non-functional faults for \textit{Energy, Heat} in \xavier.]{
        \scriptsize
        \label{tab:rq2}
        \resizebox{\textwidth}{!}{
            \begin{tabular}{@{}r@{}ll|llll|llll|llll|llll|llll|ll|}
            \clineB{4-25}{2}
            &  &  & \multicolumn{4}{c|}{Accuracy} & \multicolumn{4}{c|}{Precision} & \multicolumn{4}{c|}{Recall} & \multicolumn{4}{c|}{Gain (Energy)} & \multicolumn{4}{c|}{Gain (Heat)}  & \multicolumn{2}{c|}{Time$^\dagger$} \bigstrut\\ \clineB{4-25}{2}
            
            &  &  & \rotatebox{90}{\bfseries \ourapproach~} & \rotatebox{90}{\cbi} & \rotatebox{90}{\encore} & \rotatebox{90}{\bugdoc} & \rotatebox{90}{\bfseries \ourapproach~}  & \rotatebox{90}{\cbi} & \rotatebox{90}{\encore} & \rotatebox{90}{\bugdoc} & \rotatebox{90}{\bfseries \ourapproach~} & \rotatebox{90}{\cbi} & \rotatebox{90}{\encore} & \rotatebox{90}{\bugdoc} & \rotatebox{90}{\bfseries \ourapproach~} & \rotatebox{90}{\cbi} & \rotatebox{90}{\encore} & \rotatebox{90}{\bugdoc} & \rotatebox{90}{\bfseries \ourapproach~} & \rotatebox{90}{\cbi} & \rotatebox{90}{\encore} & \rotatebox{90}{\bugdoc} & \rotatebox{90}{\bfseries \ourapproach~}  & \rotatebox{90}{Others} \\ \clineB{4-25}{2}
            
            \multicolumn{1}{l}{}&\multicolumn{1}{l}{}  & \multicolumn{1}{l}{} & \multicolumn{1}{l}{} & \multicolumn{1}{l}{} & \multicolumn{1}{l}{} & \multicolumn{1}{l}{} & \multicolumn{1}{l}{} & \multicolumn{1}{l}{} & \multicolumn{1}{l}{} \\[-0.85em]\hlineB{2}

            & \multicolumn{1}{l|}{} & \textsc{Xception} & \cellcolor{blue!10}\textbf{65} & 55 & 57 & 63 & \cellcolor{blue!10}\textbf{64} & 55 & 51 & 62 & \cellcolor{blue!10}\textbf{67} & 47 &53 & 60 & \cellcolor{blue!10}\textbf{58} & 44 & 51 & 54 & \cellcolor{blue!10}\textbf{3} & 1 & 1 & 1 & \cellcolor{blue!10}\textbf{0.8} & 4 \\
            
            & \multicolumn{1}{l|}{} & \textsc{BERT} & \cellcolor{blue!10}\textbf{69} & 55 & 51 & 59 & \cellcolor{blue!10}\textbf{65} & 53 & 47 & 61 & \cellcolor{blue!10}\textbf{71} & 53 & 61 & 67 & \cellcolor{blue!10}\textbf{65} & 41 & 51& 61 & \cellcolor{blue!10}\textbf{4}  & 2 & 2 & 3 & \cellcolor{blue!10}\textbf{0.4}  & 4 \\
            
            & \multicolumn{1}{l|}{} & \textsc{Deepspeech} & \cellcolor{blue!10}\textbf{72}  & 55 & 49 & 61 & \cellcolor{blue!10}\textbf{73} & 51 & 51 & 61 & \cellcolor{blue!10}\textbf{71}  & 57 & 53 & 64 & \cellcolor{blue!10}\textbf{69}  & 47 & 51 & 64 & \cellcolor{blue!10}\textbf{4}  & 1 & 1 & 3  & \cellcolor{blue!10}\textbf{0.3}  & 4 \\
            
            \multirow{-4}{*}{\rotatebox{90}{Energy +}} & \multicolumn{1}{l|}{\multirow{-4}{*}{\rotatebox{90}{Heat}}} & \multicolumn{1}{l|}{\textsc{x264}} & \cellcolor{blue!10}\textbf{72}  & 59 & 57 & 66 & \cellcolor{blue!10}\textbf{71}  & 51 & 55 & 62 & \cellcolor{blue!10}\textbf{69}  & 61 & 59 & 59 & \cellcolor{blue!10}\textbf{67}  & 51 & 51 & 61 & \cellcolor{blue!10}\textbf{5}  & 2 & 3 & 4  & \cellcolor{blue!10}\textbf{0.5}  & 4 \\
            \multicolumn{1}{l}{}&\multicolumn{1}{l}{}  & \multicolumn{1}{l}{} & \multicolumn{1}{l}{} & \multicolumn{1}{l}{} & \multicolumn{1}{l}{} & \multicolumn{1}{l}{} & \multicolumn{1}{l}{} & \multicolumn{1}{l}{} & \multicolumn{1}{l}{} \\[-0.85em]\hlineB{2}

           \end{tabular}
           
    }}
    \\
    \subfloat[Multi-objective non-functional faults for \textit{Energy, Heat, and Latency} in \txtwo.]{
        \scriptsize

    \footnotesize
    \resizebox{\linewidth}{!}{
        \begin{tabular}{@{}l@{}ll|llll|llll|llll|llll|llll|llll|ll|}
            \clineB{4-29}{2}
            &  &  & \multicolumn{4}{c|}{Accuracy} & \multicolumn{4}{c|}{Precision} & \multicolumn{4}{c|}{Recall} & \multicolumn{4}{c|}{Gain (Latency)} & \multicolumn{4}{c|}{Gain (Energy)} & \multicolumn{4}{c|}{Gain (Heat)} & \multicolumn{2}{c|}{Time$^\dagger$} \bigstrut\\ \clineB{4-29}{2}
            
            &  &  & \rotatebox{90}{\bfseries\ourapproach~} & \rotatebox{90}{\cbi} & \rotatebox{90}{\encore} & \rotatebox{90}{\bugdoc} & \rotatebox{90}{\bfseries\ourapproach~} & \rotatebox{90}{\cbi} & \rotatebox{90}{\encore} & \rotatebox{90}{\bugdoc} & \rotatebox{90}{\bfseries\ourapproach~} & \rotatebox{90}{\cbi} & \rotatebox{90}{\encore} & \rotatebox{90}{\bugdoc} & \rotatebox{90}{\bfseries\ourapproach~} & \rotatebox{90}{\cbi} & \rotatebox{90}{\encore} & \rotatebox{90}{\bugdoc} & \rotatebox{90}{\bfseries\ourapproach~} & \rotatebox{90}{\cbi} & \rotatebox{90}{\encore} & \rotatebox{90}{\bugdoc} & \rotatebox{90}{\bfseries\ourapproach~} & \rotatebox{90}{\cbi} & \rotatebox{90}{\encore} & \rotatebox{90}{\bugdoc} & \rotatebox{90}{\bfseries\ourapproach~} & \rotatebox{90}{Others} \\ \clineB{4-29}{2}
            
           \multicolumn{1}{l}{}&\multicolumn{1}{l}{}  & \multicolumn{1}{l}{} & \multicolumn{1}{l}{} & \multicolumn{1}{l}{} & \multicolumn{1}{l}{} & \multicolumn{1}{l}{} & \multicolumn{1}{l}{} & \multicolumn{1}{l}{} & \multicolumn{1}{l}{} \\[-0.95em]\hlineB{2}

            & \multicolumn{1}{l|}{} & Image & \cellcolor{blue!10}\textbf{76} & 57 & 48 & 66 & \cellcolor{blue!10}\textbf{68} & 61 & 57 & 61 & \cellcolor{blue!10}\textbf{81} & 53 & 46 & 70 & \cellcolor{blue!10}\textbf{62} & 33 & 30 & 42 & \cellcolor{blue!10}\textbf{52} & 23 & 18 & 24 & \cellcolor{blue!10}\textbf{4} & 1 & 0 & 0 & \cellcolor{blue!10}\textbf{0.1} & 4 \\
            & \multicolumn{1}{l|}{} & x264 & \cellcolor{blue!10}\textbf{80} & 59 & 47 & 54 & \cellcolor{blue!10}\textbf{76} & 61 & 56 & 63 & \cellcolor{blue!10}\textbf{81} & 56 & 46 & 51 & \cellcolor{blue!10}\textbf{12} & 2 & 1 & 2 & \cellcolor{blue!10}\textbf{15} & 4 & 2 & 4 & \cellcolor{blue!10}\textbf{4} & 1 & 0 & 1 & \cellcolor{blue!10}\textbf{0.1} & 4 \\
           \multirow{-3}{*}{\rotatebox{90}{All}} & \multicolumn{1}{l|}{\multirow{-3}{*}{\rotatebox{90}{Three}}} & SQLite & \cellcolor{blue!10}\textbf{73} & 56 & 51 & 53 & \cellcolor{blue!10}\textbf{68} & 59 & 56 & 60 & \cellcolor{blue!10}\textbf{78} & 54 & 45 & 51 & \cellcolor{blue!10}\textbf{12} & 1 & 1 & 4 & \cellcolor{blue!10}\textbf{8} & 4 & 2 & 5 & \cellcolor{blue!10}\textbf{1} & 1 & \cellcolor[HTML]{FFCCC9}-1 & \cellcolor[HTML]{FFCCC9}-1 & \cellcolor{blue!10}\textbf{0.1} & 4 \\ \hlineB{2}
           \multicolumn{10}{l}{$^\dagger$ Wallclock time in hours}\bigstrut
           \end{tabular}
    }}
    \label{tab:rq_effectiveness_appendix}
\end{table*}


\subsubsection{Effectiveness.}
~\Cref{tab:rq_effectiveness_appendix}(a) shows the effectiveness of \ourapproach in resolving single objective faults due to heat in NVIDIA \txone. Here, \ourapproach ~{ outperforms correlation-based methods in all cases}. For example, in  \textsc{Bert} on TX1, \ourapproach achieves 9\% more accuracy, 11\% more precision, and 10\% more recall compared to the next best method, \bugdoc. We observed heat gains as high as $7\%$ ($2\%$ more than \bugdoc) on \textsc{x264}. The results confirm that \ourapproach~{\em can recommend repairs for faults that significantly improve latency and energy usage}. Applying the changes to the configurations recommended by \ourapproach increases the performance drastically.

\ourapproach~{\em can resolve misconfiguration faults significantly faster than correlation-based approaches}.~In~\Cref{tab:rq_effectiveness_appendix}, the last two columns indicate the time taken (in hours) by each approach to diagnosing the root cause. \ourapproach can do resolve faults significantly faster, \eg, \ourapproach is $13\times$ faster in diagnosing and resolving latency and heat faults for \textsc{Deepspeech}.

\subsubsection{Transferability.}
\begin{table}
    \caption{\small Transferring causal models across hardware platforms. Cells highlighted in \colorbox{blue!10}{\bfseries blue} indicate the transferability potential of \ourapproach when compared to \ourapproach \textsc(Rerun).  }
    \label{tab:transfer}
    \centering
    \resizebox{\linewidth}{!}{
    \begin{tabular}{ll|rrr|rrr|rrr|rrr}
        \hlineB{2}
        \multicolumn{14}{c}{TX1 (source) $\longrightarrow$ TX2 (target)} \bigstrut\\ \hlineB{2}

        & \multicolumn{1}{l|}{} & \multicolumn{3}{c|}{Accuracy} & \multicolumn{3}{c|}{Recall} & \multicolumn{3}{c|}{Precision} & \multicolumn{3}{c}{$\Delta_{gain}$} 
        
        \bigstrut\\ \clineB{3-14}{2} 
        
        &  Software & \rotatebox{90}{\ourapproach \textsc{(Reuse)}} & \multicolumn{1}{r}{\rotatebox{90}{\ourapproach+25}} & \rotatebox{90}{\ourapproach \textsc{(Rerun)}} & \rotatebox{90}{\ourapproach \textsc{(Reuse)}} & \multicolumn{1}{r}{\rotatebox{90}{\ourapproach+25}} & \rotatebox{90}{\ourapproach \textsc{(Rerun)}} & \rotatebox{90}{\ourapproach \textsc{(Reuse)}} & \multicolumn{1}{r}{\rotatebox{90}{\ourapproach+25}} & \rotatebox{90}{\ourapproach \textsc{(Rerun)}} & \rotatebox{90}{\ourapproach \textsc{(Reuse)}} & \multicolumn{1}{r}{\rotatebox{90}{\ourapproach+25}} & \rotatebox{90}{\ourapproach \textsc{(Rerun)}} \bigstrut\\ \hlineB{2}       

    \multicolumn{1}{l|}{\multirow{5}{*}{\rotatebox{90}{Latency}}} & \textsc{Xception} & 52 & \cellcolor{blue!10}83 & 86 & 70 & \cellcolor{blue!10}79 & 86 & 46 & \cellcolor{blue!10}78 & 83 & 46 & \cellcolor{blue!10}71 & 82 \bigstrut\\
    \multicolumn{1}{l|}{} & \textsc{Bert} & 55 & \cellcolor{blue!10}75 & 81 & 57 & \cellcolor{blue!10}70 & 71 & 45 & \cellcolor{blue!10}67 & 76 & 43 & \cellcolor{blue!10}70 & 74 \bigstrut\\
    \multicolumn{1}{l|}{} & \textsc{Deepspeech} & 45 & \cellcolor{blue!10}71 & 81 & 56 & \cellcolor{blue!10}79 & 81 & 49 & \cellcolor{blue!10}73 & 76 & 54 & \cellcolor{blue!10}73 & 76 \bigstrut\\
    \multicolumn{1}{l|}{} & \textsc{x264} & 57 & \cellcolor{blue!10}79 & 83 & 70 & \cellcolor{blue!10}75 & 78 & 58 & \cellcolor{blue!10}77 & 82 & 45 & \cellcolor{blue!10}73 & 85 \bigstrut\\
     \hlineB{2}
    \multicolumn{14}{c}{TX2 (source) $\longrightarrow$ \xavier (target)} \bigstrut\\ \hlineB{2}
    \multicolumn{1}{l|}{\multirow{5}{*}{\rotatebox{90}{Energy}}} & \textsc{Xception} & 53 & \cellcolor{blue!10}74 & 84 & 48 & \cellcolor{blue!10}73 & 80 & 51 & \cellcolor{blue!10}69 & 78 & 43 & \cellcolor{blue!10}73 & 83 \bigstrut\\
    \multicolumn{1}{l|}{} & \textsc{Bert} & 50 & \cellcolor{blue!10}61 & 66 & 53 & \cellcolor{blue!10}71 & 79 & 49 & \cellcolor{blue!10}66 & 70 & 40 & \cellcolor{blue!10}55 & 62 \bigstrut\\
    \multicolumn{1}{l|}{} & \textsc{Deepspeech} & 57 & \cellcolor{blue!10}70 & 73 & 45 & \cellcolor{blue!10}74 & 78 & 43 & \cellcolor{blue!10}69 & 75 & 49 & \cellcolor{blue!10}71 & 78 \bigstrut\\
    \multicolumn{1}{l|}{} & \textsc{x264} & 54 & \cellcolor{blue!10}72 & 77 & 46 & \cellcolor{blue!10}72 & 78 & 42 & \cellcolor{blue!10}75 & 83 & 46 & \cellcolor{blue!10}79 & 87 \bigstrut\\
    \hlineB{2}
    \multicolumn{14}{c}{\xavier (source) $\longrightarrow$ TX1 (target)} \bigstrut\\ \hlineB{2}
    \multicolumn{1}{l|}{\multirow{5}{*}{\rotatebox{90}{Heat}}} & \textsc{Xception}& 63 & \cellcolor{blue!10}64 & 69 & 61 & \cellcolor{blue!10}67 & 68 & 58 & \cellcolor{blue!10}74 & 75 & 3 & \cellcolor{blue!10}4 & 4 \bigstrut\\
    \multicolumn{1}{l|}{} & \textsc{Bert} & 55 & \cellcolor{blue!10}65 & 71 & 59 & \cellcolor{blue!10}67 & 72 & 52 & \cellcolor{blue!10}64 & 72 & 3 & \cellcolor{blue!10}4 & 5 \bigstrut\\
    \multicolumn{1}{l|}{} & \textsc{Deepspeech} & 57 & \cellcolor{blue!10}64 & 71 & 59 & \cellcolor{blue!10}63 & 69 & 53 & \cellcolor{blue!10}63 & 71 & 1 & \cellcolor{blue!10}2 & 3 \bigstrut\\
    \multicolumn{1}{l|}{} & \textsc{x264} & 51 & \cellcolor{blue!10}65 & 74 & 53 & \cellcolor{blue!10}64 & 74 & 54 & \cellcolor{blue!10}69 & 74 & 3 & \cellcolor{blue!10}5 & 7 \bigstrut\\
     \hlineB{2}
    \end{tabular}}
    \end{table}
    
Table~\ref{tab:transfer} indicates the results for different transfer scenarios: (I) We learn a causal model from \txone and use them to resolve the latency faults in \txtwo,  (I) We learn a causal model from \txtwo and use them to resolve the energy faults in \xavier, and (III) We learn a causal model from \xavier and use them to resolve the heat faults in \txone. Here, we determine how transferable is \ourapproach by comparing with \ourapproach \textsc(Reuse), \ourapproach+25, and \ourapproach \textsc(Rerun). For all systems, we observe that the performance of \ourapproach \textsc(Reuse) is close to the performance of \ourapproach \textsc(Rerun) which confirms the high transferability property of \ourapproach. For example, in \textsc{Xception} and \textsc{SQLite}, \ourapproach \textsc(Reuse) has the exact gain as of \ourapproach \textsc(Rerun) for heat faults. For latency and energy faults, the main difference between \ourapproach \textsc(Reuse) and \ourapproach \textsc(Rerun) is less than 5\% for all systems. We also observe that with little updates, \ourapproach+25 ($\sim$24 minutes) achieves a similar performance of \ourapproach \textsc{(Rerun)} ($\sim$40 minutes), on average. This confirms that as the causal mechanisms are sparse, the causal performance model from source in \ourapproach quickly reaches a fixed structure in the target using incremental learning by judiciously evaluating the most promising fixes until the fault is resolved.

\subsubsection{Scalability.}

The scalability of \ourapproach depends on the scalability of each phase. Therefore, we design scenarios to test the scalability of each phase to determine the overall scalability. Since the initial number of samples and the underlying phases for each task is the same, it is sufficient to examine the scalability of \ourapproach for the debugging non-functional fault task.

\textsc{SQLite} was chosen because it offers a large number of configurable options, much more than neural applications, and video encoders. Further, each of these options can take on a large number of permitted values, making \textsc{Deepstream} a useful candidate to study the scalability of \ourapproach. \textsc{Deepstream} was chosen as it has a higher number of components than others, and it is interesting to determine how \ourapproach behaves when the number of options and events are increasing. As a result, SQLite exposes new system design opportunities to enable efficient inference and many complex interactions between software options.

In large systems, there are significantly more causal paths and therefore, causal learning and estimations of queries take more time. However, with as many as 242 configuration options and 19 events (\tab{scalability}, row 2), causal graph discovery takes roughly one minute, evaluating all 2234 queries takes roughly two minutes, and the total time to diagnose and fix a fault is roughly 22 minutes for \textsc{SQLite}. This trend is observed even with 242 configuration options, 288 events (\tab{scalability}, row 3), and finer granularity of configuration values---the time required to causal model recovery is a little over 1 minute and the total time to diagnose and fix a fault is less than 2 hours. Similarly, in \textsc{Deepstream}, with 53 configuration options and 288 events, causal model discovery is less than two minutes and the time needed to diagnose and fix a fault is less than an hour. 
The results in ~\tab{scalability} indicate that \ourapproach can scale to a much larger configuration space without an exponential increase in runtime for any of the intermediate stages. This can be attributed to the sparsity of the causal graph (average degree of a node for \textsc{SQLite} in \tab{scalability} is at most 3.6, and it reduces to 1.6 when the number of configurations increase and reduces from 3.1 to 2.3 in \textsc{Deepstream} when systems events are increased). This makes sense because not all variables (\ie, configuration options and/or system events) affect non-functional properties and a high number of variables in the graph end up as isolated nodes. Therefore, the number of paths and consequently the evaluation time do not grow exponentially as the number of variables increases.


Finally, the latency gain associated with repairs from larger configuration space with configurations was similar to the original space of 34 and 53 configurations for \textsc{SQLite} and \textsc{Deepstream}, respectively. This indicates that: (a)~imparting domain expertise to select most important configuration options can speed up the inference time of \ourapproach, and (b)~if the user chooses instead to use more configuration options (perhaps to avoid initial feature engineering), \ourapproach can still diagnose and fix faults satisfactorily within a reasonable time.

\end{document}